\documentclass[10pt,journal,compsoc]{IEEEtran}
\usepackage{booktabs}
\usepackage{multirow}
\usepackage{xcolor}
\usepackage{graphicx}
\usepackage{ifpdf}
\usepackage{ragged2e}
\definecolor{mygreen}{RGB}{0,102,0}
\definecolor{myblue}{RGB}{0,122,122}
\usepackage[colorlinks,linkcolor=black,citecolor=black]{hyperref}
\usepackage{graphicx}
\usepackage{amsmath,amsfonts}
\usepackage{algorithmic}
\usepackage{array}
\usepackage[caption=false,font=tiny,labelfont=sf,textfont=sf]{subfig}
\usepackage{textcomp}
\usepackage{stfloats}
\usepackage{url}
\usepackage{verbatim}
\usepackage{tikz}
\usetikzlibrary{calc}
\usetikzlibrary{arrows} 
\usetikzlibrary{arrows.meta, calc}
\usepackage{caption}
\DeclareCaptionFormat{myformat}{#1#2#3}
\usepackage{balance}

\ifCLASSOPTIONcompsoc
  \usepackage[nocompress]{cite}
\else
  \usepackage{cite}
\fi
\ifCLASSINFOpdf
\else
\fi
\begin{document}
\title{DualGazeNet: A Biologically Inspired Dual-Gaze Query Network for Salient Object Detection}
%
%
%
%

\author{Yu Zhang, Haoan Ping, Yuchen Li, Zhenshan Bing, Fuchun Sun,~\IEEEmembership{Fellow,~IEEE,}\\ Alois Knoll,~\IEEEmembership{Fellow,~IEEE} 
\IEEEcompsocitemizethanks{


\IEEEcompsocthanksitem Yu Zhang, Haoan Ping, and Alois Knoll are with the School of Computation, Information and Technology, Technical University of Munich, 80333 München, Germany.

\IEEEcompsocthanksitem Yuchen Li is with the Department of Informatics, Technical University of Munich, Munich, 85748, Germany, and the University Research and Innovation Center, Obuda University, Budapest, H-1034, Hungary.

\IEEEcompsocthanksitem Zhenshan Bing is with the School of Computation, Information and Technology, Technical University of Munich, 80333 München, Germany, also with the State Key Laboratory for Novel Software Technology, Nanjing University Suzhou Campus, Suzhou 215163, China, and also with the School of Science and Technology, Nanjing University Suzhou Campus, Suzhou 215163, China.


\IEEEcompsocthanksitem Fuchun Sun is with the Department of Computer Science and Technology, Tsinghua University, Beijing 100190, China.}
}

\IEEEtitleabstractindextext{%
\begin{abstract}
Recent salient object detection (SOD) methods aim to improve performance in four key directions: semantic enhancement, boundary refinement, auxiliary task supervision, and multi-modal fusion. In pursuit of continuous gains, these approaches have evolved toward increasingly sophisticated architectures with multi-stage pipelines, specialized fusion modules, edge-guided learning, and elaborate attention mechanisms. However, this complexity paradoxically introduces feature redundancy and cross-component interference that obscure salient cues, ultimately reaching performance bottlenecks. In contrast, human vision achieves efficient salient object identification without such architectural complexity. This contrast raises a fundamental question: can we design a biologically grounded yet architecturally simple SOD framework that dispenses with most of this engineering complexity, while achieving state-of-the-art accuracy, computational efficiency, and interpretability? In this work, we answer this question affirmatively by introducing DualGazeNet, a biologically inspired pure Transformer framework that models the dual biological principles of robust representation learning and magnocellular–parvocellular dual-pathway processing with cortical attention modulation in the human visual system. Extensive experiments on five RGB SOD benchmarks show that DualGazeNet consistently surpasses 25 state-of-the-art CNN- and Transformer-based methods. On average, DualGazeNet achieves about 60$\%$ higher inference speed and 53.4$\%$ fewer FLOPs than four Transformer-based baselines of similar capacity (VST++, MDSAM, Sam2unet, and BiRefNet). Moreover, DualGazeNet exhibits strong cross-domain generalization, achieving leading or highly competitive performance on camouflaged and underwater SOD benchmarks without relying on additional modalities. Code and results are available at: \justifying{
 \href{https://github.com/jeremypha/DualGazeNet}{https://github.com/jeremypha/DualGazeNet}.}

\end{abstract}

\begin{IEEEkeywords}
Salient object detection, vision transformer, query learning, and bio-inspired mechanism.
\end{IEEEkeywords}}

\maketitle

\IEEEdisplaynontitleabstractindextext

%
\IEEEpeerreviewmaketitle

\IEEEraisesectionheading{\section{Introduction}\label{sec:introduction}}

%
%
%
%
\IEEEPARstart{S}ALIENT object detection (SOD) represents a fundamental computer vision task that models human visual attention mechanisms to identify and segment the most perceptually significant objects within an image at pixel-level precision \cite{11072100}. As a pre-attentive vision task, SOD suppresses background clutter and irrelevant information, thereby benefiting numerous downstream computer vision applications, such as semantic segmentation \cite{9933726,10203634}, object recognition and detection \cite{10549838,6587754}, and object tracking \cite{zhang2020non,hong2015online}.

\begin{figure}[!t]
\vspace{-5pt}
  \centering
  \begin{tikzpicture}
    \begin{scope}[xshift=-2.7cm, yshift=0cm ]
        \node[above right] (fig63) at (0,0){\includegraphics[ height=0.311\linewidth]{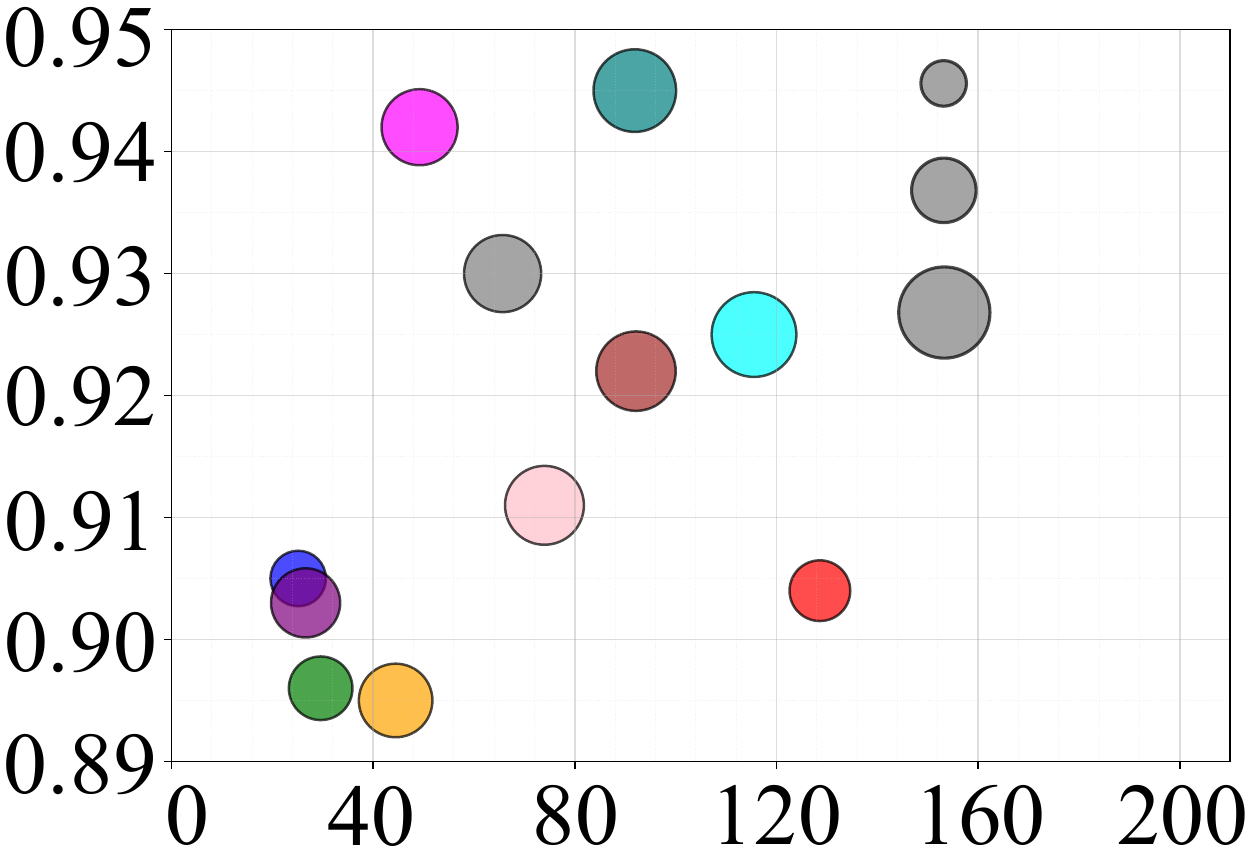}};
        \node at ($(fig63.south)+(0,-0.35)$) {\footnotesize (a) latency under 20 ms};
        \node[inner sep=1pt, font=\footnotesize] at (3.75,1.68+0.22-0.03) {20ms};
        \node[inner sep=1pt, font=\footnotesize] at (3.75,2.08+0.22-0.04) {10ms};
        \node[inner sep=1pt, font=\footnotesize] at (3.82,2.43+0.22-0.03) {5ms};
        \draw[black, line width=0.4pt] (3.05-0.05,1.45+0.22) rectangle (4.1,2.6+0.19); 
            \node at ($(fig63.south)+(0.05,-0.02)$) {\scriptsize Parameters (M)};
            \node[rotate=90] at ($(fig63.west)+(-0.1,0.08)$) {\scriptsize max F-measure (DUTS)};
    \end{scope}
    \begin{scope}[xshift=1.65cm, yshift=0.04cm]
        \node[above right] (fig64) at (0,0){\includegraphics[ height=0.31\linewidth]{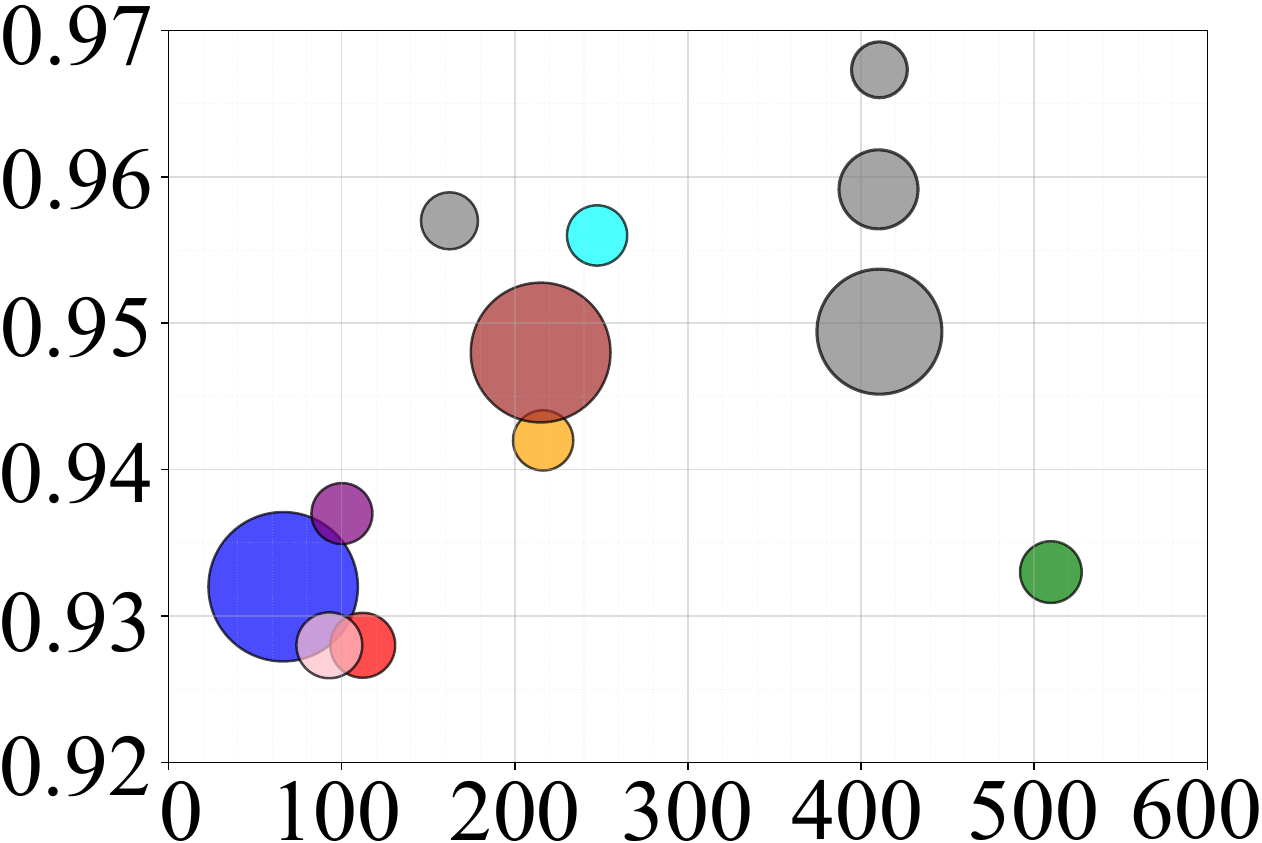}};
        \node at ($(fig64.south)+(0,-0.35)$) {\footnotesize (b) latency above 20 ms};
        \node[inner sep=1pt, font=\footnotesize] at (3.75-0.1,1.68+0.15-0.04) {100ms};
        \node[inner sep=1pt, font=\footnotesize] at (3.75-0.04,2.08+0.2-0.02) {40ms};
        \node[inner sep=1pt, font=\footnotesize] at (3.82-0.1,2.43+0.21) {20ms};
        \draw[black, line width=0.4pt] (3.05-0.30,1.45+0.10) rectangle (4.067,2.6+0.18);
            \node[rotate=90] at ($(fig64.west)+(-0.05,0.08)$) {\scriptsize max F-measure (DUTS)};
            \node at ($(fig64.south)+(0.10,-0.02)$) {\scriptsize Parameters (M)};
    \end{scope}
  \end{tikzpicture}
  \begin{tikzpicture}
    \definecolor{blue}{HTML}{0000FF}      
    \definecolor{green}{HTML}{008000}     
    \definecolor{red}{HTML}{FF0000}       
    \definecolor{purple}{HTML}{800080}    
    \definecolor{orange}{HTML}{FFA500}    
    \definecolor{brown}{HTML}{8B4513}     
    \definecolor{pink}{HTML}{FFC0CB}      
    \definecolor{gray}{HTML}{808080}      
    \definecolor{cyan}{HTML}{00FFFF}      
    \definecolor{magenta}{HTML}{FF00FF}   
    \definecolor{teal}{HTML}{008080}      
    
    \node[right, text=black, font=\footnotesize] at (0.15-2.8,0.6) {(a):};
    \filldraw[fill=orange, fill opacity=0.7, draw=black, line width=0.2pt] (-2+0.05,0.6) circle (2.5pt);
    \node[right, text=black, font=\footnotesize] at (-1.9+0.05,0.6) {VST \cite{Liu_2021_ICCV}};
    \filldraw[fill=green, fill opacity=0.7, draw=black, line width=0.2pt] (-0.5,0.6) circle (2.5pt);
    \node[right, text=black, font=\footnotesize] at (-0.4,0.6) {DFI \cite{dfi}};
    \filldraw[fill=purple, fill opacity=0.7, draw=black, line width=0.2pt] (1.0,0.6) circle (2.5pt);
    \node[right, text=black, font=\footnotesize] at (1.1,0.6) {RMFDNet \cite{zhou2025rmfdnet}};
    \filldraw[fill=red, fill opacity=0.7, draw=black, line width=0.2pt] (3.3,0.6) circle (2.5pt);
    \node[right, text=black, font=\footnotesize] at (3.4,0.6) {GateNet \cite{gatenet}};

    \filldraw[fill=blue, fill opacity=0.7, draw=black, line width=0.2pt] (-2+0.05,0.3) circle (2.5pt);
    \node[right, text=black, font=\footnotesize] at (-1.9+0.05,0.3) {LDF \cite{ldf}};
    \filldraw[fill=pink, fill opacity=0.7, draw=black, line width=0.2pt] (-0.35,0.3) circle (2.5pt);
    \node[right, text=black, font=\footnotesize] at (-0.25,0.3) {BBRF \cite{bbrf}};

    \filldraw[fill=brown, fill opacity=0.7, draw=black, line width=0.2pt] (1.4,0.3) circle (2.5pt);
    \node[right, text=black, font=\footnotesize] at (1.5,0.3) {ICON-S \cite{icon}};
    \filldraw[fill=cyan, fill opacity=0.7, draw=black, line width=0.2pt] (3.45,0.3) circle (2.5pt);
     \node[right, text=black, font=\footnotesize] at (3.55,0.3) {DGN-Swin};

     \filldraw[fill=gray, fill opacity=0.7, draw=black, line width=0.2pt] (-2+0.05,0.0) circle (2.5pt);
    \node[right, text=black, font=\footnotesize] at (-1.9+0.05,0.0) {SwinSOD \cite{WU2024105039}};

    \filldraw[fill=magenta, fill opacity=0.7, draw=black, line width=0.2pt] (0.3,0.0) circle (2.5pt);
    \node[right, text=black, font=\footnotesize] at (0.4,0.0) {DGN$^{*}$-B (pruned-ours)};

    \filldraw[fill=teal, fill opacity=0.7, draw=black, line width=0.2pt] (3.7,0.0) circle (2.5pt);
    \node[right, text=black, font=\footnotesize] at (3.8,0.0) {DGN-B (ours)};

    \draw[black, line width=0.5pt] (-2.55,-0.15) rectangle (6.25,0.85);
  \end{tikzpicture}
  \begin{tikzpicture}
   \definecolor{blue}{HTML}{0000FF}      
    \definecolor{green}{HTML}{008000}     
    \definecolor{red}{HTML}{FF0000}       
    \definecolor{purple}{HTML}{800080}    
    \definecolor{orange}{HTML}{FFA500}    
    \definecolor{brown}{HTML}{8B4513}     
    \definecolor{pink}{HTML}{FFC0CB}      
    \definecolor{gray}{HTML}{808080}      
    \definecolor{cyan}{HTML}{00FFFF}      
    
    \node[right, text=black, font=\footnotesize] at (0.15-2.8,0.6) {(b):};
    \filldraw[fill=red, fill opacity=0.7, draw=black, line width=0.2pt] (-2+0.05,0.6) circle (2.5pt);
    \node[right, text=black, font=\footnotesize] at (-1.9+0.05,0.6) {VST++ \cite{10497889}};
    \filldraw[fill=pink, fill opacity=0.7, draw=black, line width=0.2pt] (0.0,0.6) circle (2.5pt);
    \node[right, text=black, font=\footnotesize] at (0.1,0.6) {DGN-pvt};
    \filldraw[fill=blue, fill opacity=0.7, draw=black, line width=0.2pt] (1.7,0.6) circle (2.5pt);
    \node[right, text=black, font=\footnotesize] at (1.8,0.6) {TE7 \cite{tracer}};
    \filldraw[fill=green, fill opacity=0.7, draw=black, line width=0.2pt] (3.25,0.6) circle (2.5pt);
    \node[right, text=black, font=\footnotesize] at (3.35,0.6) {DC-Net-S \cite{dcnet}};

    \filldraw[fill=purple, fill opacity=0.7, draw=black, line width=0.2pt] (-2+0.05,0.3) circle (2.5pt);
    \node[right, text=black, font=\footnotesize] at (-1.9+0.05,0.3) {SwinSOD \cite{WU2024105039}};
    \filldraw[fill=orange, fill opacity=0.7, draw=black, line width=0.2pt] (0.3,0.3) circle (2.5pt);
    \node[right, text=black, font=\footnotesize] at (0.4,0.3) {Sam2unet \cite{sam2unet}};
    \filldraw[fill=brown, fill opacity=0.7, draw=black, line width=0.2pt] (2.6,0.3) circle (2.5pt);
    \node[right, text=black, font=\footnotesize] at (2.7,0.3) {BiRefNet \cite{birefnet}};

    \filldraw[fill=gray, fill opacity=0.7, draw=black, line width=0.2pt] (-2+0.05,0.0) circle (2.5pt);
    \node[right, text=black, font=\footnotesize] at (-1.9+0.05,0.0) {DGN$^{*}$-L (pruned-ours)};
    \filldraw[fill=cyan, fill opacity=0.7, draw=black, line width=0.2pt] (1.5,0.0) circle (2.5pt);
    \node[right, text=black, font=\footnotesize] at (1.6,0.0) {DGN-L (ours)};

    \draw[black, line width=0.5pt] (-2.55,-0.15) rectangle (6.25,0.85);
  \end{tikzpicture}
  \caption{Accuracy-efficiency trade-offs of SOTA RGB-SOD methods on DUTS-TE. Circle size indicates inference latency (ms). Results are illustrated separately for (a) lightweight and (b) heavy models to avoid visual clutter and enhance readability.}
  \label{fig2}
    \vspace{-0.5em}
\end{figure}
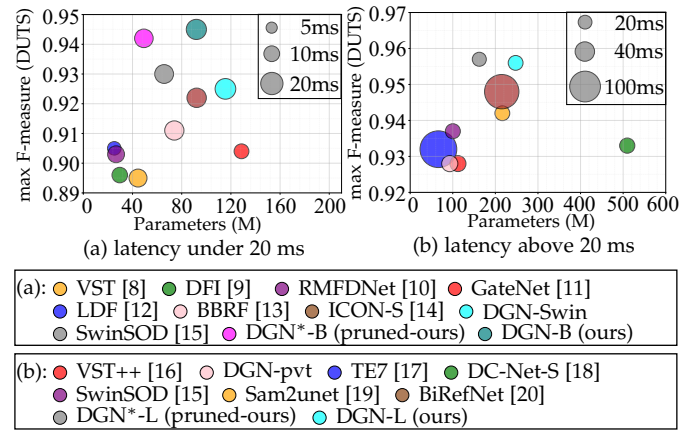

\begin{figure}[!t]
  \centering
  \begin{tikzpicture}%
    \begin{scope}[xshift=0cm, yshift=0cm ]
        \node[above right] (fig63) at (0.0,0.4){\hspace{-0.1cm}\includegraphics[ height=0.5\linewidth]{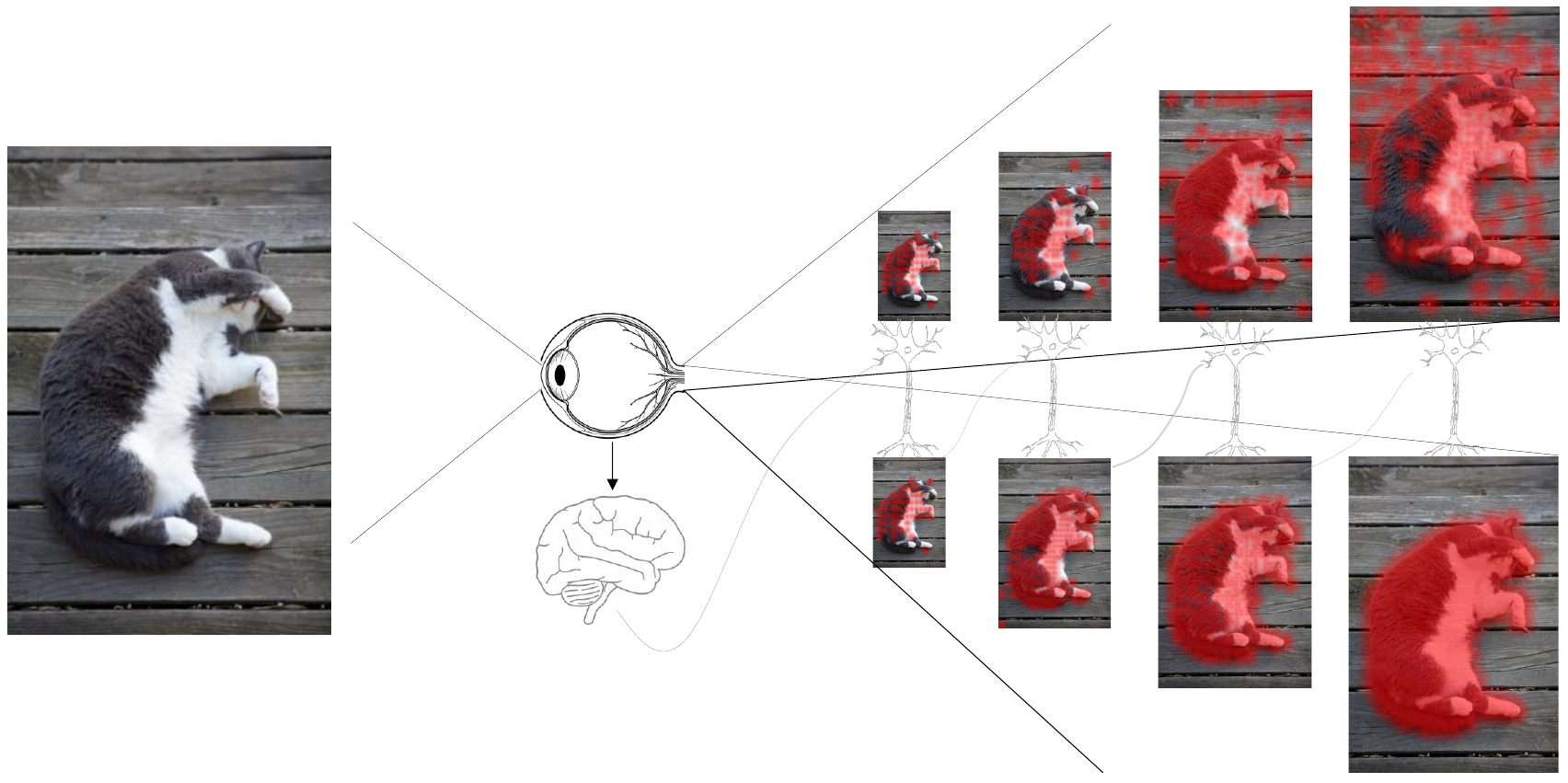}};
        \node[inner sep=1pt, font=\footnotesize] at (1.0,0.8) {Original Image};
        \node[inner sep=1pt, font=\footnotesize] at (3.5,1.0) {Human Brain};
        \node[inner sep=1pt, font=\footnotesize] at (3.5,0.6) {(Query Learning)};
        \node[inner sep=1pt, font=\footnotesize] at (3.5,3.6) {Human Eyes};
        \node[inner sep=1pt, font=\footnotesize] at (6.6,5.2) {Magnocellular Pathway (First Gaze)};
        \node[inner sep=1pt, font=\footnotesize] at (6.5,0.2) {Parvocellular Pathway (Second Gaze)};
    \end{scope}
    
  \end{tikzpicture}
  \caption{Biological inspiration from human visual perception. Human vision operates through two complementary pathways: the magnocellular pathway rapidly processes coarse spatial information and global scene structure, and the parvocellular pathway refines fine spatial details and precise object boundaries, with cortical attention modulation in the brain dynamically coordinating information flow between these streams.}
  \label{fig1}
  \vspace{-0.5em}
\end{figure}
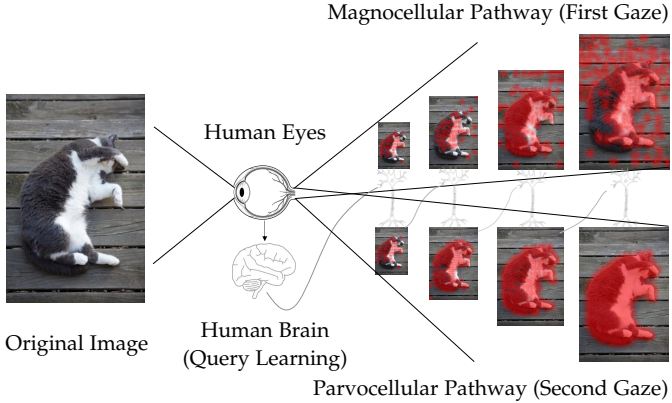

Recent SOD methods pursue performance improvements through four directions, including semantic enhancement \cite{liu2021tritransnet,10227346}, boundary refinement \cite{8954094,10155248}, auxiliary task supervision \cite{yuan2024fgnet,zhu2024separate}, and multi-modal fusion \cite{10475351,10778650}. However, in pursuit of continuous performance improvements, each direction has evolved toward increasingly sophisticated architectures involving multi-stage processing pipelines, specialized fusion modules, and elaborate attention mechanisms \cite{10268450,10184101}. Some existing methods even assume and empirically validate that achieving the optimal SOD performance further requires combining these directions within unified architectural frameworks \cite{9320524,10227346}. Such integration further increases architectural complexity and introduces potential conflicts between different enhancement strategies. These conflicts may create feature redundancy that wastes computational resources and creates cross-component interference that obscures important saliency cues, hindering the realization of optimal performance gains from increased architectural sophistication, as shown in Fig. \ref{fig2}. Actually, SOD fundamentally aims to emulate human visual attention mechanisms that identify prominent objects in complex scenes \cite{730558}. In contrast to the efficient and adaptive processing of human vision, current SOD methods employ increasingly complex architectures with a limited biological basis. This naturally raises questions: \textbf{can biological fidelity alone deliver competitive performance without the architectural complexity prevalent in current methods?} \textbf{Which biomimetic principles of human vision can be implemented as simple, trainable modules for SOD? How can we design a unified framework that embodies these principles with interpretability?}  

To address these questions, we examine the biological foundations of human visual processing, which reveal two complementary mechanisms that enable effective visual processing. First, the human visual system maintains robust object recognition when local visual information is missing. V2 neurons in the human brain respond to illusory contours with similar intensity to real edges \cite{vonderHeydt1984Science}, and higher cortical areas exhibit enhanced activation when fragmented elements trigger object completion \cite{Kovacs1995JNeuro,Lerner2002CerebCortex}. This completion mechanism enables \textbf{coherent feature extraction from complex visual scenes where salient objects may be partially obscured or embedded within cluttered backgrounds}. Second, visual processing operates through two complementary pathways \cite{atkinson1992early,han2022diversity}: a magnocellular stream that captures global scene structure and coarse spatial relationships, and a parvocellular stream that processes fine spatial details and precise object boundaries, with cortical attention modulation dynamically coordinating information flow between these two streams shown in Fig. \ref{fig1}. The coarse pathway with cortical attention modulation's ability to narrow the search space to candidate regions, combined with the fine pathway's capacity for pixel-accurate boundary recovery within identified areas, \textbf{minimizes wasteful processing of irrelevant background regions and consequently reduces feature interference from non-salient areas}. Inspired by these dual biological principles of robust representation learning and dual-pathway processing, we propose DualGazeNet, a biologically motivated pure Transformer framework for efficient and accurate RGB-SOD.


DualGazeNet operationalizes these dual biological principles through four synergistic components. First, the hierarchical feature extraction module (HFEM) models the magnocellular pathway's global processing ability. it also implements robust representation learning by employing a hierarchical vision Transformer backbone with lightweight parameter-efficient adapters. This backbone is pretrained through masked autoencoding on partial information, enabling robust feature extraction from complete images during inference. Second, the multi-scale gaze-localization query module (MGQM) models the attentional modulation in cortex by introducing learnable universal query tokens. These tokens progressively interact with hierarchical features through cross-attention mechanisms to aggregate semantic cues and identify salient regions. Third, the multi-scale gaze-guided feature reconstruction module (MGFRM) mirrors the parvocellular pathway's detailed processing by leveraging these enriched query tokens to guide fine-grained spatial feature reconstruction through reverse cross-attention, progressively refining multi-scale representations with emphasis on identified salient regions. Finally, the saliency mask generation module (SMGM) integrates information from both processing streams, projecting the reconstructed features into pixel-level saliency predictions.

DualGazeNet follows a minimalist design philosophy. The main contributions are summarized as follows: i) DualGazeNet first adopts and integrates dual biological principles within a single Transformer stream, avoiding redundant multi-branch designs in prior models and mitigating interference from irrelevant features, thus enhancing both efficiency and precision. ii) We first introduce a dual-gaze processing paradigm that models the magnocellular–parvocellular pathways with cortical attention modulation via cross-level query-guided coupling. The learnable query tokens refine hierarchical semantics and reconstruct fine-grained spatial details, thereby adaptively allocating computational focus on identified salient regions for improved accuracy and efficient resource utilization. iii) We develop an efficient mask generation module that fuses dual-pathway outputs through element-wise similarity estimation. This design features a concise fusion process that preserves pixel-level accuracy and avoids much computational cost. iv) Extensive experiments across multiple benchmark datasets demonstrate that our framework achieves state-of-the-art performance and strong generalization. Across transformer-based SOD baselines of similar capacity, DualGazeNet improves average inference throughput by 60$\%$ and reduces computational cost by 53.4$\%$ (FLOPs), all while maintaining better SOD accuracy.

\section{Related Works}
\subsection{CNN-based Salient Object Detection}\label{Vision Tasks in Autonomous Driving}

As neural networks demonstrated superior feature extraction capabilities, traditional handcrafted methods \cite{ecssd,6871397,730558,6247743,7410522,6243147} were gradually replaced by learning-based approaches for SOD. A key breakthrough came with fully convolutional networks (FCNs) \cite{long2015fully}, which substituted fully connected layers with convolutional layers to enable pixel-wise predictions instead of classification scores \cite{9320524}. 

Existing CNN-based SOD methods pursue different strategies to enhance detection performance. Some methods prioritize the extraction, utilization, and fusion of multi-level features to capture comprehensive semantic and spatial information \cite{minet,menet,9756227,fang2022lc3net,9989433,liu2019simple}. For instance, Wu et al. \cite{9756227} introduced an extremely downsampled block that gradually downsamples feature maps to learn a global view of the whole image, demonstrating that enhancing high-level feature learning is essential for accurate SOD. Fang et al. \cite{fang2022lc3net} proposed a ladder context correlation network that enhances multi-level feature fusion through specialized convolution blocks and cross-level aggregation modules. Similarly, Wu et al. \cite{9989433} designed a "U-Net in U-Net" framework that embeds a smaller U-Net into a larger U-Net backbone to enable multi-level and multi-scale representation learning without relying on classification backbones. Another approach in \cite{liu2019simple}  further proposed PoolNet, which expanded the role of pooling techniques within pyramid-based architectures through a global guidance module and feature aggregation module to capture global context and reduce upsampling aliasing effects for real-time SOD. Beyond multi-level feature engineering, some methods leverage boundary cues or other multi-task learning frameworks to refine object boundaries and improve localization accuracy \cite{dfi,wei2020f3net,9008371,10155248,yuan2024fgnet}. In \cite{wei2020f3net}, the authors introduced F$^{3}$Net that employs a pixel position-aware loss that emphasizes boundary and hard pixels, yielding sharper object contours. In \cite{9008371}, EGNet was proposed to explicitly model the complementarity between salient edge and object features within a single network, leveraging edge supervision to refine boundaries and enhance localization. More recently, Zhang et al. \cite{10155248} proposed ELSA-Net, which embeds edge cues into saliency features through edge-guided learning and specific aggregation, thereby enhancing boundaries. Yuan et al. \cite{yuan2024fgnet} proposed FGNet, a fixation-guided multi-branch framework where fixation cues enhance saliency and edge prediction, and a multi-resolution interaction module improves localization and boundary clarity in complex scenes.

However, these approaches typically rely on complex multi-component architectures \cite{dcnet} that may suffer from feature interference and increased computational overhead. Additionally, methods that depend on boundary information face inherent limitations, as demonstrated in \cite{zhou2025rmfdnet}, where boundary-dependent approaches can degrade SOD performance due to overly cluttered boundaries \cite{wang2021feature}, inaccurate edge prediction \cite{liu2022generative}, and improperly weighted edge loss functions \cite{zhao2024combining}. In this paper, the proposed DualGazeNet unifies the detection pipeline into a streamlined architecture. Rather than combining multiple specialized components for feature fusion and boundary refinement, our method achieves state-of-the-art performance through a simple dual-pathway design that naturally captures both semantic and spatial information without relying on edge cues or complex fusion strategies.

\subsection{Transformer-based Salient Object Detection}\label{Autonomous Driving Datasets}
Transformers, originally designed for sequence transduction tasks \cite{vaswani2017attention}, have fundamentally transformed various domains due to their self-attention mechanism that effectively captures long-range dependencies and global contextual relationships than CNN. The computer vision community has successfully adapted this architecture by reformulating image processing as sequence modeling, leading to the development of Vision Transformer (ViT) \cite{Dosovitskiy2021AnII}. ViT partitions input images into non-overlapping patches, applies linear embeddings, and processes the resulting token sequences through standard Transformer encoders. This patch-based paradigm has demonstrated exceptional performance across diverse computer vision tasks, including pose estimation \cite{mao2022poseur}, object tracking \cite{9964258}, object detection \cite{9577946}, image classification \cite{Dosovitskiy2021AnII}, and dense prediction \cite{zheng2021rethinking,10377177}, establishing ViT as a compelling alternative to convolutional architectures.


The ViT family has evolved into diverse architectural variants \cite{9710747,9710031,9711179} that address the trade-offs between local inductive biases, global modeling capacity, and computational efficiency.  Among these architectural variants, several have demonstrated successful applications in SOD tasks \cite{Liu_2021_ICCV,NEURIPS2021_82898892,WU2024105039,10834569}. For example, Liu et al. \cite{Liu_2021_ICCV} adopted a T2T-ViT-based encoder-decoder architecture that maintains a pure transformer design, utilizing task-related tokens and reverse tokens-to-token upsampling to capture global dependencies and preserve sharp boundaries. Zhang et al. \cite{NEURIPS2021_82898892} also followed a pure transformer design but augmented it with an energy-driven latent prior that is optimized through Langevin dynamics, yielding uncertainty-aware saliency maps robust to ambiguity. To further enhance performance, recent approaches have pursued the integration of CNN components into transformer architectures, achieving performance gains at the cost of increased structural complexity. For instance, Wu et al. \cite{WU2024105039} combine a Swin encoder with convolutional recalibration and fusion in a progressive two-stage decoder, achieving precise boundary refinement and robust multi-scale integration. Additionally, Sun et al. \cite{10834569} introduced a diffusion-based framework that integrates PVT transformer conditioning with CNN denoising components, further demonstrating the effectiveness of hybrid designs. 

However, these aforementioned transformer-based SOD approaches still pursue architectural mixing and rely on exhaustive feature extraction paradigms. They process complete image representations through hierarchical feature pyramids and dense attention mechanisms to enhance performance. In contrast, our approach maintains a pure transformer architecture and embraces a minimalist design philosophy, demonstrating that pure transformer frameworks can achieve superior performance even compared to existing hybrid CNN-Transformer approaches in SOD tasks through biologically-inspired visual processing.

\subsection{Feature Redundancy Compensation, Missing Feature Recovery, and Lightweight Strategies}\label{Feature Redundancy Compensation}
The complex multi-component designs in some SOD architectures inherently lead to feature redundancy and key information loss issues. Recognizing these challenges, several works have proposed specialized modules for feature reduction and compensation \cite{deng2018r3net,zhou2025rmfdnet,Hu2024IJCV_CPNet}. For instance, Deng et al. \cite{deng2018r3net} introduced residual refinement blocks that recurrently integrate complementary multi-level features. Similarly, Zhou et al. propose RMFDNet \cite{zhou2025rmfdnet} utilizing a decoupling strategy for feature optimization. This method separates redundant and missing features through specialized complement and removal decoders, improving SOD accuracy. Beyond these explicit feature processing designs, attention mechanisms have emerged as another effective strategy. Attention provides learnable weights that can dynamically prioritize important features \cite{10516304,10155248,cagnet,tracer,10497889}. For example, pixel-wise contextual attention selectively aggregates long-range cues around each location to highlight salient regions and suppress background clutter \cite{10155248}.  Channel-spatial attention mechanisms enhance feature representation by weighting both spatial locations and feature channels based on their relevance to salient regions \cite{10497889}. However, these explicit feature processing designs continue to increase architectural complexity through additional specialized components, and they also lack interpretability, as the learning outcomes of each module cannot be visualized to verify the effective identification of redundant and missing feature information. Additionally, attention mechanisms depend on careful or complex loss function design to achieve proper weight allocation, but more attention utilization requires more computational resources. 

\begin{figure}[!t]
\vspace{-2pt}
  \centering
  \begin{tikzpicture}
    \begin{scope}[xshift=-2.7cm, yshift=0cm ]
        \node[above right] (fig63) at (0,0){\includegraphics[ height=0.265\linewidth]{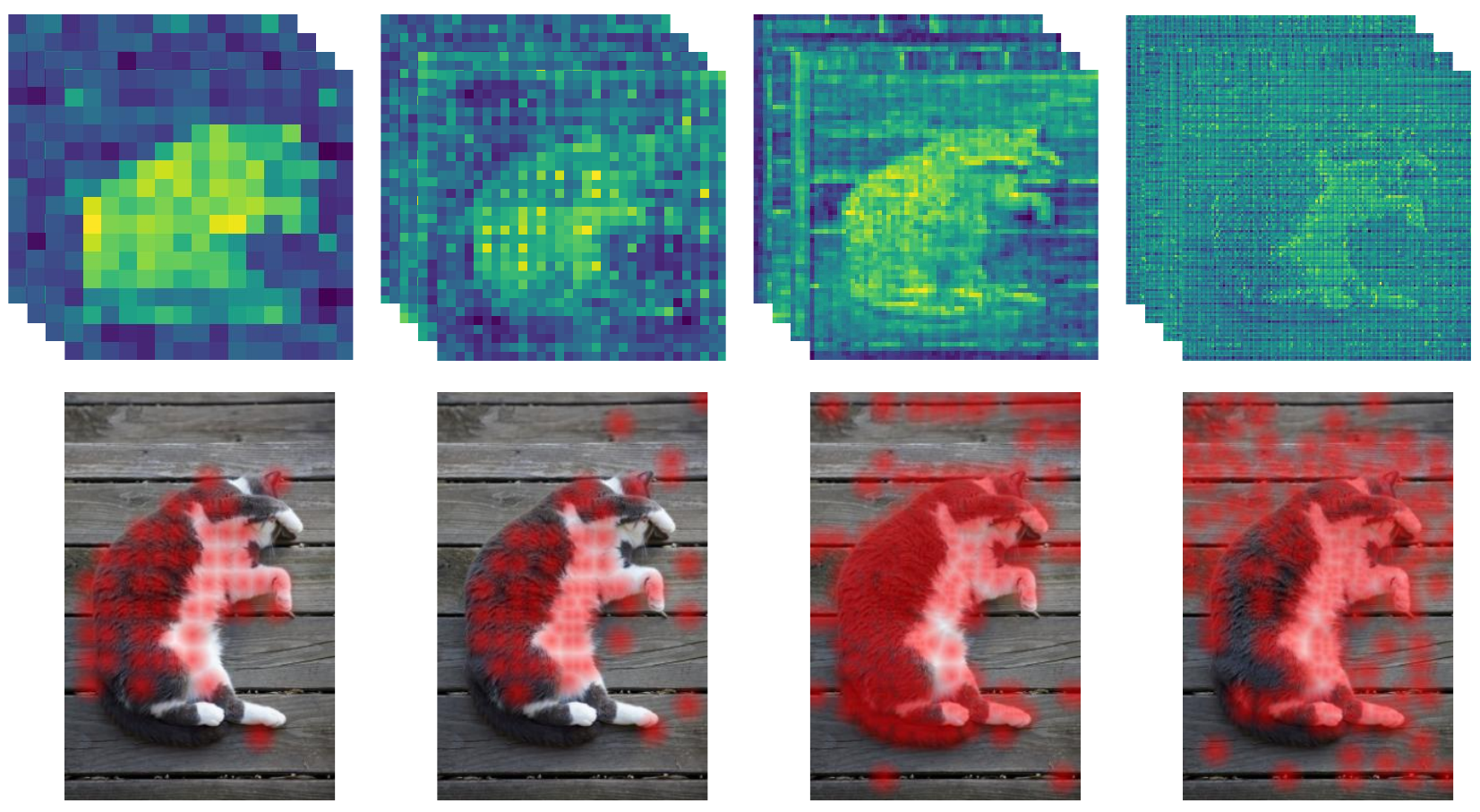}};
        \node at ($(fig63.south)+(0,-0.35)$) { \small (a) first gaze results};
    \end{scope}
    \begin{scope}[xshift=1.7cm, yshift=-0.01cm]
        \node[above right] (fig64) at (0,0){\includegraphics[ height=0.27\linewidth]{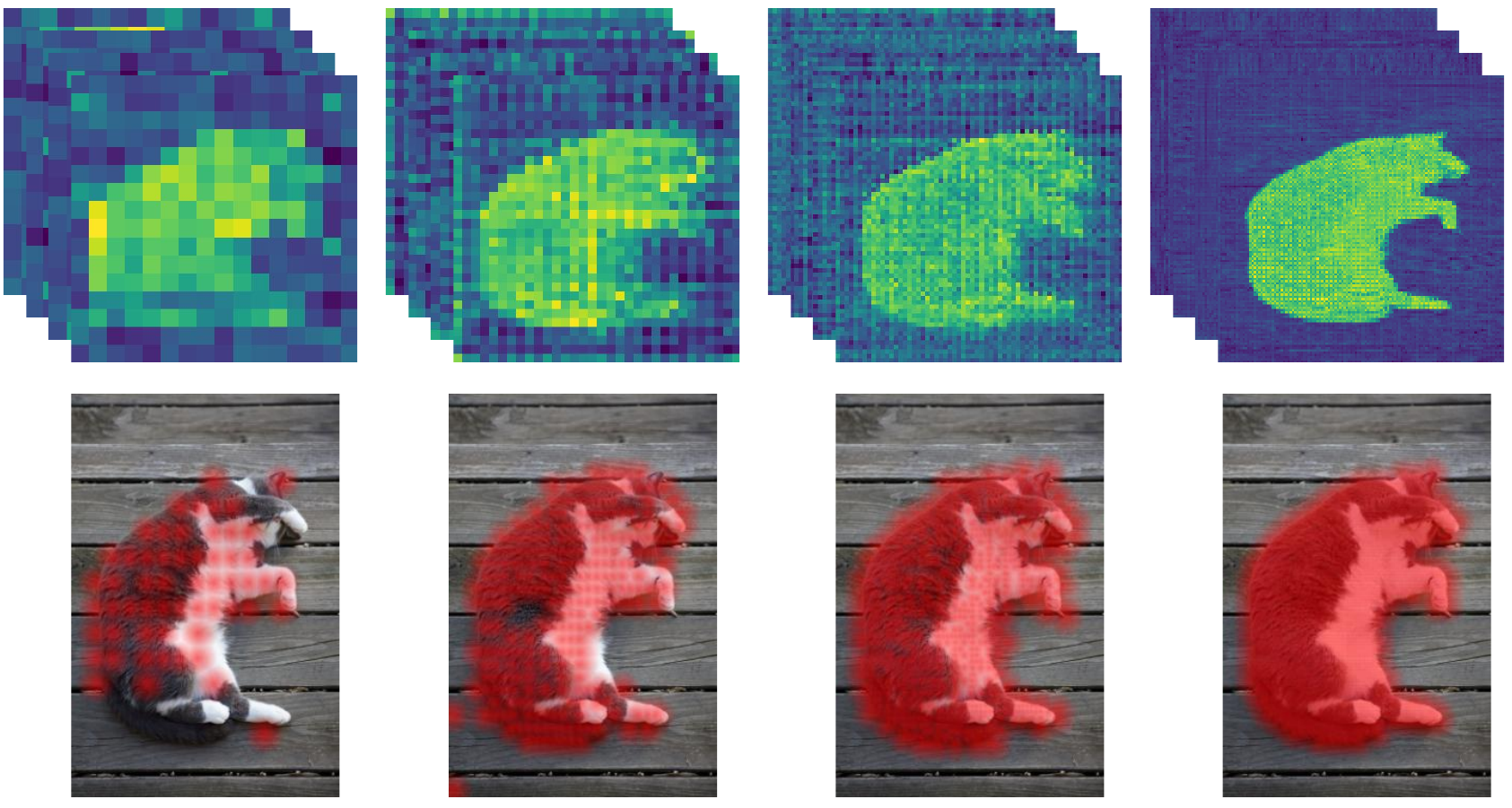}};
        \node at ($(fig64.south)+(0,-0.35)$) { \small (b) second gaze results};
    \end{scope}
  \end{tikzpicture}
  
  \caption{Attention map visualization of the DualGazeNet. (a) First gaze generates dispersed attention patterns to capture global scene context. (b) Second gaze, guided by query tokens, refines attention to focus on salient object regions. This visualization demonstrates how our method progressively refines attention from global scene understanding to salient object localization.}
  \label{fig3}
  \vspace{-0.5em}
\end{figure}

To address these computational and architectural complexity issues, lightweight design strategies have gained increasing attention in SOD research. These methods include compact architectural design \cite{9669083,11072100}, network pruning \cite{10330640,ma2023llm,10428086}, knowledge distillation \cite{gou2021knowledge}, and quantization \cite{xiao2023smoothquant,li2023q}. However, these lightweight approaches primarily focus on achieving competitive performance with significantly reduced computational cost rather than pursuing maximum accuracy. In contrast, by mimicking how humans naturally process visual scenes through coordinated coarse and fine attention, DualGazeNet achieves improved efficiency-accuracy trade-offs. This approach reduces computational redundancy by selectively focusing on salient regions and suppressing irrelevant visual information. The method also does not require complex loss function design, which simplifies optimization compared to existing frameworks. Furthermore, each gaze component provides interpretable attention visualizations shown in Fig. \ref{fig3}, enabling a better understanding of the model's decision-making process.


\subsection{Query-Based Learning in Computer Vision}\label{Query-Based Learning in Computer Vision}
The success of DETR architecture \cite{detr} has demonstrated the effectiveness of explicit learnable queries for guiding models toward specific objects in a direct and interpretable manner. This breakthrough led to the widespread adoption of query-based learning structures across diverse computer vision tasks \cite{10657220,10203147,Jin2025KeypointDETR,seqrank}.  

Recently, query-based learning structures have begun to be explored in the SOD domain \cite{Nam2025_SA_DETR,Cen2025_PDDNet,Yang2024_SEF_E2E}.  SA-DETR \cite{Nam2025_SA_DETR} extended DETR's cross-attention mechanism to learn relationships between global image semantics and salient objects, but it focuses on object-level saliency rather than pixel-wise accuracy. PDNet \cite{Cen2025_PDDNet} has been proposed in a parallel dual-decoder architecture, where learnable queries establish global saliency dependencies and maintain local spatial features through CNN-transformer integration. Moreover, SEFNet \cite{Yang2024_SEF_E2E} employed set prediction methods with query vectors to achieve end-to-end fusion between saliency and edge features through Hungarian matching algorithms. However, these approaches still rely on complex multi-component architectures that combine CNN-Transformer hybrids, auxiliary task supervision, or specialized fusion modules, inheriting the architectural complexity and feature interference issues prevalent in existing SOD methods. Furthermore, they also process full spatial coverage through dense attention mechanisms rather than adopting selective processing strategies, resulting in computational redundancy when handling irrelevant regions.

In contrast, DualGazeNet redefines query-based SOD through biologically-inspired dual-gaze mechanisms. Unlike existing methods that employ queries within complex CNN-Transformer hybrid designs, auxiliary task supervision, or specialized fusion modules, our approach introduces learnable universal query tokens that directly model human visual attention pathways. These queries enable selective processing by focusing computational resources on real salient regions rather than full spatial coverage.

\section{Methods}
\subsection{Method Overview} \label{Platform and Sensors}

\begin{figure*}[t]
\centering
\includegraphics[width=0.99\textwidth]{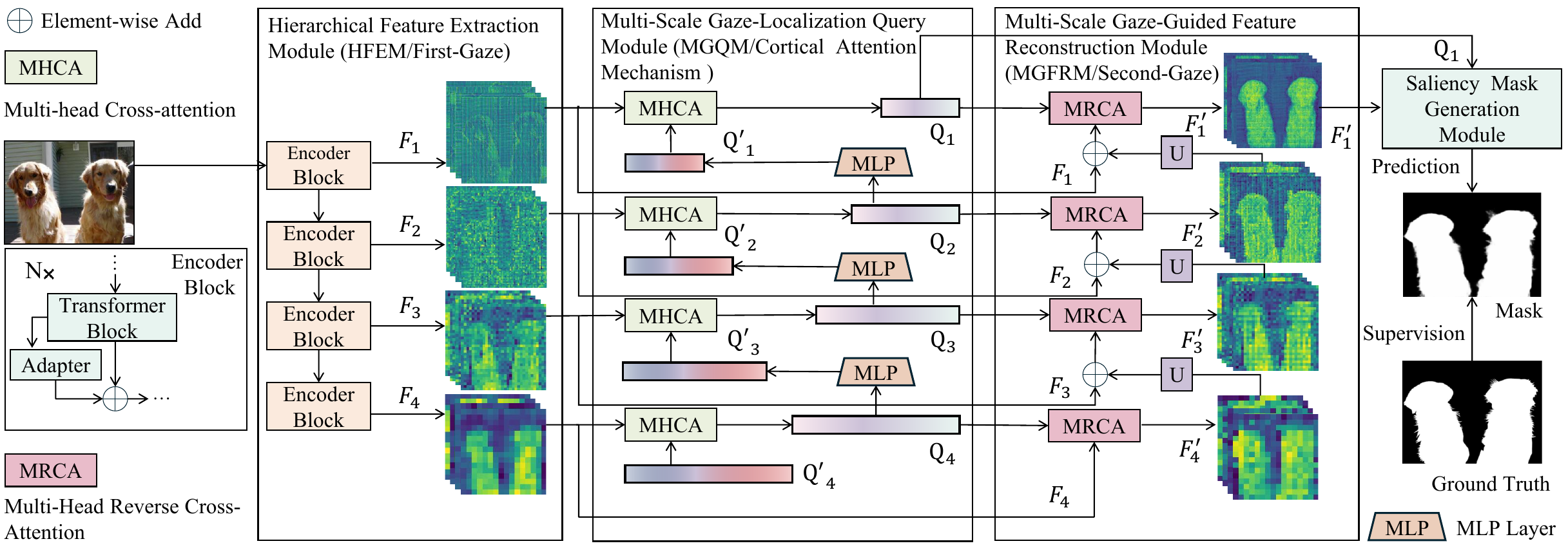}
\caption{Architecture of DualGazeNet. Inspired by the dual pathways in human vision illustrated in Fig. \ref{fig1}, DualGazeNet implements a two-stage processing mechanism: the first ``gaze'' extracts coarse global structure (magnocellular-like pathway), and the second ``gaze'' performs fine spatial refinement (parvocellular-like pathway). Query learning mimics cortical attention modulation by selectively enhancing task-relevant features and suppressing background interference. The architecture comprises four key components: (1) HFEM generates multi-level features $F_1$--$F_4$; (2) MGQM forms pseudo-queries $Q_1^{'}$--$Q_4^{'}$ that interact with multi-level features $F_1$--$F_4$ via MHCA to yield cortical queries $Q_1$--$Q_4$; (3) MGRFM refines features $F_1$--$F_4$ with cortical queries $Q_1$--$Q_4$ and obtain enhanced hierarchical features $F^{'}_1$--$F^{'}_4$ through MRCA; (4) SMGM outputs the final prediction.}
\label{architecture}
\vspace{-0.5em}
\end{figure*}

\textbf{Overall Pipeline:}. As illustrated in Fig.~\ref{architecture}, the DualGazeNet realizes the aforementioned biological principles through four synergistic components:

\textit{1) Hierarchical Feature Extraction Module (First-Gaze):} Given an input image $I \in \mathbb{R}^{H \times W \times 3}$, the hierarchical feature extraction module (HFEM) emulates the magnocellular pathway's rapid global processing by employing a frozen hierarchical Vision Transformer backbone pretrained through masked autoencoding. During pretraining, this backbone learns from randomly masked image regions, achieving robust representation learning that generalizes to complete images. During inference, HFEM processes the complete input image to extract multi-scale feature representations $\{F_i \in \mathbb{R}^{N_i \times D_i}\}_{i=1}^4$ at different hierarchical levels, where $N_{i}$ represents the spatial token count and $D_i$ denotes the embedding dimension at level $i$. Lightweight adapters are incorporated for task-specific adaptation. 

\textit{2) Multi-Scale Gaze-Localization Query Module (Cortical Attention Mechanism):} The multi-scale gaze-localization query module (MGQM) realizes the cortical attention mechanism by introducing pseudo-queries tokens $\{Q_{i}^{'} \in \mathbb{R}^{1 \times D_i}\}_{i=1}^4$ that progressively interact with hierarchical features $\{F_i \in \mathbb{R}^{N_i \times D_i}\}_{i=1}^4$ through cross-attention mechanisms to obtain cortical queries $\{Q_{i} \in \mathbb{R}^{1 \times D_i}\}_{i=1}^4$. This progressive interaction aggregates semantic cues across multiple scales and identifies abstract spatial locations of salient regions through top-to-down attention propagation.

\textit{3) Multi-Scale Gaze-Guided Feature Reconstruction Module (Second-Gaze):}
The multi-scale gaze-guided feature reconstruction module (MGFRM) mirrors the parvocellular pathway’s detail-oriented processing and is guided by cortical queries $\{Q_i \in \mathbb{R}^{1 \times D_i}\}_{i=1}^4$ produced by MGQM.
MGFRM employs reverse cross-attention to progressively reconstruct refined multi-scale representations $\{F_i^{'} \in \mathbb{R}^{N_i' \times D_i'}\}_{i=1}^4$ with an emphasis on the identified salient regions.
This query-guided reconstruction enables the recovery of spatial detail within salient regions while suppressing irrelevant background responses.

\textit{4) Saliency Mask Generation Module:} The saliency mask generation module (SMGM) integrates information from aforementioned stages, projecting the reconstructed features into pixel-level saliency predictions $S \in \mathbb{R}^{H \times W}$.

Next, we present the detailed technical design of each component in the DualGazeNet architecture.

\subsection{Hierarchical Feature Extraction Module} \label{Object Detection and Tracking}
We employ the Hierarchical Vision Transformer (Hiera) encoder~\cite{hiera} as the backbone and the first gaze to extract hierarchical feature representations. During pretraining, the model learns to reconstruct complete representations from randomly masked regions, developing coherent feature extraction even when substantial portions of the input are obscured. This mirrors the human visual system's ability to maintain robust object recognition when local visual information is missing, enabling processing of complex visual scenes where salient objects may be partially obscured or embedded within cluttered backgrounds. 

\begin{figure}[!t]
\vspace{-3pt}
  \centering
  \begin{tikzpicture}%
    \begin{scope}[xshift=0cm, yshift=0cm ]
        \node[above right] (fig63) at (0.0,0.4){\hspace{-0.1cm}\includegraphics[ height=0.428\linewidth]{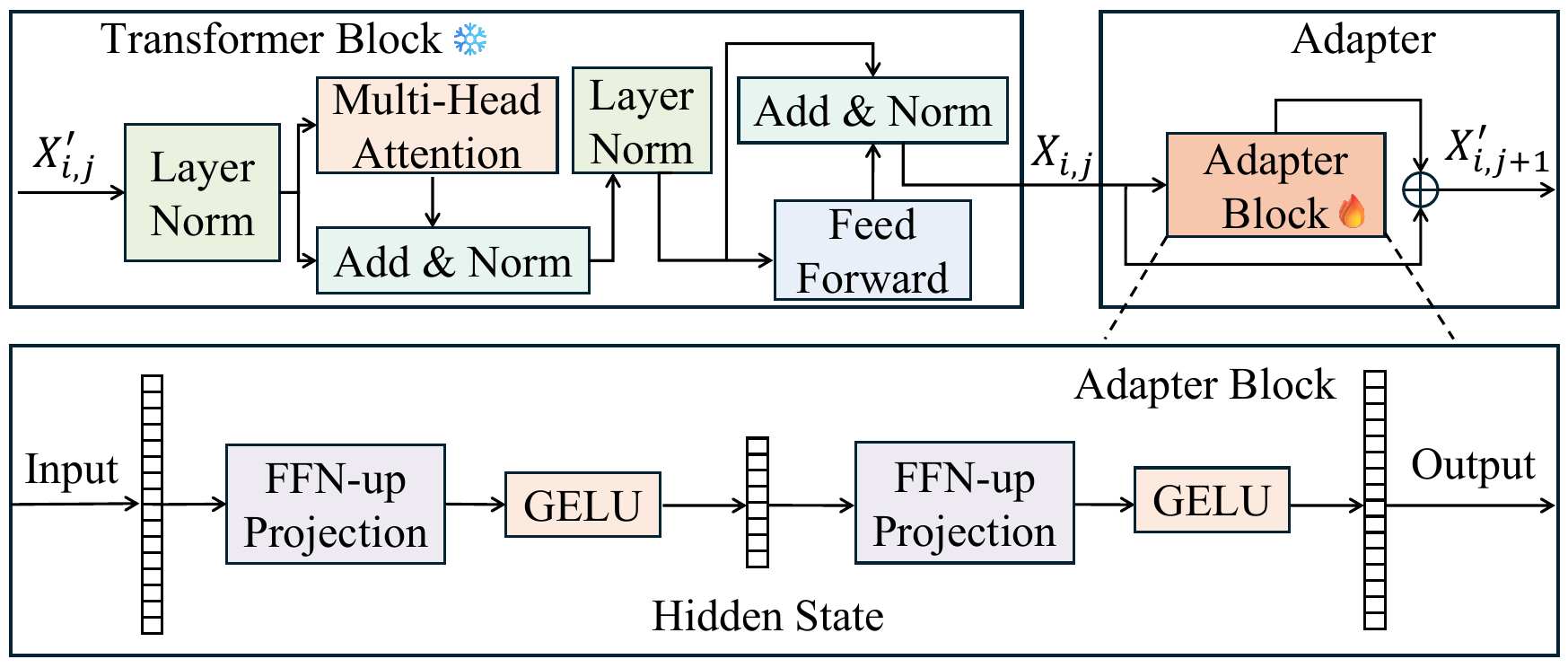}};
    \end{scope}
    
  \end{tikzpicture}
  \caption{Illustration of the proposed adapter-enhanced Transformer block within the $i$-th stage.}
  \label{fig3t}
\end{figure}

Hiera's minimalist design philosophy aligns well with our DualGazeNet framework. It eliminates all specialized components such as convolutions, shifted windows, and relative position embeddings, demonstrating that pure Transformers can effectively learn spatial relationships without additional inductive biases. However, since Hiera was originally designed for general vision tasks rather than SOD, task-specific adaptation is necessary. To preserve the pretrained representations while enabling domain adaptation, we insert lightweight adapter modules after each Transformer block within every hierarchical stage, as illustrated in Fig. \ref{fig3t}. The backbone encoder remains frozen, and each adapter follows a bottleneck architecture:
\begin{equation}
    X_{i,j+1}^{'} = X_{i,j} + \text{MLP}(X_{i,j})
\end{equation}
where $X_{i,j} \in \mathbb{R}^{N_{i,j} \times D_{i,j}}$ represents the feature output from the $j$-th Transformer block at the $i$-th hierarchical stage, with $N_{i,j}$ and $D_{i,j}$  denoting the number of spatial tokens and embedding dimension, respectively. The function $\operatorname{MLP}(\cdot)$ is a lightweight multi-layer perceptron that generates task-specific adaptations. In particular, it is defined as:
\begin{equation} \label{MLP}
\operatorname{MLP}(X_{i,j})=\sigma((\sigma(X_{i,j} W_1^{i,j}+b_1^{i,j})) W_2^{i,j}+b_2^{i,j})
\end{equation}
where $W_1^{i,j} \in \mathbb{R}^{D_{i,j}\times d}$, $W_2^{i,j} \in \mathbb{R}^{d \times D_{i,j}}$, and $\sigma(\cdot)$ denoting the $\operatorname{GELU}$ activation function. Here, $d \ll D_{i,j}$ is a bottleneck dimension that limits the number of additional parameters. Through this parameter-efficient adaptation strategy, HFEM produces hierarchical features $\{F_{i}\}_{i=1}^{4}$ that serve as the foundation for the subsequent modules.

\begin{figure}[!t]
  \centering
  \begin{tikzpicture}%
    \begin{scope}[xshift=0cm, yshift=0cm ]
        \node[above right] (fig63) at (0.0,0.4){\hspace{-0.1cm}\includegraphics[ height=0.69\linewidth]{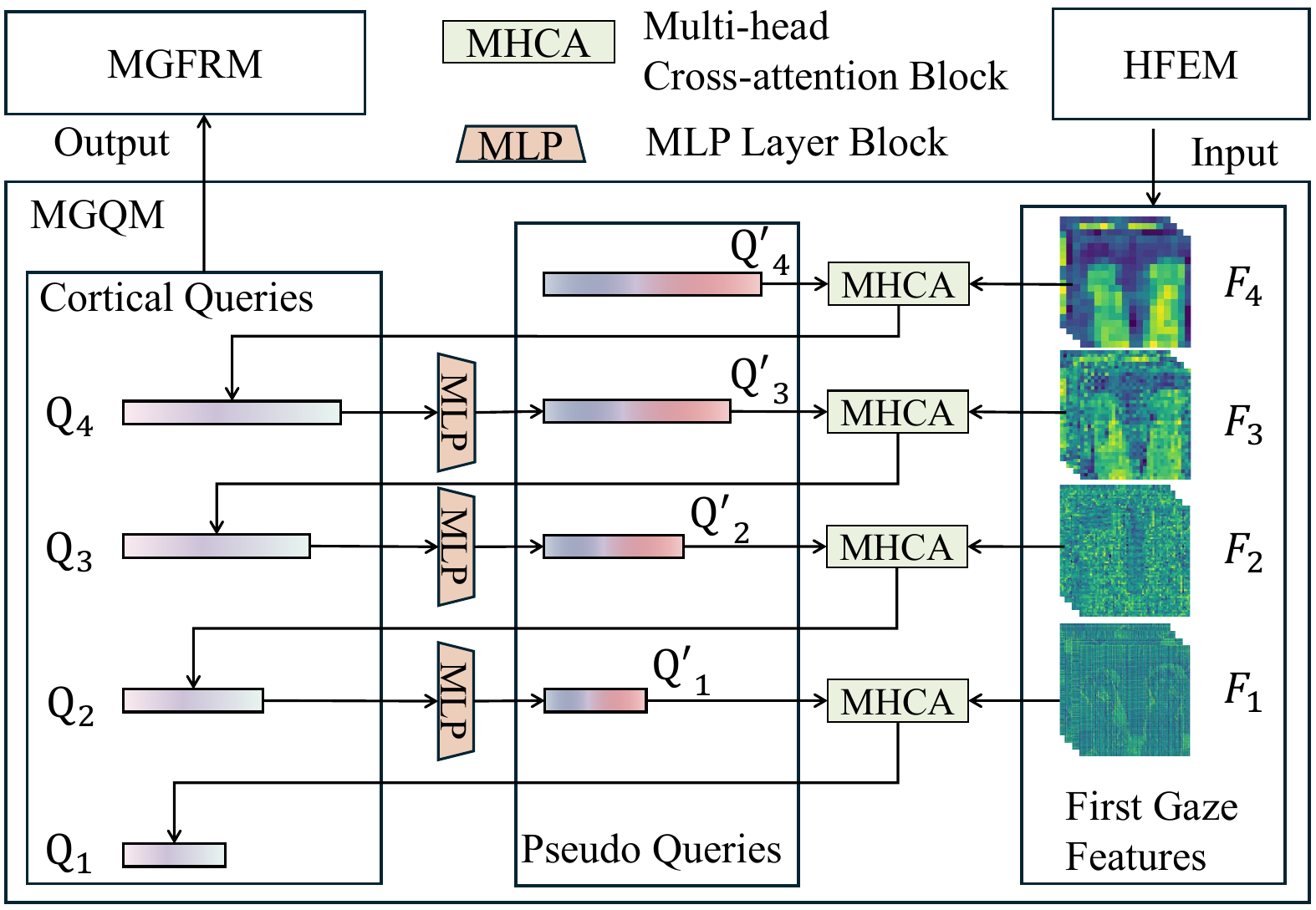}};
    \end{scope}
    
  \end{tikzpicture}
  \caption{Detailed pipeline of MGQM. This stage progressively refines cortical query representations across hierarchical levels through cross-attention mechanisms, enabling top-to-down semantic propagation from global scene understanding to scale-specific saliency localization.}
  \label{Fig.MGQM}
  \vspace{-0.5em}
\end{figure}

\subsection{Multi-Scale Gaze-Localization Query Module}
In this subsection, we model the cortical attention mechanism to establish scene understanding and identify regions of interest, as illustrated in Fig.~\ref{Fig.MGQM}. Specifically, given the hierarchical features $\{F_{i}\in \mathbb{R}^{N_i \times D_i}\}^{4}_{i=1}$ extracted by HFEM, the query localization process begins with an initial learnable pseudo-query token $Q_{\text{init}} \in \mathbb{R}^{1 \times D_{4}}$ at the deepest semantic level. To enable cross-scale information propagation, the pseudo-query $Q^{'}_{i}$ of each stage is recursively obtained via dimensional adaptation as follows:
\begin{equation} \label{queryy}
Q_i^{'}= \begin{cases}Q_{\text{init}}, & i=4 \\
\text{MLP}_i(Q_{i+1}), & 1 \leq i \leq 3\end{cases}
\end{equation}
where $\operatorname{MLP}_{i}(\cdot): \mathbb{R}^{D_{i+1}} \rightarrow \mathbb{R}^{D_i}$ performs dimensional adaptation across hierarchical stages. This design ensures consistent embedding spaces across different scales and enables seamless bottom-up information flow.

The cortical query token $Q_{i}$ is derived through a multi-head cross-attention (MHCA) mechanism between each pseudo-query $Q_{i}^{'}$ and hierarchical features $F_{i}$.
\begin{equation} \label{MHCA}
Q_i = \text{MHCA}(Q^{'}_i, F_i)
\end{equation}

In MHCA, the pseudo-query $Q_{i}^{'}$ and feature $F_{i}$  are projected into multi-head representations. For the $h$-th attention head $(h=1,\ldots,h_{i})$, the query, key, and value are computed as $Q_i^{(h)}  =Q_i^{'} W_i^{(Q, h)} \in \mathbb{R}^{1 \times d_{h, i}}$, $K_i^{(h)} =F_i W_i^{(K, h)} \in \mathbb{R}^{N_i \times d_{h, i}}$, and $V_i^{(h)}  =F_i W_i^{(V, h)} \in \mathbb{R}^{N_i \times d_{h, i}}$, where $d_{h,i} = D_{i}/h_{i}$ and $W_i^{(Q, h)}$, $W_i^{(K, h)}$, $W_i^{(V, h)} \in \mathbb{R}^{D_{i} \times d_{h,i}}$. The outputs of all attention heads are obtained follow a standard scaled dot-product attention formulation, i.e., $O_i^{(h)} =A_i^{(h)} V_i^{(h)} \in \mathbb{R}^{1 \times d_{h, i}}$ and $A_i^{(h)} =\operatorname{Softmax}(\frac{Q_i^{(h)}(K_i^{(h)})^{\top}}{\sqrt{d_{h, i}}}) \in \mathbb{R}^{1 \times N_i}$. These outputs are then concatenated and linearly projected, i.e., $Q_i^{''}=\operatorname{Concat}(O_i^{(1)}, \ldots, O_i^{\left(h_i\right)}) W_i^{(O)}+b_i^{(O)} \in \mathbb{R}^{1\times D_{i}}$, and residual connections and layer normalization are applied to stabilize the training process:  
\begin{equation} \label{query}
\begin{aligned}
Q_i & = \text{LN}(Q_{i}^{'}+Q_i^{''}) \in \mathbb{R}^{1\times D_{i}}
\end{aligned}
\end{equation}
where $W_i^{(O)} \in \mathbb{R}^{D_i \times D_i}$ and $b_i^{(O)} \in \mathbb{R}^{D_i}$. 

\begin{figure}[!t]
\vspace{-0.9pt}
  \centering
  \begin{tikzpicture}%
    \begin{scope}[xshift=0cm, yshift=0cm ]
        \node[above right] (fig63) at (0.0,0.4){\hspace{-0.1cm}\includegraphics[ height=0.754\linewidth]{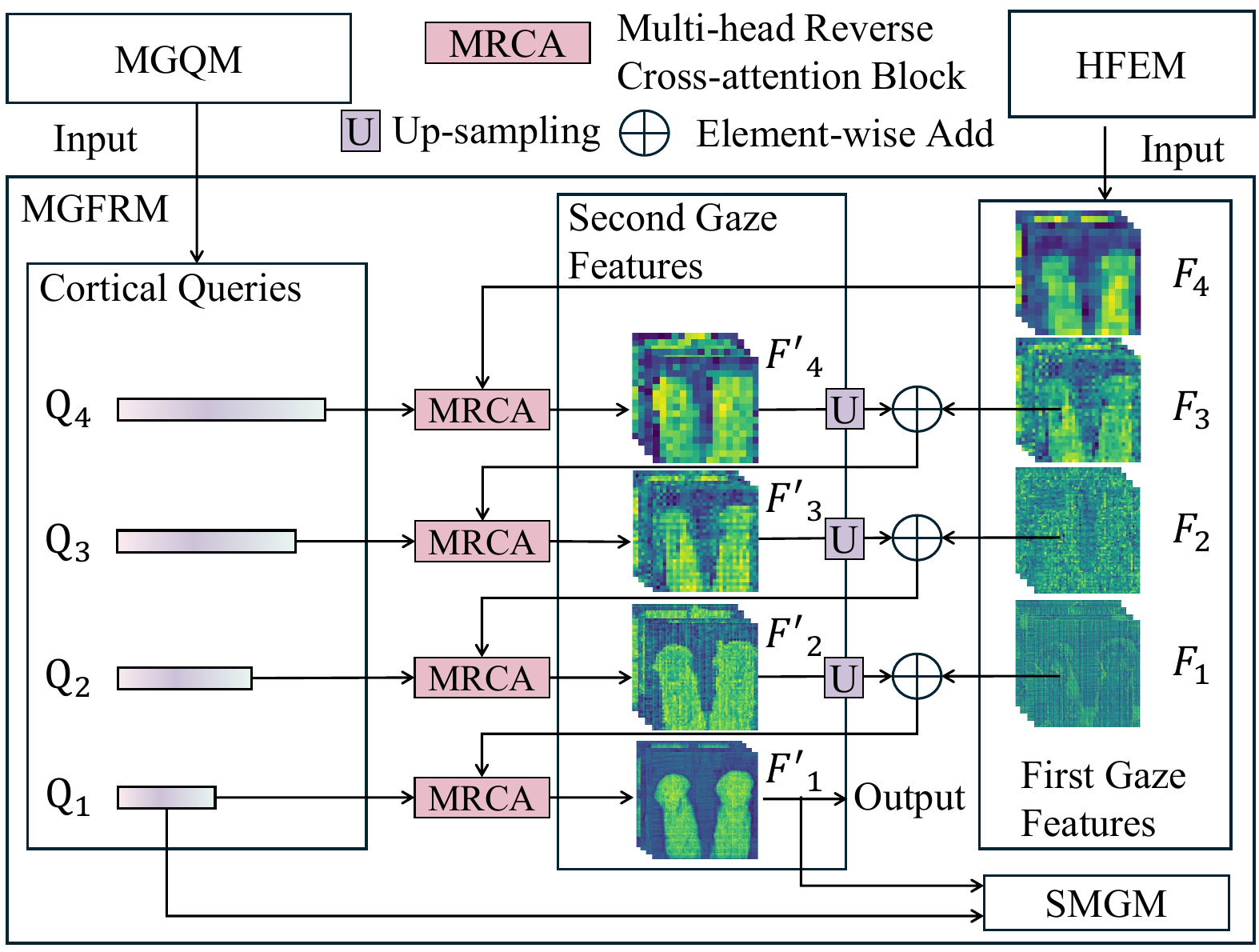}};
    \end{scope}
    
  \end{tikzpicture}
  \caption{Detailed pipeline of MGFRM. Distinct from conventional encoder–decoder fusion, MGFRM reuses query tokens as dynamic semantic probes to modulate multi-level features in a top-down manner, enabling adaptive reconstruction of spatial details under strong semantic guidance.}
  \label{Fig.MGFRM}
  \vspace{-0.5em}
\end{figure}

Through this hierarchical cross-attention mechanism, MGQM progressively refines the query representations at each scale, with each refined query token $Q_{i}$ encoding scale-specific semantic information about potential salient regions. This top-down semantic propagation allows deeper-level semantic knowledge to guide query refinement at shallower levels, thereby accumulating hierarchical saliency cues across multiple scales. The resulting enriched query tokens $\{Q_{i}\}_{i=1}^{4}$ serve as semantic anchors that encode saliency understanding at multiple scales, providing the necessary guidance for the next second gaze stage.

\subsection{Multi-Scale Gaze-Guided Feature Reconstruction Module}\label{sec:MGFRM}
MGQM generated cortical query tokens that encode hierarchical saliency information across multiple scales. While these semantic anchors can identify regions of interest through global scene understanding, they lack the spatial resolution required for pixel-wise saliency prediction. To bridge this semantic-spatial gap, we introduce the MGFRM, which constitutes the second gaze of our dual-gaze pipeline.

Distinct from conventional encoder–decoder architectures that rely on fixed fusion heuristics, MGFRM introduces a query-guided modulation strategy. The cortical query tokens $\{Q_{i}\}_{i=1}^{4}$ from MGQM act as dynamic semantic guides to modulate the hierarchical features $\{F_{i}\}_{i=1}^{4}$ from HFEM.  As illustrated in Fig.~\ref{Fig.MGFRM}, this query-guided reconstruction process produces enhanced hierarchical features $\{F_{i}^{'}\}_{i=1}^{4}$ through a top-down pathway as follows:
\begin{equation}
    F^{'}_{i} = 
    \begin{cases}
         \text{MRCA}(F_{i}, Q_{i}),& i = 4 \\
        \begin{aligned}
            &\text{MRCA}(\text{Up}_i(F^{'}_{i+1}) + {F}_{i}, {Q}_{i}), \\ 
        \end{aligned} & 1 \le i \le 3
    \end{cases}
\end{equation}
where $\text{Up}_i(\cdot): \mathbb{R}^{H_{i+1} \times W_{i+1} \times D_{i+1}} \rightarrow \mathbb{R}^{H_i \times W_i \times D_i}$ performs upsampling for dimensional adaptation. The multi-head reverse cross-attention (MRCA) mechanism reverses the role assignment from MHCA: hierarchical feature tokens serve as queries, and cortical query tokens act as keys and values. 

Specifically, in MRCA, the input features $\tilde{F}_{i}$ are constructed hierarchically. At the deepest stage $i = 4$, we set  $\tilde{F}_4=F_4$. For shallower stages $1\leq i \leq 3 $, we have $\tilde{F}_{i} = \text{Up}_i({F'}_{i+1}) + {F}_{i} \in \mathbb{R}^{N_{i} \times D_{i}}$ to combine upsampled deeper features with current-level features. The query, key, and value in this reverse cross-attention is then computed as $Q_i^{(h)} =\tilde{F}_i W_i^{(Q, h)} \in \mathbb{R}^{N_{i} \times d_{h, i}}$, $K_i^{(h)} ={Q}_i W_i^{(K, h)} \in \mathbb{R}^{1 \times d_{h, i}}$, and $V_i^{(h)} ={Q}_i W_i^{(V, h)} \in \mathbb{R}^{1 \times d_{h, i}}$, where $W_i^{(Q, h)}$, $W_i^{(K, h)}$, $W_i^{(V, h)} \in \mathbb{R}^{D_{i}\times d_{h,i}}$, and $d_{h,i} = D_{i}/h_{i}$. The outputs
of all above attention heads are also obtained following the standard scaled dot-product attention, i.e., $O_i^{(h)} =A_i^{(h)} V_i^{(h)} \in \mathbb{R}^{N_{i} \times d_{h, i}}$ and $A_i^{(h)}=\operatorname{Softmax}(\frac{Q_i^{(h)}(K_i^{(h)})^{\top}}{\sqrt{d_{h, i}}}) \in \mathbb{R}^{N_i \times 1}$. Next, these multi-head outputs are concatenated and projected as $F_i^{''}=\operatorname{Concat}(O_i^{(1)}, \ldots, O_i^{\left(h_i\right)}) W_i^O   + b_i^{(O)} \in \mathbb{R}^{N_i \times D_i}$, where $W_i^{(O)} \in \mathbb{R}^{D_i \times D_i}$ and $b_i^{(O)} \in \mathbb{R}^{D_i}$. Finally, the reconstructed features are obtained through residual connection and layer normalization as follows:
\begin{equation}
F^{'}_i = \text{LN}(\tilde{F}^{'} + F_{i}^{''}) \in \mathbb{R}^{N_i \times D_i} 
\end{equation}

Through this reverse cross-attention, MGFRM achieves two critical objectives: utilizing the compact semantic guidance from MGQM to reconstruct spatially detailed features, and ensuring that spatial feature reconstruction remains consistent with the saliency understanding established by MGQM and the features extracted from HFEM. The complementary roles of semantic localization and spatial reconstruction in the dual-gaze architecture serve as the foundation for the subsequent precise saliency mask generation.

\subsection{Saliency Mask Generation Module}\label{sec:mask-generator}
Following the dual-gaze processing pipeline, we obtain enriched query tokens $\{Q_{i}\}_{i=1}^{4}$ from MGQM and enhanced hierarchical features $\{F^{'}_{i}\}_{i=1}^{4}$ from MGFRM, encoding semantic saliency understanding and detailed spatial information respectively. The SMGM synthesizes these dual-pathway outputs into unified pixel-level saliency predictions.

Through the hierarchical processing in MGFRM, the finest-scale feature $F_{1}^{'}$  already incorporates information from all deeper levels via progressive upsampling and feature fusion from $F_{4}^{'}$ to $F_{1}^{'}$ and $F_{4}$ to $F_{1}$. Similarly, $Q_{1}$ inherits a multi-scale semantic understanding through top-down query propagation in MGQM, progressing from deeper to shallower levels. Therefore, these finest-scale representations encode the complete hierarchical information from both pathways, and SMGM focuses solely on the finest-scale representations $F_{1}^{'}$ and $Q_{1}$.

\begin{figure}[!t]
\vspace{-4pt}
  \centering
  \begin{tikzpicture}%
    \begin{scope}[xshift=0cm, yshift=0cm ]
        \node[above right] (fig63) at (0.0,0.4){\hspace{-0.1cm}\includegraphics[ height=0.554\linewidth]{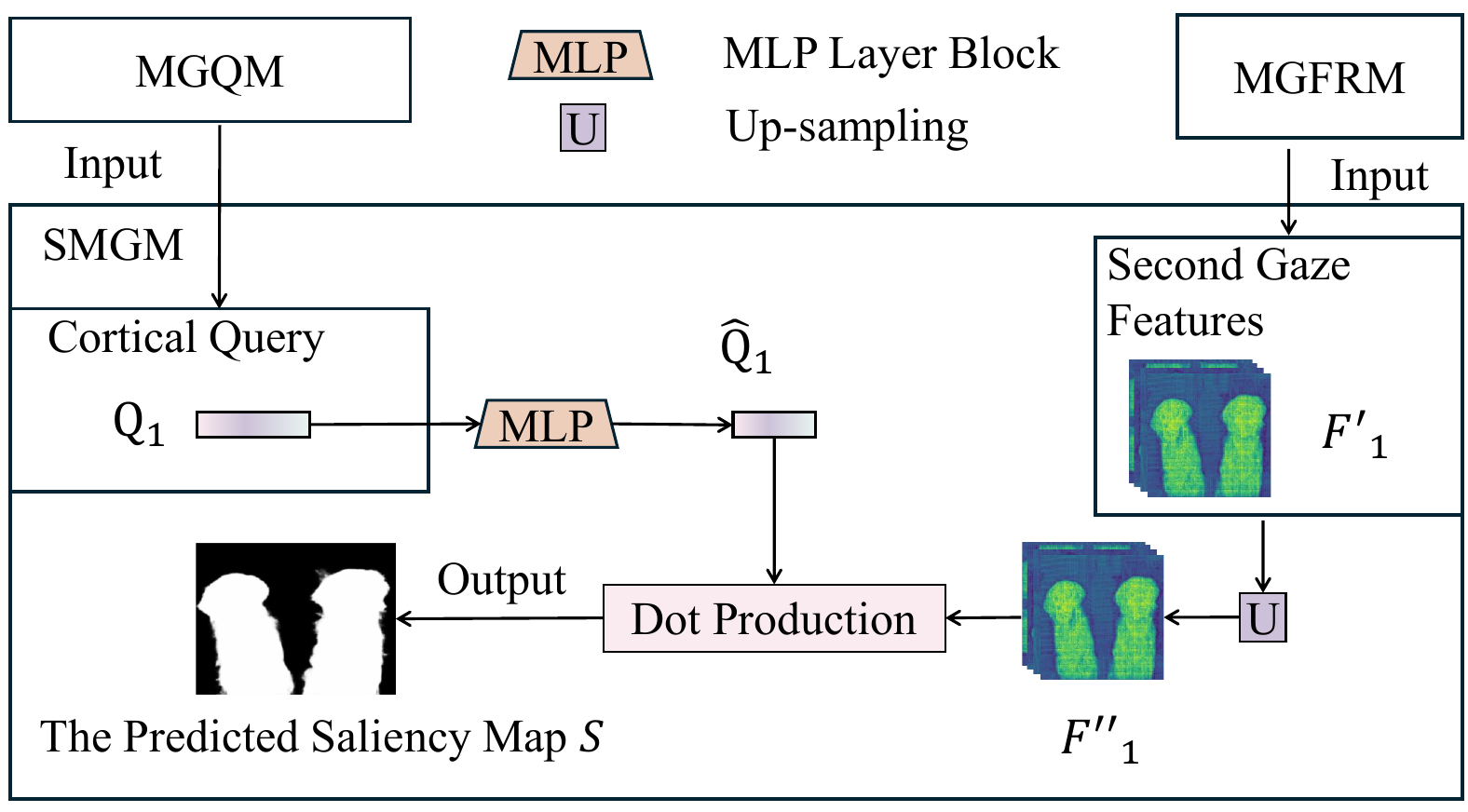}};
    \end{scope}
    
  \end{tikzpicture}
  \caption{Detailed pipeline of the SMGM. The module integrates spatial feature $F_{1}^{'}$ from the second gaze and semantic guidance $Q_{1}$ from the cortical query through a two-branch processing pipeline, culminating in pixel-level saliency prediction via element-wise similarity computation.}
  \label{Fig.MG}
  \vspace{-0.5em}
\end{figure}

SMGM processes these representations through a two-branch pipeline that aligns their dimensions before generating the final prediction, as illustrated in Fig.~\ref{Fig.MG}. 

\textit{a) Finest-scale reconstructed feature upsampling.} The mask generation process starts by upsampling the lowest-level reconstructed features $F_{1}^{'}$. Instead of collapsing the channel dimensionality in a single step, this stage provides a smooth transition toward the final resolution, thereby mitigating information loss and preserving discriminative cues necessary for the following mask prediction: 
\begin{equation} \label{feature1}
\begin{aligned}
 &F^{(1)}_1 = \text{ConvTranspose}(F'_1) \in \mathbb{R}^{H^{\uparrow 1} \times W^{\uparrow 1} \times D_m} \\
 &F^{(2)}_1 = \text{GELU}(F^{(1)}_1) \in \mathbb{R}^{H^{\uparrow 1} \times W^{\uparrow 1} \times D_m} \\
 &F^{(3)}_1 = \text{ConvTranspose}(F^{(2)}_1)\in \mathbb{R}^{H^{\uparrow 2} \times W^{\uparrow 2} \times D_m}  \\
 &F^{(4)}_1 = \text{GELU}(F^{(3)}_1) \in \mathbb{R}^{H^{\uparrow 2} \times W^{\uparrow 2} \times D_m} \\
 & F^{''}_1= \text{MLP}(F^{(4)}_1) \in \mathbb{R}^{H \times W \times D_{\text{mask}}}  
\end{aligned}
\end{equation}
where the two transposed convolution $\text{ConvTranspose}(\cdot)$ operations progressively increase spatial resolution, and the three-layer $\text{MLP}(\cdot)$  performs gradual channel dimension reduction to avoid abrupt transitions.

\textit{b) Finest-scale Query Processing.} In parallel, we process the query token $Q_{1}$ through an MLP to align its dimensions with the upsampled features:  
\begin{equation}\label{query1}
\hat{Q}_1= \text{MLP}(Q_1) \in \mathbb{R}^{1 \times D_{\text{mask}}}
\end{equation} 

Finally, the saliency mask integrates the above spatial and semantic information through a element-wise similarity computation detailed as $S \in \mathbb{R}^{H \times W}$ with $S(x,y) =\left\langle F_1^{\prime \prime}(x, y,:), \hat{Q}_1\right\rangle=\sum_{c=1}^{D_{\text {mask }}} F_1^{\prime \prime}(x, y, c) \hat{Q}_1(c) \in \mathbb{R}$, where $F_1^{\prime \prime}(x, y,:)$ is the feature vector at pixel $(x,y)$, and this dot product directly measures the similarity between the spatial feature at $(x,y)$ and the global semantic guidance vector $\hat{Q}_{1}$. 


\subsection{Loss Function}
Consistent with minimalist design philosophy, we employ a streamlined loss formulation that avoids complex schemes or heuristic weighting strategies. The training objective only combines binary cross-entropy (BCE) loss and Dice loss to provide dual supervision for saliency prediction. Given the predicted saliency logits $S \in \mathbb{R}^{H \times W}$ and binary ground-truth mask $G \in \{0,1\}^{H\times W} $, the loss is formulated as: 
\begin{equation}
\mathcal{L}_{\text{total}} =  \mathcal{L}_{\text{Dice}} + \mathcal{L}_{\text{BCE}}.
\end{equation}
where $\mathcal{L}_{\text{Dice}} = 1 - \frac{2 \sum_{x=1}^{H}\sum_{y=1}^{W} \sigma(S(x,y)) G(x,y) + \epsilon}{\sum_{x=1}^{H}\sum_{y=1}^{W} \sigma(S(x,y))^2 + \sum_{x=1}^{H}\sum_{y=1}^{W} G(x,y)^2 + \epsilon}$, $\mathcal{L}_{\text{BCE}} = -\frac{1}{HW} \sum_{x=1}^{H}\sum_{y=1}^{W} [ G(x,y) \log \sigma(S(x,y)) 
+ (1-G(x,y)) \log (1-\sigma(S(x,y)))$, $\sigma(\cdot)$ denotes the sigmoid activation function that transforms logits to probabilities, and $\epsilon > 0$ is a small constant for numerical stability. 

This hybrid loss design directly reflects our dual-gaze architecture. Specifically, the BCE loss ensures fine-grained pixel-level accuracy that aligns with the detailed spatial reconstruction from MGFRM, and the Dice loss enforces holistic region consistency that reflects the semantic understanding from MGQM. The following experiments show that this balanced design is sufficient to yield strong performance across diverse SOD benchmarks.

\begin{table*}[!t]
\centering
  \definecolor{mygreen}{RGB}{0,102,0}
  \definecolor{myred}{RGB}{153,0,0}
  \definecolor{myblue}{RGB}{0,122,122}
    \definecolor{myorange}{RGB}{204,106,0}
\caption{: Quantitative comparison on five benchmark SOD datasets. The best results are in \textcolor{myred}{red} marked \textbf{in bold}, second best in \textcolor{myorange}{orange} marked \textbf{in bold}, and third best in \textcolor{mygreen}{green} marked \textbf{in bold}. MAE: lower is better ($\downarrow$); other metrics: higher is better ($\uparrow$). Numbers in parentheses indicate dataset sizes. Please refer to the text for more details.}
\label{tab:sod_comparison}
\renewcommand{\arraystretch}{0.8}
\setlength{\tabcolsep}{2pt}
\begin{tabular*}{\textwidth}{@{\extracolsep{\fill}}l|cccc|cccc|cccc|cccc|cccc@{}}
\toprule
\multirow{2}{*}{\textbf{Method}} &
\multicolumn{4}{c|}{\textbf{DUTS-TE (5019)}} &
\multicolumn{4}{c|}{\textbf{DUT-OMRON (5168)}} &
\multicolumn{4}{c|}{\textbf{ECSSD (1000)}} & 
\multicolumn{4}{c|}{\textbf{HKU-IS (4447)}} & 
\multicolumn{4}{c}{\textbf{PASCAL-S (850)}} \\
\cmidrule(lr){2-5} \cmidrule(lr){6-9} \cmidrule(lr){10-13} \cmidrule(lr){14-17} \cmidrule(lr){18-21}
& MAE & $F_\beta^{\max}$ & $S_m$ & $E_m$ 
& MAE & $F_\beta^{\max}$ & $S_m$ & $E_m$ 
& MAE & $F_\beta^{\max}$ & $S_m$ & $E_m$ 
& MAE & $F_\beta^{\max}$ & $S_m$ & $E_m$ 
& MAE & $F_\beta^{\max}$ & $S_m$ & $E_m$ \\
\midrule
\multicolumn{21}{c}{\textit{CNN-based Methods}} \\
\midrule
CaGNet (2020) \cite{cagnet} & .029 & .898 & .897 & .939 & .047 & .818 & .845 & .882 & .026 & .950 & .930 & .959 & .024 & .940 & .923 & .961 & .063 & .878 & .870 & .917 \\
DFI (2020) \cite{dfi} & .039 & .896 & .887 & .912 & .055 & .839 & .840 & .862 & .035 & .956 & .927 & .947 & .031 & .940 & .919 & .948 & .065 & .892 & .865 & .898 \\
GateNet (2020) \cite{gatenet} & .032 & .904 & .900 & .930 & .053 & .828 & .842 & .867 & .028 & .960 & .936 & .959 & .026 & .946 & .925 & .957 & .058 & .882 & .868 & .905 \\
MINet (2020) \cite{minet} & .037 & .884 & .884 & .917 & .056 & .831 & .833 & .860 & .033 & .954 & .925 & .950 & .029 & .942 & .919 & .952 & .064 & .881 & .856 & .896 \\
LDF (2020) \cite{ldf} & .034 & .905 & .892 & .925 & .052 & .835 & .839 & .865 & .034 & .956 & .924 & .948 & .028 & .943 & .919 & .953 & .060 & .887 & .863 & .903 \\
EDN (2020) \cite{9756227} & .035 & .893 & .892 & .934 & .050 & .821 & .849 & .885 & .027 & .940 & .924 & .962 & .033 & .950 & .927 & .958 & .062 & .879 & .865 & .908 \\
TE7 (2022) \cite{tracer} & .023 & .932 & .920 & .954 & .045 & .849 & .856 & .884 & .026 & .962 & .936 & .959 & \textcolor{mygreen}{\textbf{.021}} & .953 & .933 & .966 & \textcolor{mygreen}{\textbf{.046}} & .907 & .884 & .929 \\
MENet (2023) \cite{menet} & .028 & .918 & .905 & .938 & .045 & .845 & .850 & .871 & \textcolor{mygreen}{\textbf{.021}} & .957 & .928 & .951 & .023 & .951 & .927 & .960 & .053 & .897 & .872 & .910 \\
DC-Net (2023) \cite{dcnet} & .023 & .932 & .925 & .952 & \textcolor{mygreen}{\textbf{.039}} & .868 & .875 & .898 & .023 & .968 & .947 & .965 & \textcolor{mygreen}{\textbf{.021}} & .957 & \textcolor{mygreen}{\textbf{.941}} & .966 & .049 & .904 & .887 & .917 \\
BBRF (2023) \cite{bbrf} & .025 & .911 & .909 & .949 & .044 & .839 & .861 & .896 & .022 & .961 & .939 & \textcolor{mygreen}{\textbf{.969}} & \textcolor{myorange}{\textbf{.020}} & .949 & .932 & \textcolor{myorange}{\textbf{.969}} & .049 & .887 & .878 & .923 \\
ELSA-Net (2024) \cite{10155248} & .034 & .882 & -- & .934 & .050 & .794 & -- & .891 & .030 & .934 & -- & .961 & .025 & .935 & -- & .964 & .059 & .862 & -- & .912 \\
PAM (2025) \cite{pam} & .029 & .901 & .903 & .944 & .048 & .831 & .859 & .904 & .026 & .957 & .942 & .966 & .023 & .944 & .933 & .965 & .058 & .895 & .887 & .917 \\
RMFDNet (2025) \cite{zhou2025rmfdnet} & .034 & .903 & .897 & .927 & .052 & .832 & .845 & .869 & .033 & .953 & .927 & .947 & .028 & .944 & .923 & .953 & .060 & .884 & .865 & .901\\
\midrule
\multicolumn{21}{c}{\textit{Transformer-based Methods}} \\
\midrule
VST (2021) \cite{Liu_2021_ICCV} & .037 & .895 & .896 & .919 & .058 & .836 & .850 & .871 & .033 & .954 & .932 & .951 & .029 & .946 & .928 & .952 & .061 & .882 & .872 & .902 \\
ICON (2022) \cite{icon} & .026 & .922 & .916 & .950 & .044 & .860 & .868 & .897 & .025 & .960 & .940 & .962 & .023 & .952 & .934 & .964 & .049 & .903 & .886 & .923 \\
VST++ (2023) \cite{10497889} & .025 & .928 & .921 & .950 & .044 & .861 & .870 & .897 & .022 & .969 & .947 & .968 & \textcolor{mygreen}{\textbf{.021}} & .957 & .938 & \textcolor{mygreen}{\textbf{.967}} & .047 & .905 & .889 & .924 \\
SwinSOD (2024) \cite{WU2024105039} & .025 & .930 & .916 & .949 & .042 & .866 & .866 & .894 & .024 & .966 & .942 & .965 & \textcolor{myorange}{\textbf{.020}} & .958 & .937 & \textcolor{myorange}{\textbf{.969}} & .047 & .912 & .887 & .925\\
MDSAM (2024) \cite{mdsam} & .024 & .937 & .920 & .949 & \textcolor{mygreen}{\textbf{.039}} & \textcolor{mygreen}{\textbf{.887}} & .878 & .910 & \textcolor{mygreen}{\textbf{.021}} & \textcolor{myorange}{\textbf{.974}} & .948 & .967 & \textcolor{myred}{\textbf{.019}} & \textcolor{myorange}{\textbf{.963}} & .941 & \textcolor{myorange}{\textbf{.969}} & .052 & .907 & .882 & .917 \\
Sam2unet (2024) \cite{sam2unet} & \textcolor{myorange}{\textbf{.020}} & .942 & \textcolor{mygreen}{\textbf{.934}} & \textcolor{myorange}{\textbf{.958}} & \textcolor{mygreen}{\textbf{.039}} & .878 & \textcolor{mygreen}{\textbf{.883}} & .910& \textcolor{myorange}{\textbf{.020}} & .970 & \textcolor{mygreen}{\textbf{.950}} & \textcolor{myorange}{\textbf{.970}} & \textcolor{myred}{\textbf{.019}} & \textcolor{mygreen}{\textbf{.960}} & .941 & \textcolor{myred}{\textbf{.971}} & \textcolor{myorange}{\textbf{.043}} & .913 & .894 & \textcolor{myorange}{\textbf{.931}} \\
BiRefNet (2024) \cite{birefnet} & \textcolor{myred}{\textbf{.019}} & \textcolor{myorange}{\textbf{.948}} & \textcolor{myorange}{\textbf{.940}} & \textcolor{myorange}{\textbf{.958}} & .040 & .856 & .869 & .878 & \textcolor{mygreen}{\textbf{.021}} & \textcolor{mygreen}{\textbf{.973}} & \textcolor{mygreen}{\textbf{.950}} & .965 & \textcolor{myorange}{\textbf{.020}} & \textcolor{myorange}{\textbf{.963}} & \textcolor{myorange}{\textbf{.944}} & \textcolor{mygreen}{\textbf{.967}} & \textcolor{myred}{\textbf{.041}} & \textcolor{myorange}{\textbf{.924}} & \textcolor{myorange}{\textbf{.900}} & \textcolor{myorange}{\textbf{.931}} \\
SefNet (2024) \cite{Yang2024_SEF_E2E} & .023 & .938 & .930 & -- & .043 & .846 & .872 & -- & \textcolor{mygreen}{\textbf{.021}} & .967 & \textcolor{myorange}{\textbf{.951}} & -- & \textcolor{mygreen}{\textbf{.021}} & .956 & \textcolor{mygreen}{\textbf{.942}} & -- & \textcolor{mygreen}{\textbf{.046}} & .913 & \textcolor{mygreen}{\textbf{.897}} & --\\
PDNet (2025) \cite{Cen2025_PDDNet} & .027 & .900 & .912 & -- & .049 & .814 & .860 & -- & .024 & .956 & .943 & -- & .023 & .944 & .933 & -- & .049 & .884 & .882 & --\\
CamoDiffusion (2025) \cite{10834569} & .023 & .902 & .921 & -- & .042 & .806 & .871 & -- & -- & -- & -- & -- & .022 & .929 & .934 & -- & -- & -- & -- & -- \\
SegR1 (2025) \cite{segR1} & .025 & .922 & .925 & .953 & .045 & .850 & .878 & \textcolor{mygreen}{\textbf{.911}} & .025 & .956 & .939 & .956 & .022 & .950 & .935 & .966 & -- & -- & -- & -- \\
ECF-DT (2025) \cite{wang2025novel} & .036 & .812 & .909 & .899 & .060 & .757 & .854 & .868 & .031 & .899 & .936 & .922 & .028 & .907 &\textcolor{myred}{\textbf{.955}} & .929 & .056 & .809 & .876 & .867 \\
\midrule
\multicolumn{21}{c}{\textit{Our Methods}} \\
\midrule
DGN-B (Ours) & \textcolor{mygreen}{\textbf{.022}} & \textcolor{mygreen}{\textbf{.945}} & .930 & \textcolor{mygreen}{\textbf{.956}} & \textcolor{myorange}{\textbf{.036}} & \textcolor{myorange}{\textbf{.892}} & \textcolor{myorange}{\textbf{.890}} & \textcolor{myorange}{\textbf{.917}} & .022 & .972 & .948 & .965 & \textcolor{mygreen}{\textbf{.021}} & \textcolor{myorange}{\textbf{.963}} & .941 & \textcolor{mygreen}{\textbf{.967}} & \textcolor{mygreen}{\textbf{.046}} & \textcolor{mygreen}{\textbf{.920}} & .893 & \textcolor{mygreen}{\textbf{.926}} \\
DGN-L (Ours) & \textcolor{myred}{\textbf{.019}} & \textcolor{myred}{\textbf{.951}} & \textcolor{myred}{\textbf{.939}} & \textcolor{myred}{\textbf{.962}} & \textcolor{myred}{\textbf{.034}} & \textcolor{myred}{\textbf{.893}} & \textcolor{myred}{\textbf{.894}} & \textcolor{myred}{\textbf{.919}} & \textcolor{myred}{\textbf{.018}} & \textcolor{myred}{\textbf{.978}} & \textcolor{myred}{\textbf{.956}} & \textcolor{myred}{\textbf{.972}} & \textcolor{myred}{\textbf{.019}} & \textcolor{myred}{\textbf{.966}} & \textcolor{myorange}{\textbf{.944}} & \textcolor{myred}{\textbf{.971}} & \textcolor{myred}{\textbf{.041}} & \textcolor{myred}{\textbf{.926}} & \textcolor{myred}{\textbf{.902}} & \textcolor{myred}{\textbf{.934}} \\
\bottomrule
\end{tabular*}
\end{table*}

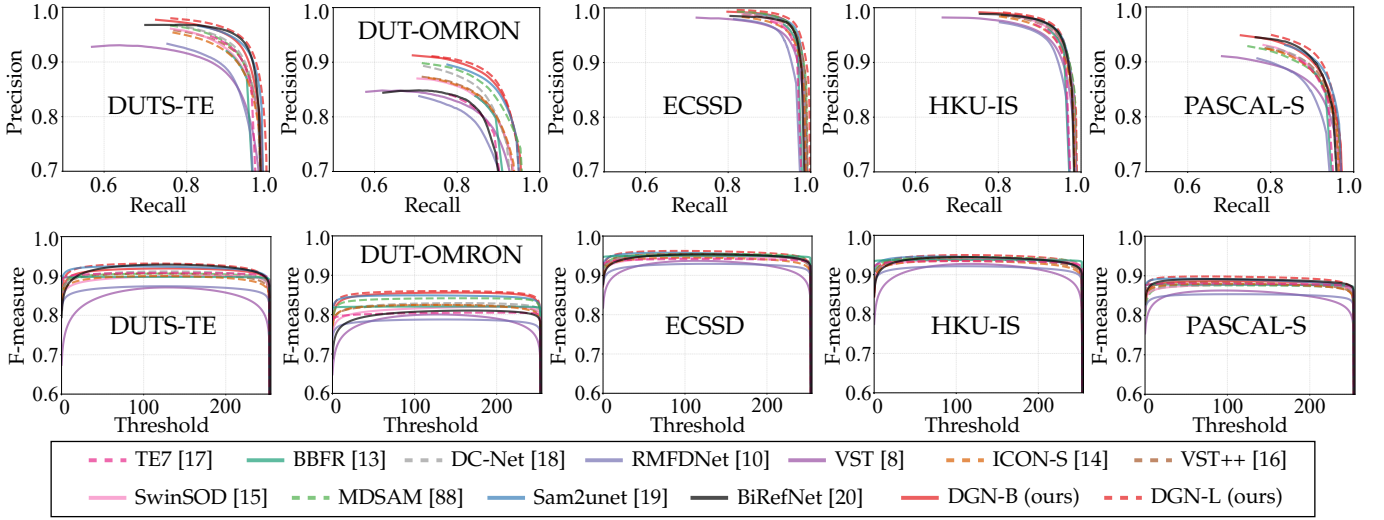
\begin{figure*}[!t]
\vspace{-5pt}
  \centering
  \begin{tikzpicture}
    \begin{scope}[xshift=0cm, yshift=0cm]
        \node[above right] (fig1) at (0,0){\includegraphics[height=0.1405\linewidth]{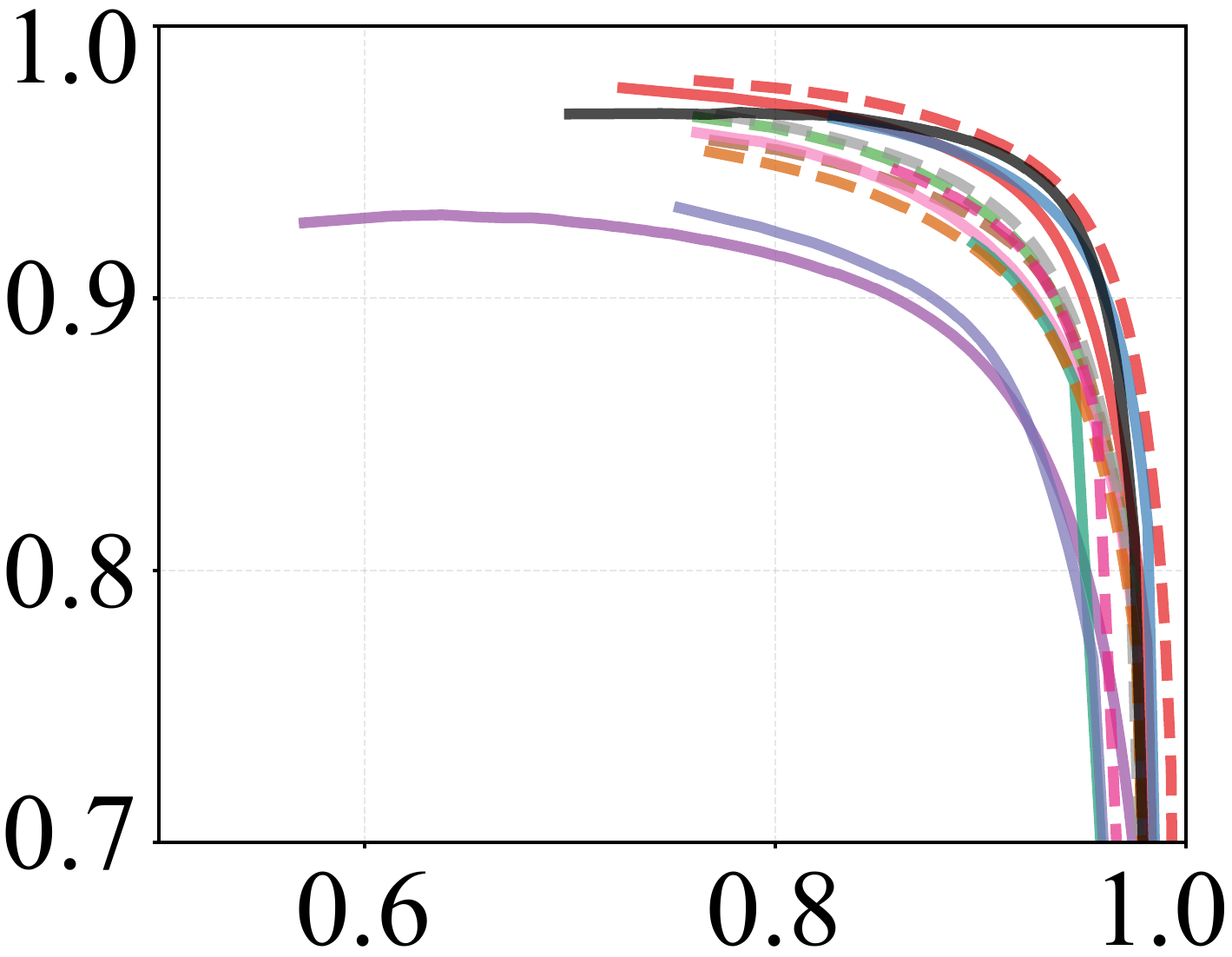}};
        \node at ($(fig1.south)+(0.05,-0.02)$) {\footnotesize Recall};
        \node[rotate=90] at ($(fig1.west)+(-0.1,0.08)$) {\footnotesize Precision};
        \node[inner sep=1pt] at (1.85,1.3) {DUTS-TE};
    \end{scope}
    
    \begin{scope}[xshift=3.6cm, yshift=0cm]
        \node[above right] (fig2) at (0,0){\includegraphics[height=0.1405\linewidth]{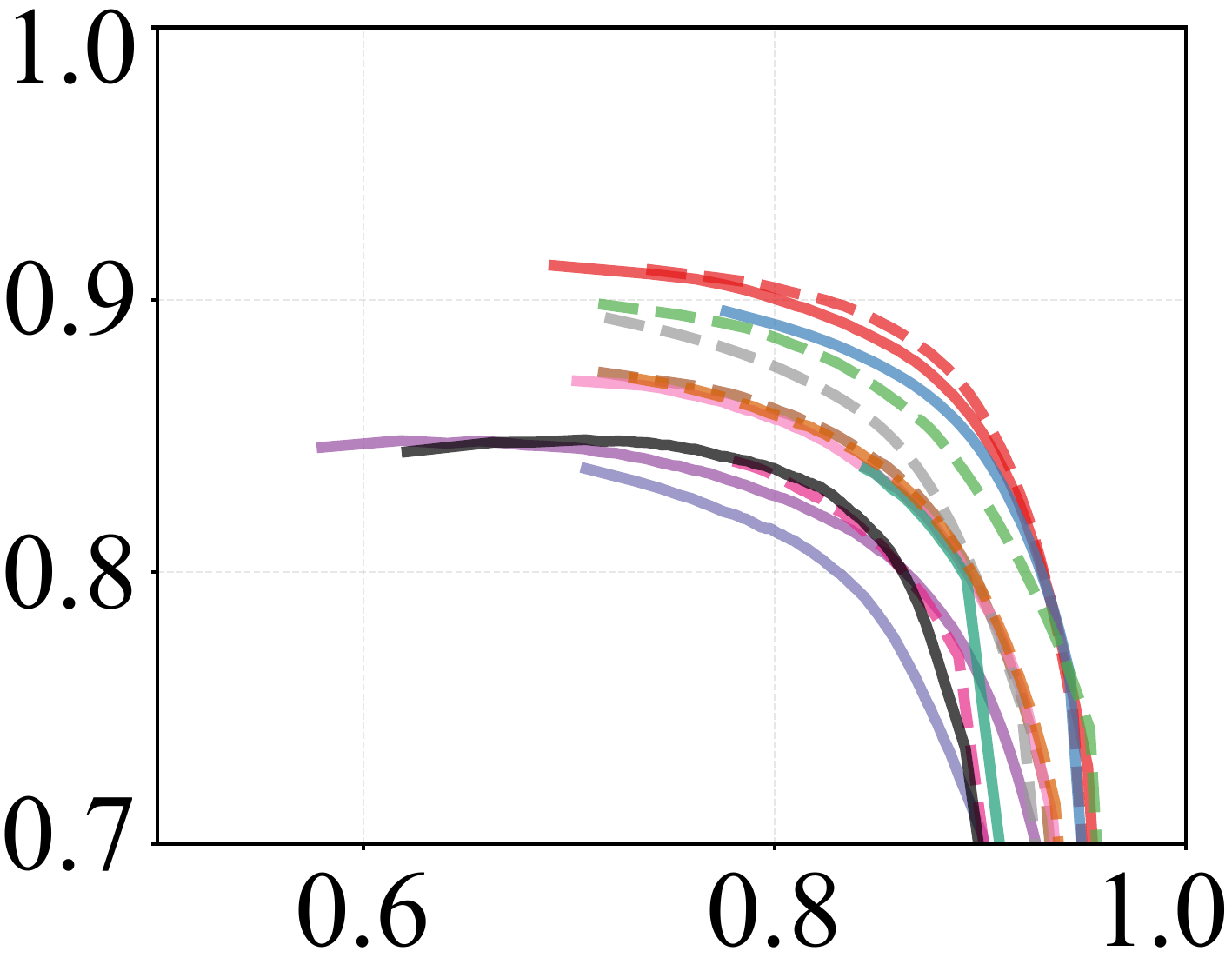}};
        \node[rotate=90] at ($(fig2.west)+(-0.05,0.08)$) {\footnotesize Precision};
        \node at ($(fig2.south)+(0.10,-0.02)$) {\footnotesize Recall};
        \node[inner sep=1pt] at (1.95,2.3) {DUT-OMRON};
    \end{scope}
    
    \begin{scope}[xshift=7.2cm, yshift=0cm]
        \node[above right] (fig3) at (0,0){\includegraphics[height=0.1405\linewidth]{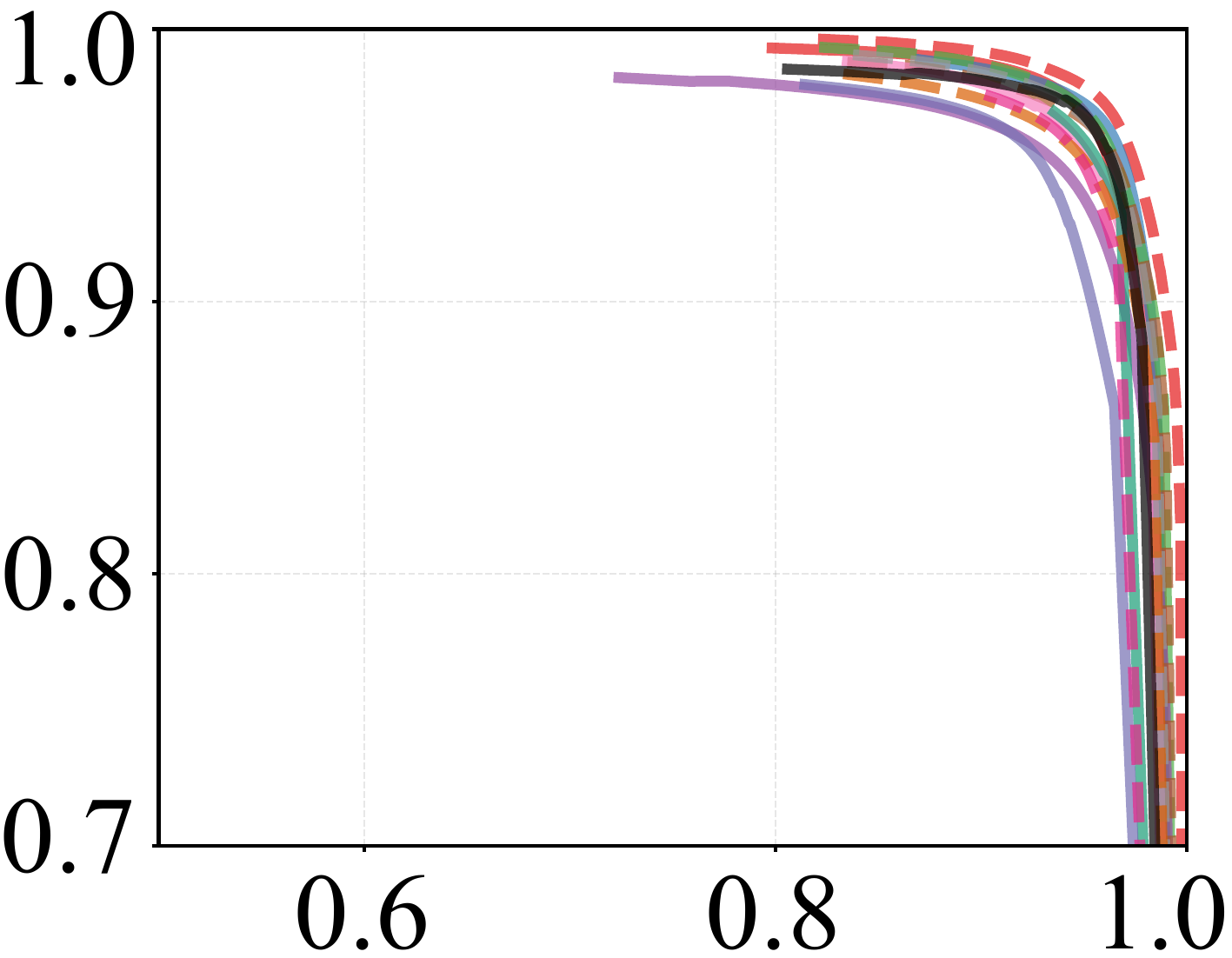}};
        \node at ($(fig3.south)+(0.05,-0.02)$) {\footnotesize Recall};
        \node[rotate=90] at ($(fig3.west)+(-0.1,0.08)$) {\footnotesize Precision};
        \node[inner sep=1pt] at (1.85,1.3) {ECSSD};
    \end{scope}
    
    \begin{scope}[xshift=10.8cm, yshift=0cm]
        \node[above right] (fig4) at (0,0){\includegraphics[height=0.1405\linewidth]{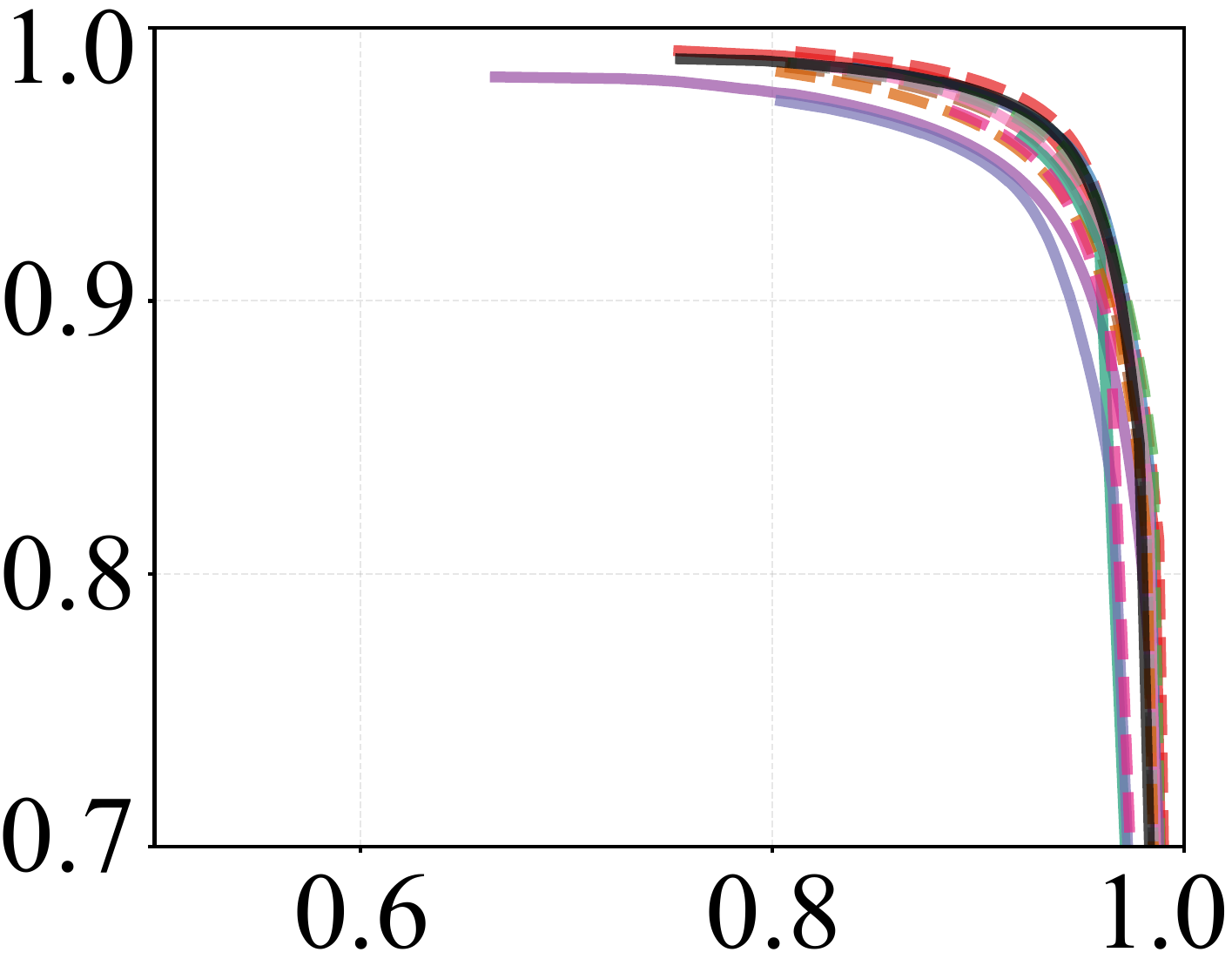}};
        \node[rotate=90] at ($(fig4.west)+(-0.05,0.08)$) {\footnotesize Precision};
        \node at ($(fig4.south)+(0.10,-0.02)$) {\footnotesize Recall};
        \node[inner sep=1pt] at (1.85,1.3) {HKU-IS};
    \end{scope}
    
    \begin{scope}[xshift=14.4cm, yshift=0cm]
        \node[above right] (fig5) at (0,0){\includegraphics[height=0.1405\linewidth]{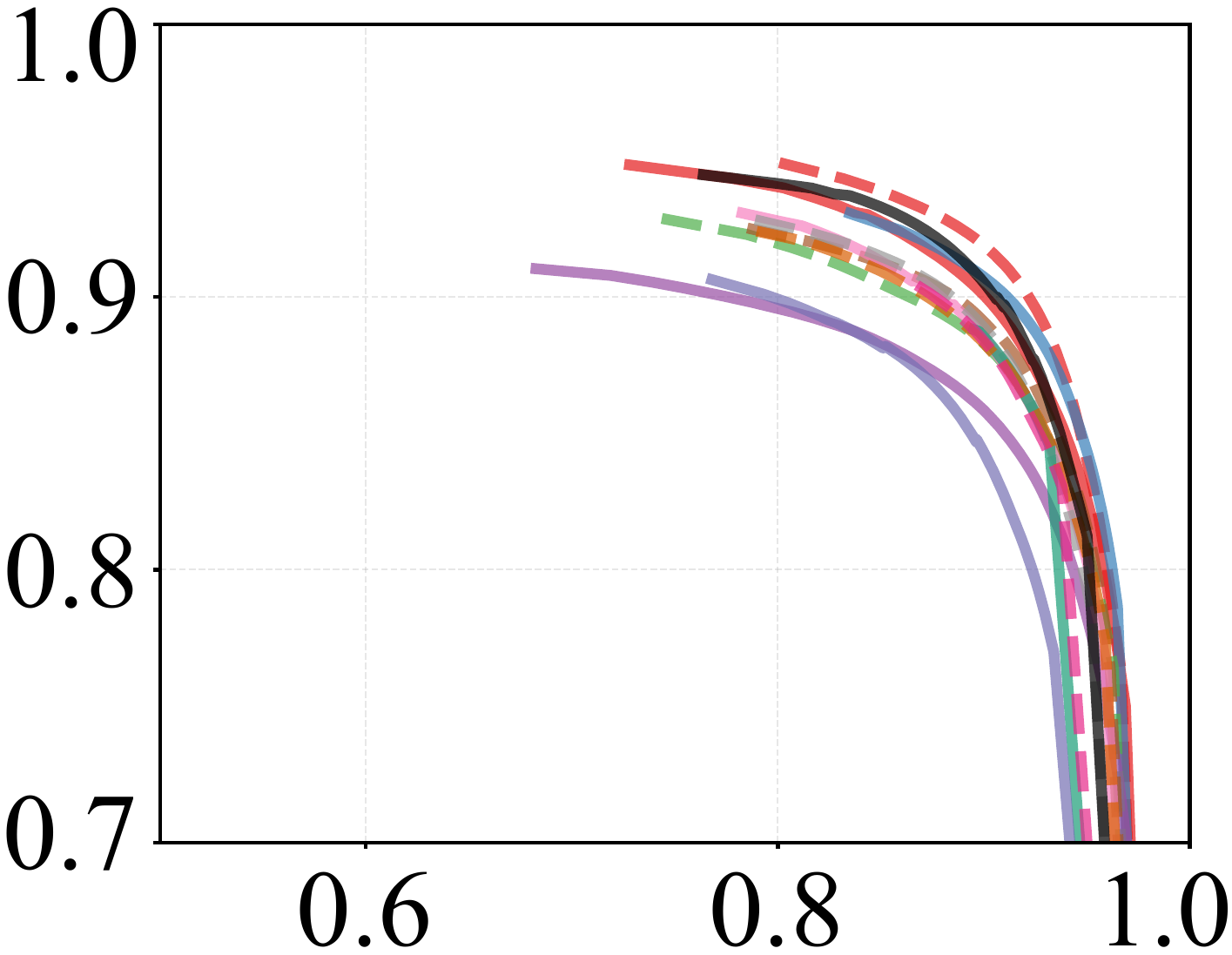}};
        \node at ($(fig5.south)+(0.05,-0.02)$) {\footnotesize Recall};
        \node[rotate=90] at ($(fig5.west)+(-0.1,0.08)$) {\footnotesize Precision};
        \node[inner sep=1pt] at (1.85,1.3) {PASCAL-S};
    \end{scope}
    
  \end{tikzpicture}

  \begin{tikzpicture}
    \begin{scope}[xshift=0cm, yshift=0cm]
        \node[above right] (fig1) at (0,0){\includegraphics[height=0.135\linewidth]{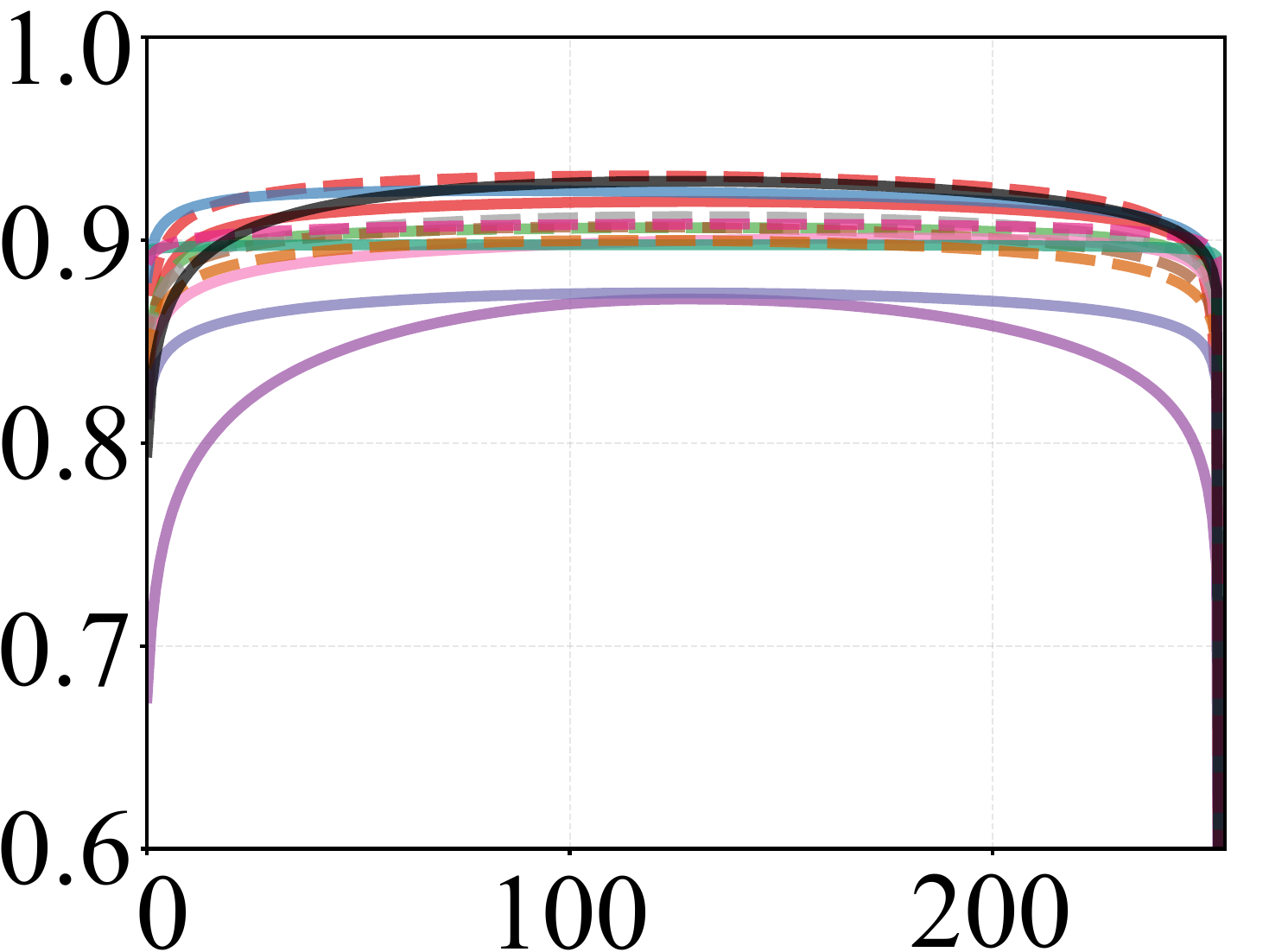}};
        \node at ($(fig1.south)+(0.05,-0.02)$) {\footnotesize Threshold};
        \node[rotate=90] at ($(fig1.west)+(-0.1,0.08)$) {\footnotesize F-measure};
        \node[inner sep=1pt] at (1.85,1.3) {DUTS-TE};
    \end{scope}
    
    \begin{scope}[xshift=3.6cm, yshift=0cm]
        \node[above right] (fig2) at (0,0){\includegraphics[height=0.135\linewidth]{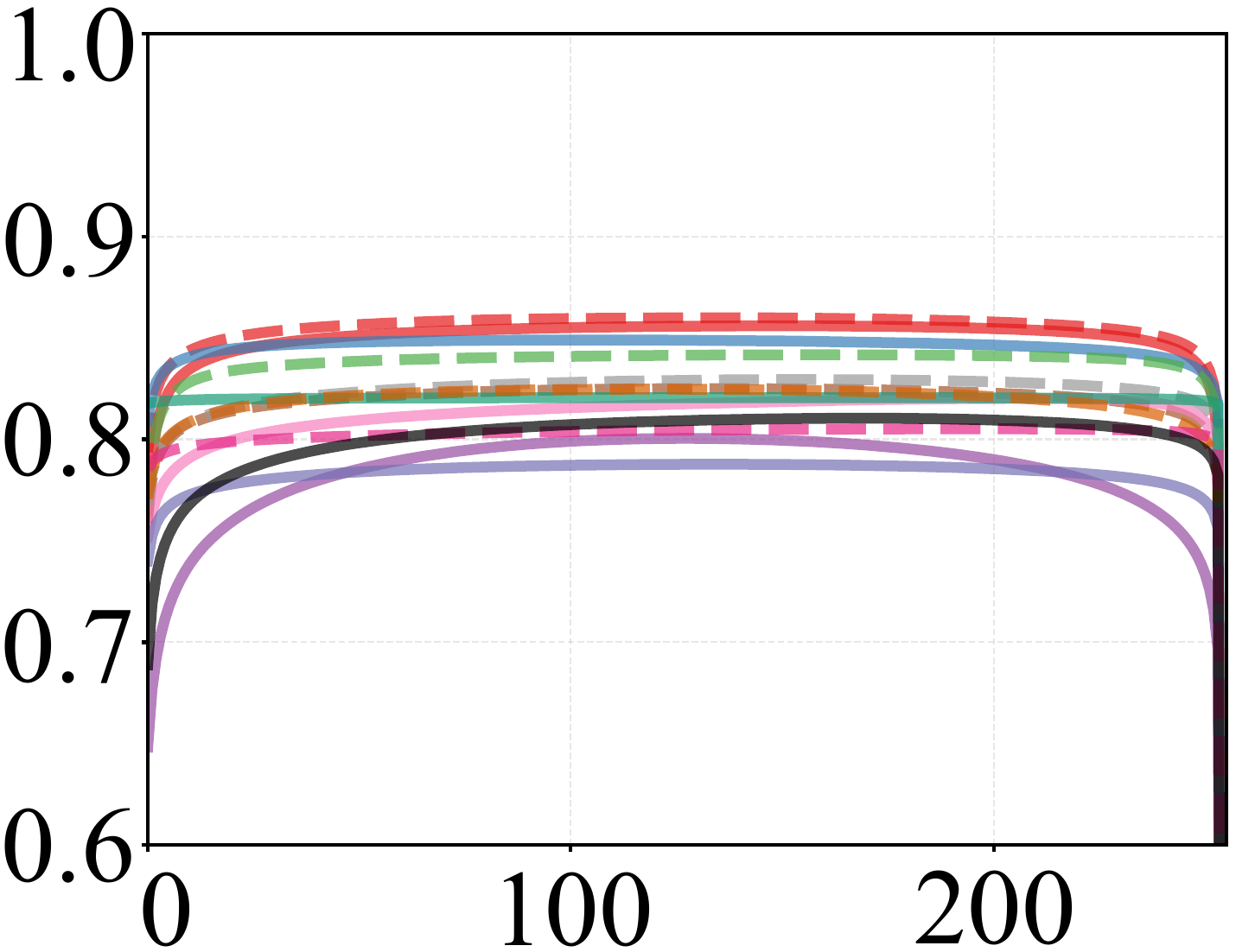}};
        \node[rotate=90] at ($(fig2.west)+(-0.05,0.08)$) {\footnotesize F-measure};
        \node at ($(fig2.south)+(0.10,-0.02)$) {\footnotesize Threshold};
        \node[inner sep=1pt] at (1.95,2.25) {DUT-OMRON};
    \end{scope}
    
    \begin{scope}[xshift=7.2cm, yshift=0cm]
        \node[above right] (fig3) at (0,0){\includegraphics[height=0.135\linewidth]{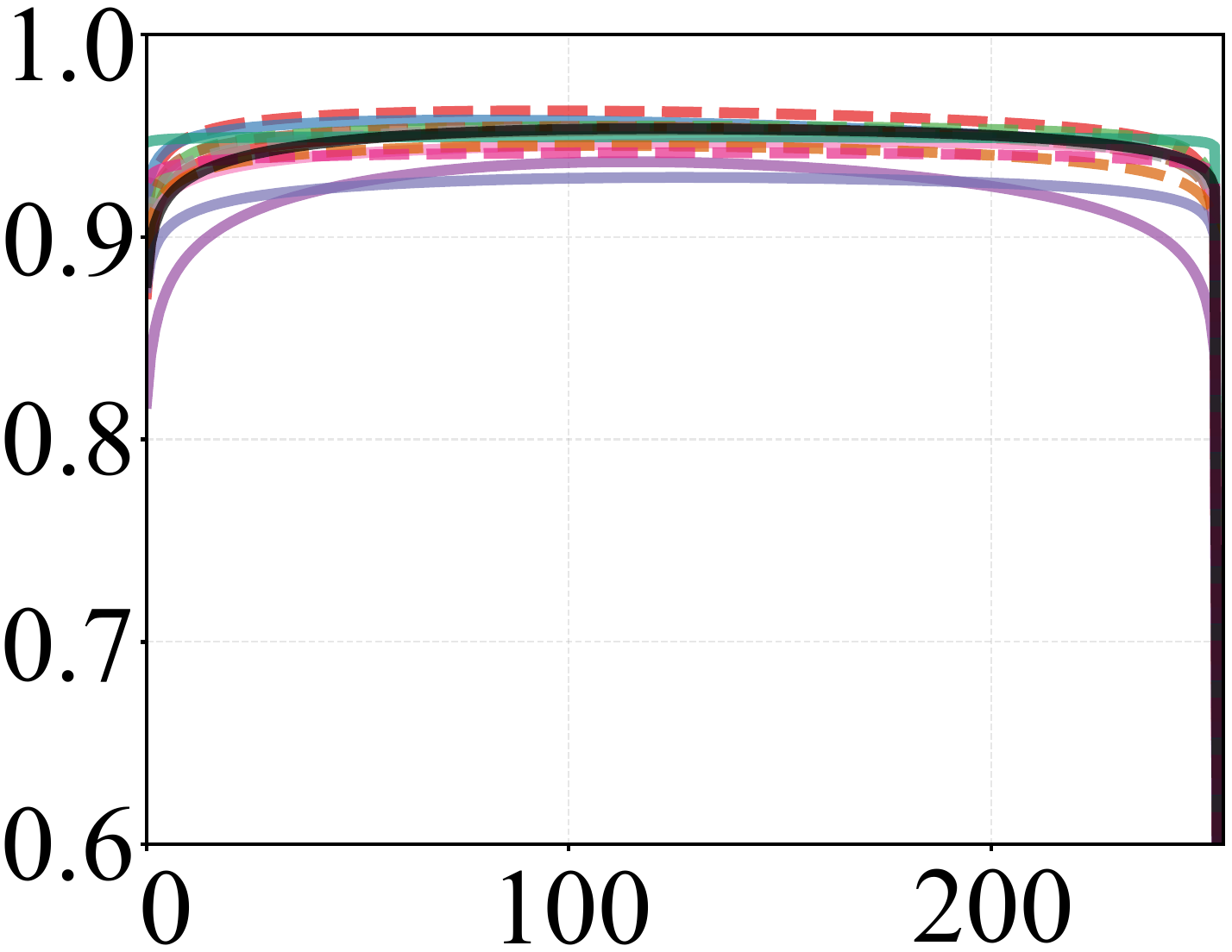}};
        \node at ($(fig3.south)+(0.05,-0.02)$) {\footnotesize Threshold};
        \node[rotate=90] at ($(fig3.west)+(-0.1,0.08)$) {\footnotesize F-measure};
        \node[inner sep=1pt] at (1.85,1.3) {ECSSD};
    \end{scope}
    
    \begin{scope}[xshift=10.8cm, yshift=0cm]
        \node[above right] (fig4) at (0,0){\includegraphics[height=0.135\linewidth]{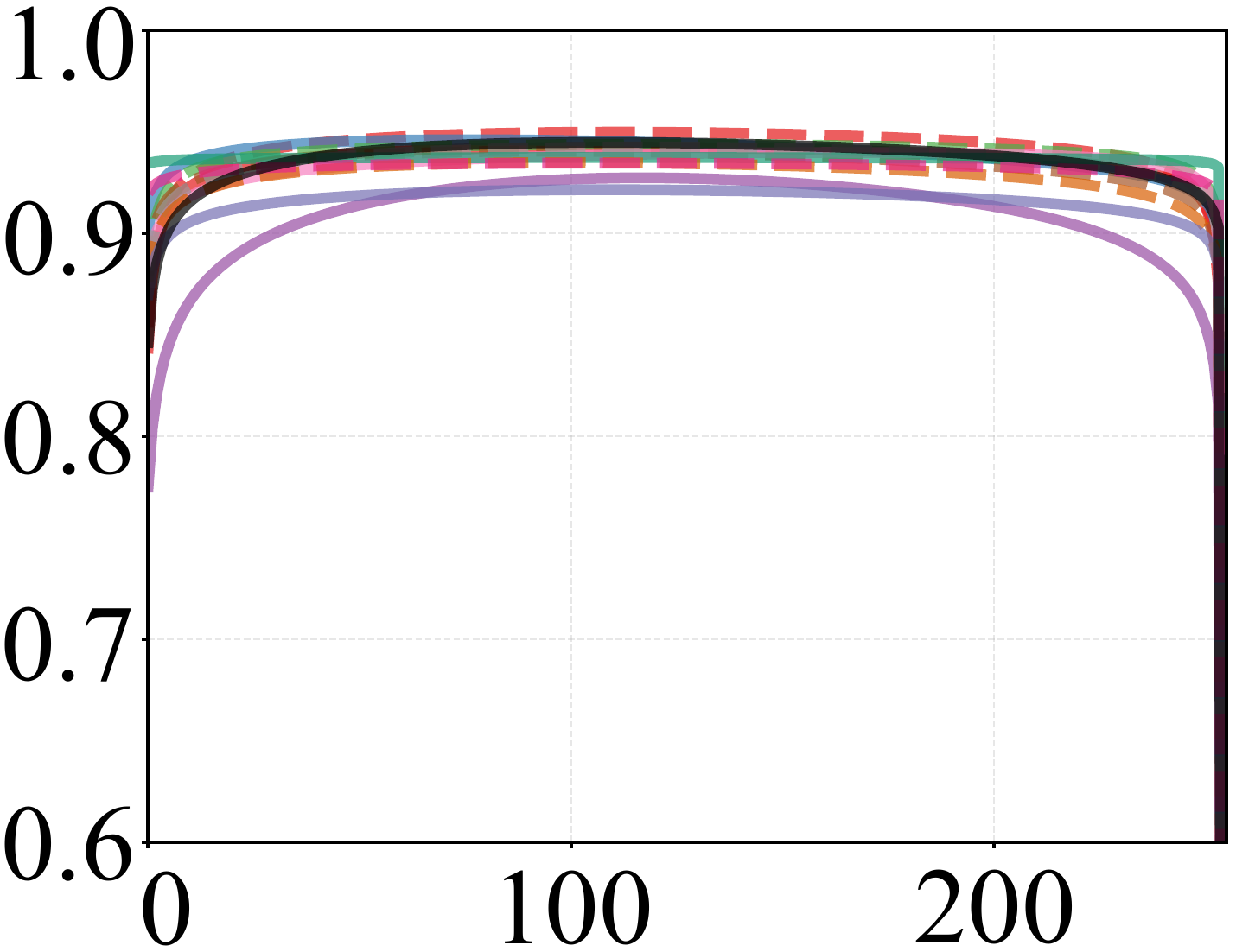}};
        \node[rotate=90] at ($(fig4.west)+(-0.05,0.08)$) {\footnotesize F-measure};
        \node at ($(fig4.south)+(0.10,-0.02)$) {\footnotesize Threshold};
        \node[inner sep=1pt] at (1.85,1.3) {HKU-IS};
    \end{scope}
    
    \begin{scope}[xshift=14.4cm, yshift=0cm]
        \node[above right] (fig5) at (0,0){\includegraphics[height=0.135\linewidth]{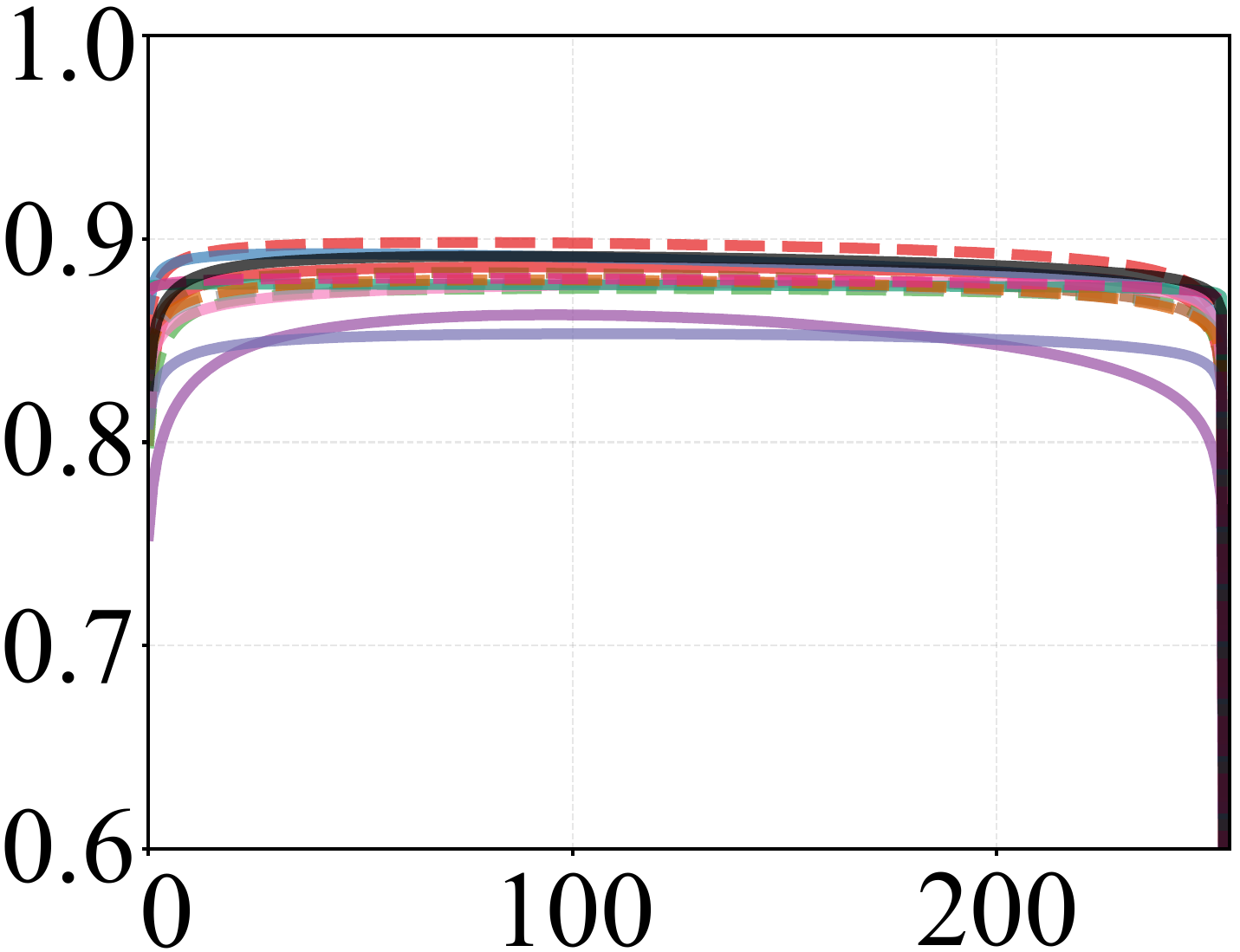}};
        \node at ($(fig5.south)+(0.05,-0.02)$) {\footnotesize Threshold};
        \node[rotate=90] at ($(fig5.west)+(-0.1,0.08)$) {\footnotesize F-measure};
        \node[inner sep=1pt] at (1.85,1.3) {PASCAL-S};
    \end{scope}
    
  \end{tikzpicture}

  \begin{tikzpicture}
    \definecolor{dgncolor}{HTML}{E41A1C}      
    \definecolor{sam2unetcolor}{HTML}{377EB8} 
    \definecolor{mdsamcolor}{HTML}{4DAF4A}    
    \definecolor{vstcolor}{HTML}{984EA3}      
    \definecolor{vstppcolor}{HTML}{A65628}    
    \definecolor{swinsodcolor}{HTML}{F781BF}  
    \definecolor{dcnetcolor}{HTML}{999999}    
    \definecolor{bbrfcolor}{HTML}{1B9E77}     
    \definecolor{iconcolor}{HTML}{D95F02}     
    \definecolor{rmfdcolor}{HTML}{7570B3}     
    \definecolor{te7color}{HTML}{E7298A}      
    \definecolor{birefnetcolor}{HTML}{000000} 
    
    \draw[te7color, line width=1.5pt, opacity=0.7, dashed] (-4.5,0.6) -- (-4.0,0.6);
    \node[right, text=black, font=\footnotesize] at (-4.0,0.6) {TE7 \cite{tracer}};
    \draw[bbrfcolor, line width=1.5pt, opacity=0.7] (-2.4,0.6) -- (-1.9,0.6);
    \node[right, text=black, font=\footnotesize] at (-1.9,0.6) {BBFR \cite{bbrf}};
    \draw[dcnetcolor, line width=1.5pt, opacity=0.7, dashed] (-0.3,0.6) -- (0.2,0.6);
    \node[right, text=black, font=\footnotesize] at (0.2,0.6) {DC-Net \cite{dcnet}};
    \draw[rmfdcolor, line width=1.5pt, opacity=0.7] (2.1,0.6) -- (2.6,0.6);
    \node[right, text=black, font=\footnotesize] at (2.6,0.6) {RMFDNet \cite{zhou2025rmfdnet}};
    \draw[vstcolor, line width=1.5pt, opacity=0.7] (4.8,0.6) -- (5.3,0.6);
    \node[right, text=black, font=\footnotesize] at (5.3,0.6) {VST \cite{Liu_2021_ICCV}};
    \draw[iconcolor, line width=1.5pt, opacity=0.7, dashed] (6.9,0.6) -- (7.4,0.6);
    \node[right, text=black, font=\footnotesize] at (7.4,0.6) {ICON-S \cite{icon}};
    \draw[vstppcolor, line width=1.5pt, opacity=0.7, dashed] (9.4,0.6) -- (9.9,0.6);
    \node[right, text=black, font=\footnotesize] at (9.9,0.6) {VST++ \cite{10497889}};
    \draw[swinsodcolor, line width=1.5pt, opacity=0.7] (-4.5,0.1) -- (-4.0,0.1);
    \node[right, text=black, font=\footnotesize] at (-4.0,0.1) {SwinSOD \cite{WU2024105039}};
    \draw[mdsamcolor, line width=1.5pt, opacity=0.7, dashed] (-1.8,0.1) -- (-1.3,0.1);
    \node[right, text=black, font=\footnotesize] at (-1.3,0.1) {MDSAM \cite{mdsam}};
    \draw[sam2unetcolor, line width=1.5pt, opacity=0.7] (0.8,0.1) -- (1.3,0.1);
    \node[right, text=black, font=\footnotesize] at (1.3,0.1) {Sam2unet \cite{sam2unet}};
    \draw[birefnetcolor, line width=1.5pt, opacity=0.7] (3.5,0.1) -- (4.0,0.1);
    \node[right, text=black, font=\footnotesize] at (4.0,0.1) {BiRefNet \cite{birefnet}};
    \draw[dgncolor, line width=1.5pt, opacity=0.7] (6.3,0.1) -- (6.8,0.1);
    \node[right, text=black, font=\footnotesize] at (6.8,0.1) {DGN-B (ours)};
    \draw[dgncolor, line width=1.5pt, opacity=0.7, dashed] (9.0,0.1) -- (9.5,0.1);
    \node[right, text=black, font=\footnotesize] at (9.5,0.1) {DGN-L (ours)};
    
    \draw[black, line width=0.5pt] (-5.0,-0.15) rectangle (11.8,0.85);
  \end{tikzpicture}
  \caption{Precision-recall and F-measure curves of our method and competing SOTA methods on five popular salient object detection benchmarks. Our DGN variants (shown in red) consistently outperform or match existing methods across all datasets.}
  \label{fig13}
\end{figure*}

\begin{table}[!t]
\centering
\definecolor{mygreen}{RGB}{0,102,0}
\definecolor{myred}{RGB}{153,0,0}
\definecolor{myorange}{RGB}{204,106,0}
\caption{: Quantitative comparison of model complexity, efficiency, and average performance across five benchmark SOD datasets. The best results are in \textcolor{myred}{red} marked \textbf{in bold}, second best in \textcolor{myorange}{orange} marked \textbf{in bold}, and third best in \textcolor{mygreen}{green} marked \textbf{in bold}. MAE: lower is better ($\downarrow$); other metrics: higher is better ($\uparrow$).}
\label{tab:average_performance}
\renewcommand{\arraystretch}{0.5}
\setlength{\tabcolsep}{0.7pt}
\begin{tabular}{@{}l|cccc|cccc@{}}
\toprule
\textbf{Method} & Size & 
Params & 
FLOPs & 
FPS & MAE & \textbf{$F_\beta^{\max}$} & \textbf{$S_m$} & \textbf{$E_m$} \\
\midrule
\multicolumn{9}{c}{\textit{CNN-based Methods}} \\
\midrule
CaGNet (2020) \cite{cagnet} & 480$^2$ & -- & -- & -- & .0378 & .8968 & .8930 & .9316 \\
DFI (2020) \cite{dfi} & 224$^2$ & 29.61 & 11.31 & 103 & .0450 & .9046 & .8876 & .9134 \\
GateNet (2020) \cite{gatenet}  & 384$^2$ & 128.63 & 162.13 & 113 & .0394 & .9040 & .8942 & .9236 \\
MINet (2020) \cite{minet} & 320$^2$ & 162.38 & 87.10 & 61 & .0438 & .8984 & .8834 & .9150 \\
LDF (2020) \cite{ldf} & 352$^2$ & 25.15 & 15.57 & 135 & .0416 & .9052 & .8874 & .9188 \\
EDN (2020) \cite{9756227} & 384$^2$ & 21.83 & 20.42 & 155 & .0414 & .8966 & .8914 & .9294 \\
TE7 (2022) \cite{tracer} & 640$^2$ & 66.27 & 10.17 & 7 & .0322 & .9206 & .9058 & .9384 \\
MENet (2023) \cite{menet} & 354$^2$ & -- & -- & -- & .0340 & .9136 & .8964 & .9260 \\
DC-Net (2023) \cite{dcnet} & 384$^2$ & 509.61 & 211.27 & 41 & .0310 & .9258 & .9150 & .9396 \\
BBRF (2023) \cite{bbrf}  & 352$^2$ & 74.00 & 67.02 & 67 & .0320 & .9094 & .9038 & .9412 \\
ELSA-Net (2024) \cite{10155248} & 320$^2$ & 31.92 & 21.77 & -- & .0396 & .8814 & -- & .9324 \\
PAM (2025) \cite{pam} & 224$^2$ & -- & -- & -- & .0368 & .9056 & .9048 & .9392 \\
RMFDNet (2025) \cite{zhou2025rmfdnet} & 352$^2$ & 26.62 & 36.70 & 87 & .0414 & .9032 & .8914 & .9194 \\
\midrule
\multicolumn{9}{c}{\textit{Transformer-based Methods}} \\
\midrule
VST (2021) \cite{Liu_2021_ICCV} & 224$^2$ & 44.48 & 23.18 & 77 & .0436 & .9026 & .8956 & .9190 \\
ICON (2022) \cite{icon} & 384$^2$ & 92.15 & 52.80 & 66 & .0334 & .9194 & .9088 & .9392 \\
VST++ (2023) \cite{10497889} & 320$^2$ & 112.23 & 69.30 & 37 & .0318 & .9240 & .9130 & .9412 \\
SwinSOD (2024) \cite{WU2024105039} & 384$^2$ & 65.70 & 51.68 & 70 & .0316 & .9264 & .9096 & .9404 \\
MDSAM (2024) \cite{mdsam} & 512$^2$ & 100.21 & 123.44 & 42 & .0310 & \textcolor{mygreen}{\textbf{.9336}} & .9138 & .9424 \\
Sam2unet (2024) \cite{sam2unet} & 352$^2$ & 216.40 & 128.40 & 43 & \textcolor{myorange}{\textbf{.0282}} & .9326 & \textcolor{mygreen}{\textbf{.9204}} & \textcolor{myorange}{\textbf{.9480}} \\
BiRefNet (2024) \cite{birefnet} & 1024$^2$ & 214.96 & 1143.00 & 8 & \textcolor{myorange}{\textbf{.0282}} & .9328 & \textcolor{myorange}{\textbf{.9206}} & .9398 \\
SefNet (2024) \cite{Yang2024_SEF_E2E} & 660$^2$ & -- & -- & -- & .0308 & .9240 & .9184 & -- \\
PDNet (2025) \cite{Cen2025_PDDNet} & 224$^2$ & 116.00 & 18.20 & -- & .0344 & .8996 & .9060 & -- \\
CDiff (2025) \cite{10834569} & 384$^2$ & 71.70 & 60.29 & 58 & \textcolor{mygreen}{\textbf{.0290}} & .8790 & .9087 & -- \\
SegR1 (2025) \cite{segR1} & 768$^2$ & -- & -- & -- & .0293 & .9195 & .9193 & \textcolor{mygreen}{\textbf{.9465}} \\
ECF-DT (2025) \cite{wang2025novel} & 384$^2$ & 23.85 & 28.10 & -- & .0422 & .8368 & .9060 & .8970 \\
\midrule
\multicolumn{9}{c}{\textit{Our Methods}} \\
\midrule
DGN-B (Ours) & 512$^2$ & 91.92 & 102.78 & 61 & .0294 & \textcolor{myorange}{\textbf{.9384}} & \textcolor{mygreen}{\textbf{.9204}} & .9462\\
DGN-L (Ours) & 512$^2$ & 247.56 & 238.52 & 43 & \textcolor{myred}{\textbf{.0262}} & \textcolor{myred}{\textbf{.9428}} & \textcolor{myred}{\textbf{.9270}} & \textcolor{myred}{\textbf{.9516}} \\
\bottomrule
\end{tabular}
\end{table}

\section{Experiments}
\subsection{Experimental Settings}
\textbf{Datasets}. We conduct comprehensive experiments on five widely-adopted RGB-SOD benchmarks to validate the effectiveness of the proposed DualGazeNet (DGN). DUTS \cite{duts}, the largest SOD dataset, contains 10,553 training images (DUTS-TR) and 5,019 test images (DUTS-TE). DUT-OMRON \cite{duto} provides 5,168 challenging images with cluttered backgrounds. HKU-IS \cite{hkuis} comprises 4,447 images with multiple disjoint salient objects. ECSSD \cite{ecssd} contains 1,000 images with complex structural patterns. PASCAL-S \cite{pascals} includes 850 images with complex scenes and multiple salient objects. Following standard practice, we train on DUTS-TR and evaluate on all five test sets.

\textbf{Evaluation metrics}. To quantitatively evaluate and compare the performance of different saliency models, we adopt four widely-utilized evaluation metrics, including Mean Absolute Error ($\text{MAE}$) \cite{mae}, maximum F-measure ($F^{\max}_\beta$) \cite{duto}, structure-measure ($S_m$) \cite{sm}, and mean enhanced-alignment measure ($E_{m}$) \cite{em}. Additionally, we provide precision-recall (PR) curves and F-measure (FM) curves \cite{9320524} for comprehensive analysis. Note that higher values indicate better performance for $F^{\max}_\beta$, $S_m$, and $E_{m}$, while lower values are preferable for $\text{MAE}$. For curve-based evaluation, a PR curve closer to the top-right corner (1,1) indicates superior performance, and an FM curve with a larger area under the curve reflects better overall quality. Note that all percentage changes reported in this section are relative improvements, i.e., $\frac{(M_{\text{ours}} - M_{\text{baseline}})}{M_{\text{baseline}}} \times 100\%$ for larger-is-better metrics and $\frac{({M}_{\text{baseline}} - {M}_{\text{ours}})}{{M}_{\text{baseline}}} \times 100\%$ for lower-is-better metrics.

\textbf{Efficiency evaluation}. To assess the computational efficiency of our model, we adopt three metrics, following recent works \cite{11072100,10339864}, including the number of parameters (Params, in millions), floating-point operations (FLOPs, in giga), and inference speed (FPS, frames per second). Params quantifies the total model parameters and serves as an indicator of storage requirements. FLOPs measures the computational complexity by counting floating-point operations. FPS directly reflects the inference speed by measuring the number of frames processed per second.

\textbf{Implementation details}. We implement our method utilizing the PyTorch framework on a NVIDIA RTX 5090 GPU with 32GB of memory. For the image encoder, we employ two backbone variants: Hiera-B and Hiera-L, both initialized with pretrained weights from the official Hiera model. All other components, including the adapter modules and decoder, are randomly initialized. Input images are resized to $512 \times 512$ pixels, with batch sizes of 16 and 8 for Hiera-B and Hiera-L, respectively. During training, the encoder backbone remains frozen to preserve its pretrained hierarchical representations, while only the adapter modules and decoder are fine-tuned. We optimize these trainable parameters utilizing the AdamW optimizer with a weight decay of $1\times10^{-3}$ and an initial learning rate of $1\times10^{-4}$. The training process spans 100 epochs, including a 5-epoch linear warm-up phase at the beginning.

\subsection{Comparison with State-of-the-Art Methods} \label{SOTA}
In this section, we compare the proposed DGN-B and DGN-L with 25 state-of-the-art models, including 13 CNN-based methods, i.e., CaGNet \cite{cagnet}, DFI \cite{dfi}, GateNet \cite{gatenet}, MINet \cite{minet}, LDF \cite{ldf}, EDN \cite{9756227}, TE \cite{tracer}, MENet \cite{menet}, DC-Net \cite{dcnet}, BBRF \cite{bbrf}, ELSA-Net \cite{10155248}, PAM \cite{pam}, RMFDNet \cite{zhou2025rmfdnet} and 12 Transformer-based methods, i.e., VST \cite{Liu_2021_ICCV}, ICON \cite{icon}, VST++ \cite{10497889}, SwinSOD \cite{WU2024105039}, MDSAM \cite{mdsam}, SAM2UNet \cite{sam2unet}, BirefNet \cite{birefnet}, PDNet \cite{Cen2025_PDDNet}, SefNet \cite{Yang2024_SEF_E2E}, CamoDiffusion \cite{10834569}, SeqR1 \cite{segR1}, and ECF-DT \cite{wang2025novel}. 

We categorize these methods according to their primary characteristics. Methods prioritizing the extraction, utilization, and fusion of multi-level features include GateNet \cite{gatenet}, MINet \cite{minet}, EDN \cite{9756227}, MENet \cite{menet}, and BBRF \cite{bbrf}. Multi-component approaches that leverage boundary cues or multi-task learning frameworks encompass DFI \cite{dfi}, DC-Net \cite{dcnet}, ELSA-Net \cite{10155248}, and PAM \cite{pam}. Methods that explicitly correct/compensate imperfect features comprise CaGNet \cite{cagnet}, LDF \cite{ldf}, TE \cite{tracer}, RMFDNet \cite{zhou2025rmfdnet}, and ICON \cite{icon}. Pure Transformer architectures are represented by VST \cite{Liu_2021_ICCV} and VST++ \cite{10497889}, and CNN-Transformer hybrid architectures include SwinSOD \cite{WU2024105039}, MDSAM \cite{mdsam}, Sam2UNet \cite{sam2unet}, BirefNet \cite{birefnet}, and ECF-DT \cite{wang2025novel}. Query-learning mechanisms are incorporated in PDNet \cite{Cen2025_PDDNet} and SefNet \cite{Yang2024_SEF_E2E}. Finally, diffusion-based methods, such as CamoDiffusion \cite{10834569}, treat SOD as a conditional denoising problem, whereas SegR1 \cite{segR1} utilizes reinforcement learning to generate prompts for a transformer-based segmenter. For fair comparisons, the prediction results of all methods are obtained either directly from the original authors or by running their publicly released source codes with pretrained models.

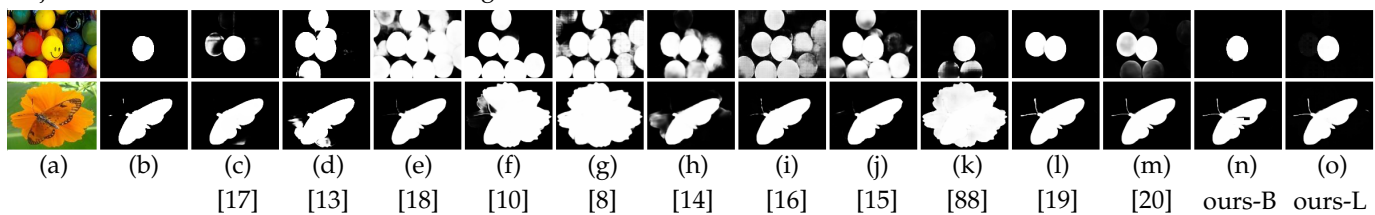
\begin{figure*}[!t]
  \centering
  \definecolor{mygreen}{RGB}{0,102,0}
  \definecolor{myred}{RGB}{153,0,0}
  \definecolor{myblue}{RGB}{0,122,122}
  \definecolor{myorange}{RGB}{204,106,0}
  \begin{tikzpicture}%
    \begin{scope}[xshift=0cm, yshift=0cm ]
        \node[above right] (fig63) at (0.0,0.4){\hspace{-0.11cm}\includegraphics[width=1.00\textwidth]{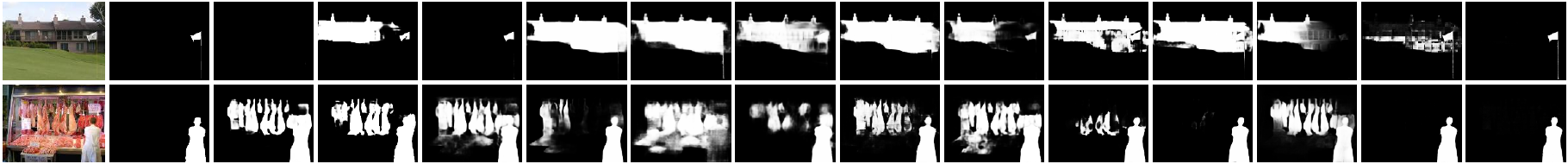}};
        \node[inner sep=1pt] at (1.95,3.6-0.95) {Object near the boundary};
        \draw[thick, black] ([xshift=0.05cm, yshift=0.4cm]fig63.north west) -- ([xshift=-0.12cm, yshift=0.4cm]fig63.north east);
    \end{scope}
    \begin{scope}[xshift=0cm, yshift=-2.6cm ]
        \node[above right] (fig63) at (0.0,0.4){\hspace{-0.11cm}\includegraphics[width=1.00\textwidth]{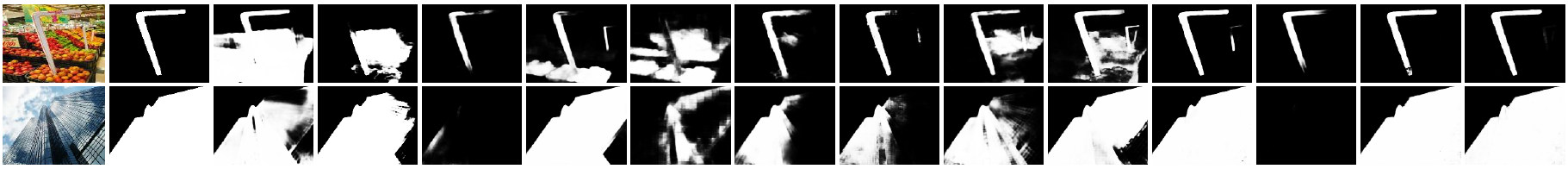}};
        \node[inner sep=1pt] at (1.00,3.6-0.95) {Large Object};
        \draw[thick, black] ([xshift=0.05cm, yshift=0.4cm]fig63.north west) -- ([xshift=-0.12cm, yshift=0.4cm]fig63.north east);
    \end{scope}
    \begin{scope}[xshift=0cm, yshift=-5.2cm ]
        \node[above right] (fig63) at (0.0,0.4){\hspace{-0.11cm}\includegraphics[width=1.00\textwidth]{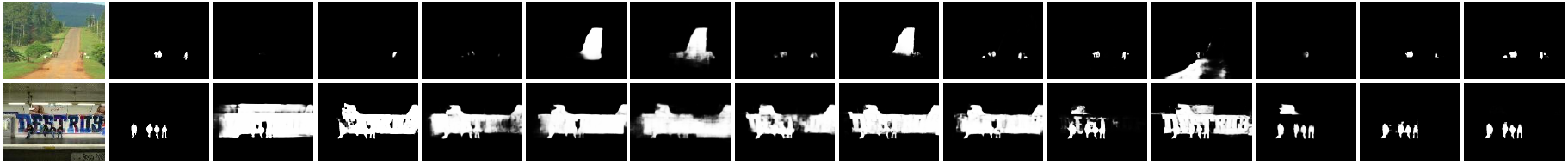}};
        \node[inner sep=1pt] at (1.68,3.6-0.95) {Multiple small objects};
        \draw[thick, black] ([xshift=0.05cm, yshift=0.4cm]fig63.north west) -- ([xshift=-0.12cm, yshift=0.4cm]fig63.north east);
    \end{scope}
    \begin{scope}[xshift=0cm, yshift=-7.8cm ]
        \node[above right] (fig63) at (0.0,0.4){\hspace{-0.11cm}\includegraphics[width=1.00\textwidth]{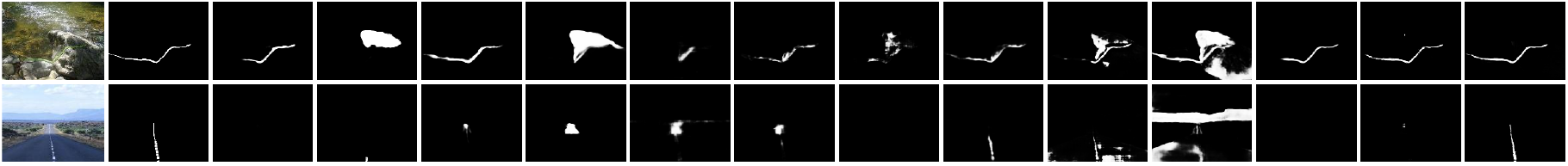}};
        \node[inner sep=1pt] at (1.34,3.6-0.95) {Elongated objects};
        \draw[thick, black] ([xshift=0.05cm, yshift=0.4cm]fig63.north west) -- ([xshift=-0.12cm, yshift=0.4cm]fig63.north east);
    \end{scope}
    \begin{scope}[xshift=0cm, yshift=-10.4cm ]
        \node[above right] (fig63) at (0.0,0.4){\hspace{-0.11cm}\includegraphics[width=1.00\textwidth]{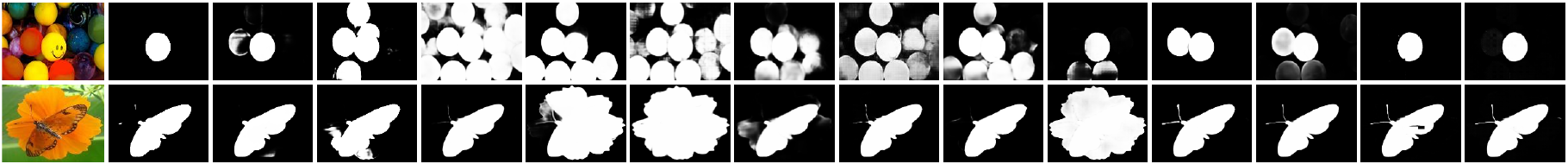}};
        \node[inner sep=1pt] at (3.80,3.6-0.95) {Objects within low-contrast cluttered backgrounds};
        \draw[thick, black] ([xshift=0.05cm, yshift=0.4cm]fig63.north west) -- ([xshift=-0.12cm, yshift=0.4cm]fig63.north east);
        \node[inner sep=1pt] at (0.65,0.3) {(a)};
        \node[inner sep=1pt] at (1.85,0.3) {(b)};
        \node[inner sep=1pt] at (3.1,0.3) {(c)};
        \node[inner sep=1pt] at (3.05,-0.15) {\cite{tracer}};
        \node[inner sep=1pt] at (4.3,0.3) {(d)};
        \node[inner sep=1pt] at (4.25,-0.15) {\cite{bbrf}};
        \node[inner sep=1pt] at (5.5,0.3) {(e)};
        \node[inner sep=1pt] at (5.45,-0.15) {\cite{dcnet}};
        \node[inner sep=1pt] at (6.7,0.3) {(f)};
        \node[inner sep=1pt] at (6.65,-0.15) {\cite{zhou2025rmfdnet}};
        \node[inner sep=1pt] at (7.95,0.3) {(g)};
        \node[inner sep=1pt] at (7.90,-0.15) {\cite{Liu_2021_ICCV}};
        \node[inner sep=1pt] at (9.15,0.3) {(h)};
        \node[inner sep=1pt] at (9.10,-0.15) {\cite{icon}};
        \node[inner sep=1pt] at (10.4,0.3) {(i)};
        \node[inner sep=1pt] at (10.35,-0.15) {\cite{10497889}};
        \node[inner sep=1pt] at (11.6,0.3) {(j)};
        \node[inner sep=1pt] at (11.55,-0.15) {\cite{WU2024105039}}; 
        \node[inner sep=1pt] at (12.8,0.3) {(k)};
        \node[inner sep=1pt] at (12.75,-0.15) {\cite{mdsam}};
        \node[inner sep=1pt] at (14.0,0.3) {(l)};
        \node[inner sep=1pt] at (13.95,-0.15) {\cite{sam2unet}};
        \node[inner sep=1pt] at (15.25,0.3) {(m)};
        \node[inner sep=1pt] at (15.20,-0.15) {\cite{birefnet}};
        \node[inner sep=1pt] at (16.45,0.3) {(n)};
        \node[inner sep=1pt ] at (16.42,-0.15) {ours-B};
        \node[inner sep=1pt] at (17.65,0.3) {(o)};
        \node[inner sep=1pt] at (17.62,-0.15) {ours-L};
    \end{scope}
  \end{tikzpicture}
  \caption{Qualitative comparison of DGN-B/L with eleven state-of-the-art methods across five challenging scenarios on SOD. From (a) to (o): input image, ground truth, TE7 \cite{tracer}, BBRF \cite{bbrf}, DC-Net \cite{dcnet}, RMFDNet \cite{zhou2025rmfdnet}, VST \cite{Liu_2021_ICCV}, ICON-S \cite{icon}, VST++ \cite{10497889}, SwinSOD  \cite{WU2024105039}, MDSAM \cite{mdsam}, Sam2UNet \cite{sam2unet}, BiRefNet \cite{birefnet}, DGN-B (ours), and DGN-L (ours). }
  \label{fig12}
\end{figure*}

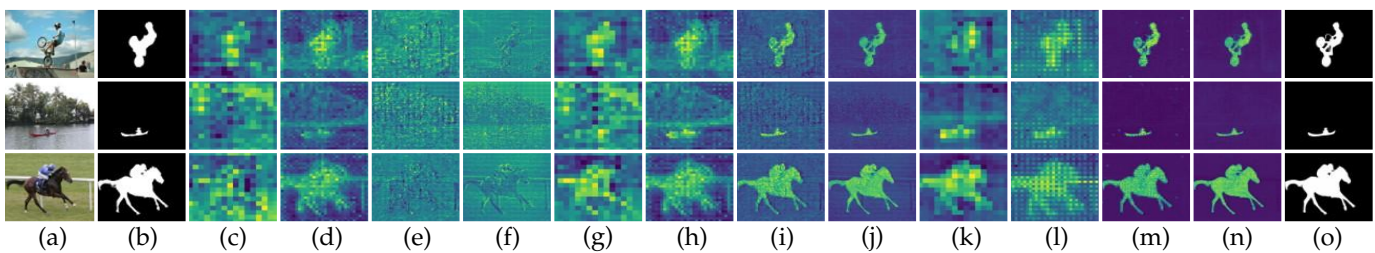
\begin{figure*}[!t]
  \centering
  \definecolor{mygreen}{RGB}{0,102,0}
  \definecolor{myred}{RGB}{153,0,0}
  \definecolor{myblue}{RGB}{0,122,122}
  \definecolor{myorange}{RGB}{204,106,0}
  \begin{tikzpicture}%
    \begin{scope}[xshift=0cm, yshift=0cm ]
        \node[above right] (fig63) at (0.0,0.4){\hspace{-0.1cm}\includegraphics[width=1.000\textwidth]{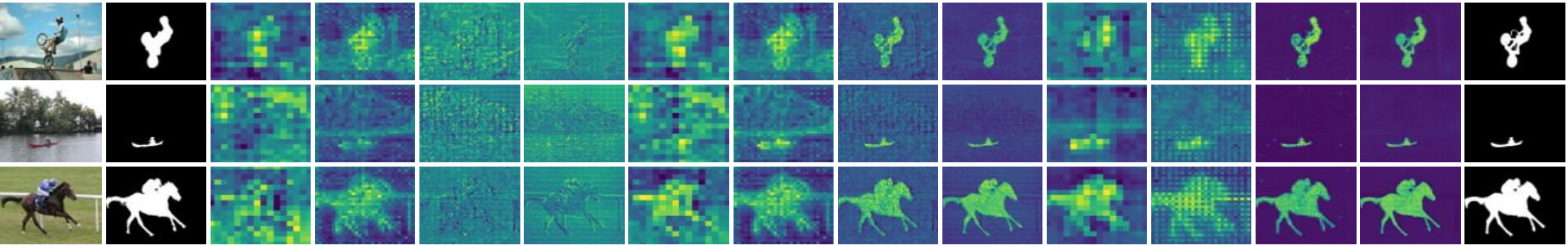}};
        \node[inner sep=1pt] at (0.65,0.3) {(a)};
        \node[inner sep=1pt] at (1.85,0.3) {(b)};
        \node[inner sep=1pt] at (3.05,0.3) {(c)};
        \node[inner sep=1pt] at (4.3,0.3) {(d)};
        \node[inner sep=1pt] at (5.5,0.3) {(e)};
        \node[inner sep=1pt] at (6.7,0.3) {(f)};
        \node[inner sep=1pt] at (7.9,0.3) {(g)};
        \node[inner sep=1pt] at (9.15,0.3) {(h)};
        \node[inner sep=1pt] at (10.35,0.3) {(i)};
        \node[inner sep=1pt] at (11.55,0.3) {(j)};
        \node[inner sep=1pt] at (12.8,0.3) {(k)};
        \node[inner sep=1pt] at (14.0,0.3) {(l)};
        \node[inner sep=1pt] at (15.25,0.3) {(m)};
        \node[inner sep=1pt] at (16.35+0.05,0.3) {(n)};
        \node[inner sep=1pt] at (17.5+0.05+0.05,0.3) {(o)};
    \end{scope}
  \end{tikzpicture}
  \caption{Visualization of features in DualGazeNet. From left to right: (a) input image; (b) ground truth GT; (c)--(f) hierarchical features $F_1$ to $F_4$ from the first gaze; (g)--(j) hierarchical features $F^{'}_1$ to $F^{'}_4$ from the second gaze guided by cortical queries; (k)--(n) per-level difference maps Diff. $F^-_1$ to Diff. $F^-_4$; (0) final prediction from DGN-L. At each level, the feature maps come from the same channel index across the two gazes, enabling a like-for-like comparison. The second gaze shows stronger focus on salient regions and clearer object boundaries, and the difference maps expose the semantic guidance provided by the cortical queries.}
  \label{fig9}
\end{figure*}

\textbf{Quantitative evaluation.} Table \ref{tab:sod_comparison} summarizes the quantitative comparison results across five datasets utilizing four evaluation metrics, and Table \ref{tab:average_performance} presents the average performance over these datasets. The results in Table \ref{tab:sod_comparison} demonstrate that out of 20 evaluation metrics in total, the DGN-L model with a heavy backbone attains the best performance on 19 metrics, and the DGN-B model with a lightweight backbone also achieves top performance on 5 metrics. The remaining metrics also demonstrate the competitive performance of our models. As shown in Table \ref{tab:average_performance}, DGN-L achieves the best performance across all averaged metrics. Specifically, DGN-L outperforms the second-best method BiRefNet \cite{birefnet} by 7.09$\%$ in MAE and 0.695$\%$ in $S_{m}$, despite BiRefNet employing high-resolution inputs (1024×1024), a complex architecture, and multiple carefully-tuned loss functions.  In terms of $F_{\beta}^{\max}$, DGN-L surpasses the third-ranked MDSAM \cite{mdsam} by 0.985$\%$, even though MDSAM leverages large-scale pretrained foundation models. Additionally, for $E_{m}$, DGN-L outperforms the second-best method Sam2unet \cite{sam2unet} by 0.380$\%$, which adopts a CNN-Transformer hybrid architecture. Furthermore, DGN-B also demonstrates comparable average performance to both BiRefNet \cite{birefnet} and Sam2UNet \cite{sam2unet}, achieving second and third place in $F_{\beta}^{\max}$ and $S_{m}$, respectively. Beyond the numerical metrics, Fig. \ref{fig13} presents the PR and FM curves, which further validate the superiority of DGN. Our method consistently exhibits curves that are positioned more outward, indicating higher precision across varying recall levels in PR curves and superior F-measure values across different thresholds in FM curves, thereby demonstrating more robust and accurate predictions. These results demonstrate that by simply emulating the dual biological principles of human vision, namely robust representation learning and dual-pathway processing, our approach achieves superior performance without requiring complex multi-component frameworks, boundary cues, multi-task learning strategies, pixel-channel attention mechanisms, CNN-Transformer hybrid architectures, or elaborate loss function designs.

\begin{table*}[!t]
\centering
  \definecolor{mygreen}{RGB}{0,102,0}
  \definecolor{myred}{RGB}{153,0,0}
  \definecolor{myblue}{RGB}{0,122,122}
  \definecolor{myorange}{RGB}{204,106,0}
\caption{: Quantitative comparison on four benchmark COD datasets. The best results are in \textcolor{myred}{red} marked \textbf{in bold}, second best in \textcolor{myorange}{orange} marked \textbf{in bold}, and third best in \textcolor{mygreen}{green} marked \textbf{in bold}. MAE: lower is better ($\downarrow$); other metrics: higher is better ($\uparrow$). Numbers in parentheses indicate dataset sizes. Please refer to the text for more details.}
\label{tab:cod_comparison}
\renewcommand{\arraystretch}{0.8}
\setlength{\tabcolsep}{1pt}
\begin{tabular*}{\textwidth}{@{\extracolsep{\fill}}l|cccc|cccc|cccc|cccc|cccc@{}}
\toprule
\multirow{2}{*}{\textbf{Method}} &
\multicolumn{4}{c|}{\textbf{CAMO (250)}} &
\multicolumn{4}{c|}{\textbf{COD10K (2026)}} &
\multicolumn{4}{c|}{\textbf{NC4K (4121)}} & 
\multicolumn{4}{c|}{\textbf{CHAMELEON (76)}} & 
\multicolumn{4}{c}{\textbf{AVERAGE}} \\
\cmidrule(lr){2-5} \cmidrule(lr){6-9} \cmidrule(lr){10-13} \cmidrule(lr){14-17} \cmidrule(lr){18-21}
& MAE & $F_\beta^{\max}$ & $S_m$ & $E_m$ 
& MAE & $F_\beta^{\max}$ & $S_m$ & $E_m$ 
& MAE & $F_\beta^{\max}$ & $S_m$ & $E_m$ 
& MAE & $F_\beta^{\max}$ & $S_m$ & $E_m$ 
& MAE & $F_\beta^{\max}$ & $S_m$ & $E_m$ \\
\midrule
Zoomnet (2022) \cite{zoomnet} & .066 & .851 & .820 & .877 & .029 & .822 & .838 & .888 & .043 & .867 & .853 & .896 & .023 & .895 & .902 & .943 & .040 & .859 & .853 & .901 \\
MRRNet (2023) \cite{10180211} & .070 & .850 & .826 & .880 & .032 & .807 & .835 & .901 & .044 & .861 & .857 & .906 & .029 & .878 & .891 & .936 & .044 & .849 & .852 & .906 \\
FSPNet (2023) \cite{huang2023feature} & .050 & .884 & .856 & .899 & .026 & .826 & .851 & .895 & .035 & .885 & .879 & .915 & .023 & .900 & .908 & .943 & .034 & .874 & .874 & .913 \\
DGNet (2023) \cite{ji2023deep} & .057 & .856 & .839 & .901 & .033 & .789 & .822 & .896 & .042 & .860 & .857 & .911 & .029 & .877 & .890 & .938 & .040 & .846 & .852 & .912 \\
CamoDiffusion (2023) \cite{chen2024camodiffusion} & \textcolor{myorange}{\textbf{.042}} & \textcolor{mygreen}{\textbf{.890}} & \textcolor{myorange}{\textbf{.878}} & \textcolor{mygreen}{\textbf{.936}} & \textcolor{myorange}{\textbf{.019}} & \textcolor{myorange}{\textbf{.862}} & \textcolor{myorange}{\textbf{.883}} & \textcolor{mygreen}{\textbf{.943}} & \textcolor{myred}{\textbf{.028}} & \textcolor{mygreen}{\textbf{.897}} & \textcolor{myorange}{\textbf{.895}} & \textcolor{mygreen}{\textbf{.942}} & -- & -- & -- & -- & \textcolor{myorange}{\textbf{.030}} & .883 & .885 & .940 \\
FEDER (2023)  \cite{10203727} & .071 & .831 & .802 & .867 & .032 & .793 & .822 & .900 & .044 & .858 & .847 & .907 & .030 & .879 & .887 & .946 & .044 & .840 & .840 & .905 \\
LSR (2023) \cite{10007893} & .080 & .811 & .787 & .838 & .037 & .765 & .804 & .880 & .048 & .843 & .840 & .895 & .030 & .871 & .890 & .935 & .049 & .822 & .830 & .887 \\
RISNet (2024) \cite{risnet} & .050 & .886 & .870 & .922 & .025 & .851 & .873 & .931 & .037 & .886 & .882 & .925 & -- & -- & -- & -- & .037 & .874 & .875 & .926 \\
PRNet (2024)  \cite{10379651} & .050 & .888 & .872 & .922 & .022 & .855 & .874 & .937 & \textcolor{mygreen}{\textbf{.031}} & \textcolor{myorange}{\textbf{.899}} & .891 & .935 & \textcolor{myorange}{\textbf{.020}} & .908 & .914 & \textcolor{myorange}{\textbf{.965}} & \textcolor{mygreen}{\textbf{.031}} & \textcolor{mygreen}{\textbf{.888}} & .888 & .940 \\
CamoFormer (2024) \cite{10623294} & .046 & .887 & .872 & .929 & .023 & .845 & .869 & .932 & \textcolor{myorange}{\textbf{.030}} & .895 & \textcolor{mygreen}{\textbf{.892}} & .939 & .022 & .907 & .910 & .957 & \textcolor{myorange}{\textbf{.030}} & .884 & .886 & .939 \\
SAM-TTT (2025) \cite{yu2025sam} & \textcolor{mygreen}{\textbf{.045}} & .879 & .868 & \textcolor{myorange}{\textbf{.938}} & .024 & .851 & .874 & \textcolor{myorange}{\textbf{.944}} & .033 & .893 & .890 & \textcolor{myred}{\textbf{.946}} & -- & -- & -- & -- & .034 & .874 & .877 & \textcolor{myorange}{\textbf{.943}} \\
CGD (2025) \cite{zhang2025cgcod} & \textcolor{mygreen}{\textbf{.045}} & -- & \textcolor{mygreen}{\textbf{.876}} & .932 & .023 & -- & .872 & .935 & .032 & -- & .889 & .935 & \textcolor{mygreen}{\textbf{.021}} & -- & \textcolor{mygreen}{\textbf{.917}} & \textcolor{myorange}{\textbf{.965}} & .030 & -- & \textcolor{mygreen}{\textbf{.889}} & \textcolor{mygreen}{\textbf{.942}} \\
CTF-Net (2025) \cite{CTF-Net} & .047 & .881 & .870 & .922 & .024 & .840 & .867 & .928 & .033 & .886 & .884 & .932 & \textcolor{mygreen}{\textbf{.021}} & \textcolor{mygreen}{\textbf{.911}} & .916 & .957 & \textcolor{mygreen}{\textbf{.031}} & .880 & .884 & .935 \\
BCLNet (2025) \cite{bclnet}  & .046 & \textcolor{myorange}{\textbf{.891}} & \textcolor{mygreen}{\textbf{.876}} & .929 & \textcolor{mygreen}{\textbf{.021}} & \textcolor{mygreen}{\textbf{.858}} & \textcolor{mygreen}{\textbf{.878}} & .941 & \textcolor{mygreen}{\textbf{.031}} & \textcolor{mygreen}{\textbf{.897}} & \textcolor{mygreen}{\textbf{.892}} & .939 & \textcolor{myorange}{\textbf{.020}} & \textcolor{myorange}{\textbf{.915}} & \textcolor{myorange}{\textbf{.918}} & \textcolor{mygreen}{\textbf{.964}} & \textcolor{myorange}{\textbf{.030}} & \textcolor{myorange}{\textbf{.890}} & \textcolor{myorange}{\textbf{.891}} & \textcolor{myorange}{\textbf{.943}} \\
DGN-L (ours) & \textcolor{myred}{\textbf{.037}} & \textcolor{myred}{\textbf{.917}} & \textcolor{myred}{\textbf{.899}} & \textcolor{myred}{\textbf{.942}} & \textcolor{myred}{\textbf{.017}} & \textcolor{myred}{\textbf{.895}} & \textcolor{myred}{\textbf{.907}} & \textcolor{myred}{\textbf{.953}} & \textcolor{myred}{\textbf{.028}} & \textcolor{myred}{\textbf{.913}} & \textcolor{myred}{\textbf{.909}} & \textcolor{myorange}{\textbf{.945}} & \textcolor{myred}{\textbf{.017}} & \textcolor{myred}{\textbf{.932}} & \textcolor{myred}{\textbf{.932}} & \textcolor{myred}{\textbf{.968}} & \textcolor{myred}{\textbf{.025}} & \textcolor{myred}{\textbf{.914}} & \textcolor{myred}{\textbf{.912}} & \textcolor{myred}{\textbf{.952}} \\
\bottomrule
\end{tabular*}
\end{table*}

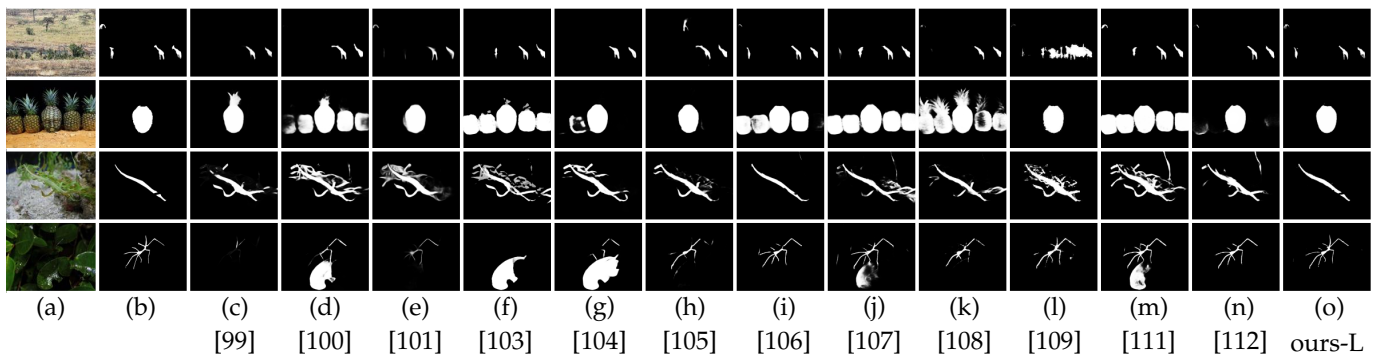
\begin{figure*}[!t]
  \centering
  \definecolor{mygreen}{RGB}{0,102,0}
  \definecolor{myred}{RGB}{153,0,0}
  \definecolor{myblue}{RGB}{0,122,122}
  \definecolor{myorange}{RGB}{204,106,0}
  \begin{tikzpicture}%
    \begin{scope}[xshift=0cm, yshift=0cm ]
        \node[above right] (fig63) at (0.0,0.4){\hspace{-0.1cm}\includegraphics[width=1.00\textwidth]{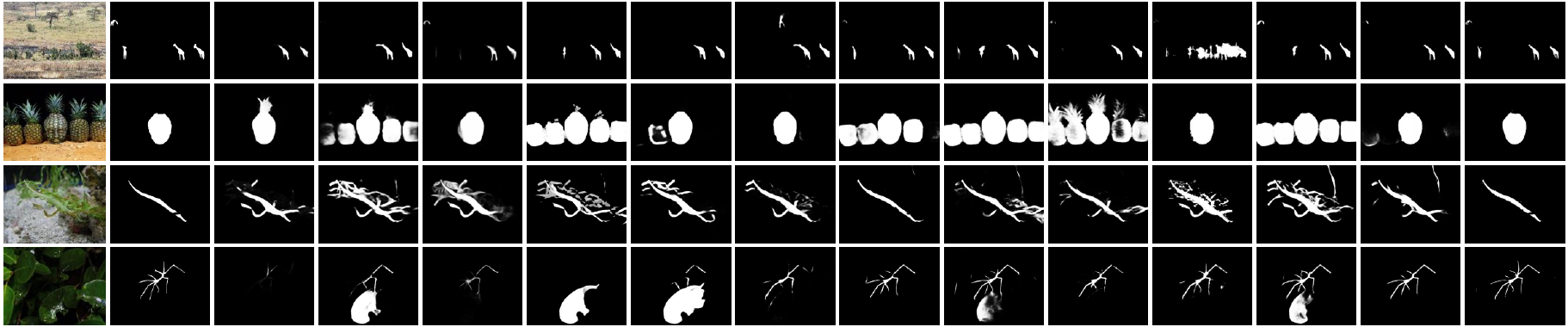}};
        \node[inner sep=1pt] at (0.65,0.3) {(a)};
        \node[inner sep=1pt] at (1.85,0.3) {(b)};
        \node[inner sep=1pt] at (3.1,0.3) {(c)};
        \node[inner sep=1pt] at (3.05,-0.15) {\cite{zoomnet}};
        \node[inner sep=1pt] at (4.3,0.3) {(d)};
        \node[inner sep=1pt] at (4.25,-0.15) {\cite{10180211}};
        \node[inner sep=1pt] at (5.5,0.3) {(e)};
        \node[inner sep=1pt] at (5.45,-0.15) {\cite{huang2023feature}};
        \node[inner sep=1pt] at (6.7,0.3) {(f)};
        \node[inner sep=1pt] at (6.65,-0.15) {\cite{chen2024camodiffusion}};
        \node[inner sep=1pt] at (7.95,0.3) {(g)};
        \node[inner sep=1pt] at (7.90,-0.15) {\cite{10203727}};
        \node[inner sep=1pt] at (9.15,0.3) {(h)};
        \node[inner sep=1pt] at (9.10,-0.15) {\cite{10007893}};
        \node[inner sep=1pt] at (10.4,0.3) {(i)};
        \node[inner sep=1pt] at (10.35,-0.15) {\cite{risnet}};
        \node[inner sep=1pt] at (11.6,0.3) {(j)};
        \node[inner sep=1pt] at (11.55,-0.15) {\cite{10379651}}; 
        \node[inner sep=1pt] at (12.8,0.3) {(k)};
        \node[inner sep=1pt] at (12.75,-0.15) {\cite{10623294}};
        \node[inner sep=1pt] at (14.0,0.3) {(l)};
        \node[inner sep=1pt] at (13.95,-0.15) {\cite{yu2025sam}};
        \node[inner sep=1pt] at (15.25,0.3) {(m)};
        \node[inner sep=1pt] at (15.20,-0.15) {\cite{CTF-Net}};
        \node[inner sep=1pt] at (16.45,0.3) {(n)};
        \node[inner sep=1pt ] at (16.42,-0.15) {\cite{bclnet}};
        \node[inner sep=1pt] at (17.65,0.3) {(o)};
        \node[inner sep=1pt] at (17.62,-0.15) {ours-L};
        
    \end{scope}
  \end{tikzpicture}
  \caption{Qualitative comparison of DGN-L with twelve state-of-the-art methods on COD. From (a) to (o): input image, ground truth, Zoomnet \cite{zoomnet}, MRRNet \cite{10180211}, FSPNet \cite{huang2023feature}, CamoDiffusion \cite{chen2024camodiffusion}, FEDER \cite{10203727}, LSR \cite{10007893}, RISNet \cite{risnet}, PRNet  \cite{10379651}, CamoFormer \cite{10623294}, SAM-TTT \cite{yu2025sam}, CTF-Net \cite{CTF-Net}, BCLNet \cite{bclnet} and DGN-L (ours).}
  \label{fig10}
\end{figure*}

\textbf{Qualitative result.}  Fig. \ref{fig12} presents several representative examples for visual comparison between our method and current state-of-the-art approaches. These examples feature complex backgrounds with salient objects exhibiting challenging characteristics, including objects near boundaries, large objects, multiple small objects, elongated objects, and objects embedded in low-contrast cluttered scenes. The results reveal that existing methods struggle to handle all these challenging scenarios simultaneously, with most producing unsatisfactory predictions. In contrast, our method achieves much improvement over other state-of-the-art approaches. This superiority stems from the cortical queries that learn and localize the semantic positions of salient objects within complex backgrounds, and subsequently guide the comprehensive reconstruction of hierarchical features, as shown in Fig. \ref{fig9}. Therefore, our method is capable of accurately highlighting salient regions with precise structural details in challenging scenes, while state-of-the-art methods lack such localization and guided reconstruction capabilities and thus fail to achieve comparable performance.

\begin{table*}[!t]
\centering
  \definecolor{mygreen}{RGB}{0,102,0}
  \definecolor{myred}{RGB}{153,0,0}
  \definecolor{myblue}{RGB}{0,122,122}
  \definecolor{myorange}{RGB}{204,106,0}
\caption{: Quantitative comparison on USOD10K. The best results are in \textcolor{myred}{red} marked \textbf{in bold}, second best in \textcolor{myorange}{orange} marked \textbf{in bold}, and third best in \textcolor{mygreen}{green} marked \textbf{in bold}. MAE: lower is better ($\downarrow$); other metrics: higher is better ($\uparrow$).}
\label{tab:underwater_comparison}
\renewcommand{\arraystretch}{0.6}
\setlength{\tabcolsep}{1pt}
\begin{tabular*}{\textwidth}{@{\extracolsep{\fill}}l|cccccccccccccc@{}}
\toprule
\textbf{Metric} & {TC-USOD} & HIDA-Net & CATNet & PICRNet & Dual-SAM & VST++ & iGAN & STAMF & STAMF-DP & DSPL-DE & HEHP & FSCDiff & {DGN-L}\\
  & \cite{10102831} {(23)}  & \cite{10091765} {(23)} & \cite{10179145} {(23)}  & \cite{cong2023point} {(23)} & \cite{10657195} {(24)} &  \cite{10497889} {(24)}  & \cite{10697198} {(24)} & \cite{ma2025stamf} {(25)} & \cite{ma2025stamf} {(25)} & \cite{10744585} {(25)} & \cite{11018233} {(25)} & \cite{li2025fscdiff} {(25)} & {(Ours)} \\
\midrule
MAE  & .020 & .026  & .021 & .023 & \textcolor{mygreen}{\textbf{.018}} & .047 & .021 & .023 & .020 & .020 & \textcolor{myred}{\textbf{.014}} & \textcolor{myorange}{\textbf{.017}} & \textcolor{myorange}{\textbf{.017}}\\
$F^{\max}_\beta$  & .924 & .911 & .924 & .921 & .925 & .858 & .925 & .913 & .926 & .927 & \textcolor{myorange}{\textbf{.940}} & \textcolor{mygreen}{\textbf{.933}} & \textcolor{myred}{\textbf{.950}}\\
$S_m$  & .922 & .911 & .921 & .921 & .924 & .870 & .920 & .912 & .924 & .923 & \textcolor{myorange}{\textbf{.935}} & \textcolor{mygreen}{\textbf{.932}} & \textcolor{myred}{\textbf{.936}}\\
$E_{m}$  & .968 & .957 & .966 & .963 & .964 & .921 & .968 & .957 & .968 & .968 & \textcolor{myred}{\textbf{.983}} & \textcolor{myorange}{\textbf{.971}} & \textcolor{mygreen}{\textbf{.969}}\\
\bottomrule
\end{tabular*}
\end{table*}

\begin{table*}[!t]
\centering
  \definecolor{mygreen}{RGB}{0,102,0}
  \definecolor{myred}{RGB}{153,0,0}
  \definecolor{myblue}{RGB}{0,122,122}
  \definecolor{myorange}{RGB}{204,106,0}
\caption{: Ablation results on different architectural settings across five benchmark SOD datasets. The best results are in \textcolor{myred}{red} marked \textbf{in bold}, second best in \textcolor{myorange}{orange} marked \textbf{in bold}, and third best in \textcolor{mygreen}{green} marked \textbf{in bold}. MAE: lower is better ($\downarrow$); other metrics: higher is better ($\uparrow$). Numbers in parentheses indicate dataset sizes. Please refer to the text for more details.}
\label{tab:sod_comparison222}
\renewcommand{\arraystretch}{0.8}
\setlength{\tabcolsep}{1pt}
\begin{tabular*}{\textwidth}{@{\extracolsep{\fill}}l|cccc|cccc|cccc|cccc|cccc|cccc@{}}
\toprule
\multirow{2}{*}{\textbf{Method}} &
\multicolumn{4}{c|}{\textbf{DUTS-TE (5019)}} &
\multicolumn{4}{c|}{\textbf{DUT-OMRON (5168)}} &
\multicolumn{4}{c|}{\textbf{ECSSD (1000)}} & 
\multicolumn{4}{c|}{\textbf{HKU-IS (4447)}} & 
\multicolumn{4}{c|}{\textbf{PASCAL-S (850)}} &
\multicolumn{4}{c}{\textbf{Average}} \\
\cmidrule(lr){2-5} \cmidrule(lr){6-9} \cmidrule(lr){10-13} \cmidrule(lr){14-17} \cmidrule(lr){18-21} \cmidrule(lr){22-25}
& MAE & $F_\beta^{\max}$ & $S_m$ & $E_{m}$ 
& MAE & $F_\beta^{\max}$ & $S_m$ & $E_{m}$ 
& MAE & $F_\beta^{\max}$ & $S_m$ & $E_{m}$ 
& MAE & $F_\beta^{\max}$ & $S_m$ & $E_{m}$ 
& MAE & $F_\beta^{\max}$ & $S_m$ & $E_{m}$ 
& MAE & $F_\beta^{\max}$ & $S_m$ & $E_{m}$ \\
\midrule
\multicolumn{25}{c}{\textit{Encoder with Full-Parameter Fine-Tuning}} \\
\midrule
DGN-Full & .024 & .946 & .926 & .949 & .039 & \textcolor{mygreen}{\textbf{.893}} & .880 & .901 & .023 & .976 & .950 & .965 & .023 & .960 & .938 & .964 & .047 & \textcolor{mygreen}{\textbf{.929}} & .896 & .924 & .0312 & .9408 & .9180 & .9406 \\
\midrule
\multicolumn{25}{c}{\textit{Model Variants with Different Transformer Encoders}} \\
\midrule
DGN-Swin & .025 & .925 & .920 & .951 & .040 & .860 & .871 & .902 & .023 & .965 & .946 & .967 & \textcolor{mygreen}{\textbf{.022}} & .954 & .936 & .966 & .049 & .905 & .887 & .925 & .0318 & .9218 & .9120 & .9422 \\
DGN-PVT  & .025 & .928 & .921 & .951 & .042 & .865 & .876 & .908 & .024 & .965 & .945 & .966 & \textcolor{mygreen}{\textbf{.022}} & .954 & .935 & .965 & .054 & .900 & .879 & .917 & .0334 & .9224 & .9112 & .9414 \\
\midrule
\multicolumn{25}{c}{\textit{Model Variant with CNN Encoder}} \\
\midrule
DGN-CNN  & \textcolor{myorange}{\textbf{.021}} & .949 & \textcolor{myorange}{\textbf{.936}} & \textcolor{myorange}{\textbf{.960}} & .039 & .892 & .889 & \textcolor{mygreen}{\textbf{.916}} & \textcolor{mygreen}{\textbf{.021}} & .976 & .952 & \textcolor{mygreen}{\textbf{.970}} & \textcolor{myorange}{\textbf{.020}} & \textcolor{mygreen}{\textbf{.965}} & .942 & \textcolor{myorange}{\textbf{.970}} & \textcolor{mygreen}{\textbf{.042}} & .927 & \textcolor{mygreen}{\textbf{.900}} & \textcolor{myorange}{\textbf{.934}} & .0286 & .9418 & .9238 & \textcolor{mygreen}{\textbf{.9500}} \\
\midrule
\multicolumn{25}{c}{\textit{Model Variants with Encoders Utilizing Different Query-Token Dimensions}} \\
\midrule
DGN-L$_{10}$  & \textcolor{myorange}{\textbf{.021}} & \textcolor{mygreen}{\textbf{.951}} & \textcolor{myorange}{\textbf{.936}} & \textcolor{mygreen}{\textbf{.958}} & \textcolor{mygreen}{\textbf{.038}} & \textcolor{myorange}{\textbf{.895}} & \textcolor{mygreen}{\textbf{.891}} & \textcolor{mygreen}{\textbf{.916}} & \textcolor{myred}{\textbf{.018}} & \textcolor{myorange}{\textbf{.979}} & \textcolor{myred}{\textbf{.957}} & \textcolor{myred}{\textbf{.973}} & \textcolor{myorange}{\textbf{.020}} & \textcolor{myred}{\textbf{.967}} & \textcolor{myorange}{\textbf{.944}} & \textcolor{myorange}{\textbf{.970}} & \textcolor{mygreen}{\textbf{.042}} & \textcolor{mygreen}{\textbf{.929}} & \textcolor{myorange}{\textbf{.902}} & \textcolor{myorange}{\textbf{.934}} & \textcolor{mygreen}{\textbf{.0278}} & \textcolor{myorange}{\textbf{.9442}} & \textcolor{mygreen}{\textbf{.9260}} & \textcolor{myorange}{\textbf{.9502}} \\
DGN-L$_{100}$ & \textcolor{mygreen}{\textbf{.023}} & \textcolor{myorange}{\textbf{.953}} & .930 & .949 & .039 & .891 & .876 & .890 & .022 & \textcolor{myorange}{\textbf{.979}} & .952 & .964 & .023 & .964 & .938 & .960 & .044 & \textcolor{myorange}{\textbf{.930}} & \textcolor{mygreen}{\textbf{.900}} & \textcolor{mygreen}{\textbf{.930}} & .0302 & \textcolor{mygreen}{\textbf{.9434}} & .9192 & .9386 \\
\midrule
\multicolumn{25}{c}{\textit{Model Variants with Encoders Utilizing Different High-Level Single-Scale Features}} \\
\midrule
DGN-L-F${_{3}}$ & \textcolor{mygreen}{\textbf{.023}} & .939 & \textcolor{mygreen}{\textbf{.935}} & .953 & \textcolor{mygreen}{\textbf{.038}} & .885 & .890 & .912 & \textcolor{mygreen}{\textbf{.021}} & .971 & .953 & .968 & \textcolor{mygreen}{\textbf{.022}} & .959 & \textcolor{mygreen}{\textbf{.943}} & .966 & \textcolor{mygreen}{\textbf{.042}} & .919 & \textcolor{myred}{\textbf{.903}} & \textcolor{myred}{\textbf{.935}} & .0292 & .9346 & .9248 & .9468 \\
DGN-L-F${_{4}}$ & .027 & .930 & .927 & .941 & .041 & .877 & .885 & .901 & .028 & .964 & .945 & .957 & .028 & .951 & .934 & .954 & .046 & .912 & .899 & .926 & .0340 & .9268 & .9180 & .9358 \\
\midrule
\multicolumn{25}{c}{\textit{Model with Its Corresponding Pruning}} \\
\midrule
DGN$^{*}$-L & \textcolor{myred}{\textbf{.019}} & \textcolor{myred}{\textbf{.957}} & \textcolor{myred}{\textbf{.939}} & \textcolor{myred}{\textbf{.962}} & \textcolor{myorange}{\textbf{.035}} & \textcolor{myred}{\textbf{.901}} & \textcolor{myorange}{\textbf{.892}} & \textcolor{myorange}{\textbf{.917}} & \textcolor{myorange}{\textbf{.020}} & \textcolor{myred}{\textbf{.980}} & \textcolor{mygreen}{\textbf{.954}} & .969 & \textcolor{myorange}{\textbf{.020}} & \textcolor{myred}{\textbf{.967}} & \textcolor{myred}{\textbf{.945}} & \textcolor{mygreen}{\textbf{.969}} & \textcolor{myred}{\textbf{.040}} & \textcolor{myred}{\textbf{.932}} & \textcolor{myorange}{\textbf{.902}} & \textcolor{myorange}{\textbf{.934}} & \textcolor{myorange}{\textbf{.0268}} & \textcolor{myred}{\textbf{.9474}} & \textcolor{myorange}{\textbf{.9264}} & \textcolor{myorange}{\textbf{.9502}} \\
DGN-L      & \textcolor{myred}{\textbf{.019}} & \textcolor{mygreen}{\textbf{.951}} & \textcolor{myred}{\textbf{.939}} & \textcolor{myred}{\textbf{.962}} & \textcolor{myred}{\textbf{.034}} & \textcolor{mygreen}{\textbf{.893}} & \textcolor{myred}{\textbf{.894}} & \textcolor{myred}{\textbf{.919}} & \textcolor{myred}{\textbf{.018}} & \textcolor{mygreen}{\textbf{.978}} & \textcolor{myorange}{\textbf{.956}} & \textcolor{myorange}{\textbf{.972}} & \textcolor{myred}{\textbf{.019}} & \textcolor{myorange}{\textbf{.966}} & \textcolor{myorange}{\textbf{.944}} & \textcolor{myred}{\textbf{.971}} & \textcolor{myorange}{\textbf{.041}} & .926 & \textcolor{myorange}{\textbf{.902}} & \textcolor{myorange}{\textbf{.934}} & \textcolor{myred}{\textbf{.0262}} & .9428 & \textcolor{myred}{\textbf{.9270}} & \textcolor{myred}{\textbf{.9516}} \\
\bottomrule
\end{tabular*}
\end{table*}

\begin{figure}[!t]
  \centering
  \definecolor{mygreen}{RGB}{0,102,0}
  \definecolor{myred}{RGB}{153,0,0}
  \definecolor{myblue}{RGB}{0,122,122}
  \definecolor{myorange}{RGB}{204,106,0}
  \begin{tikzpicture}%
    \begin{scope}[xshift=0cm, yshift=0cm ]
        \node[above right] (fig63) at (0.0,0.4){\hspace{-0.1cm}\includegraphics[width=0.49\textwidth]{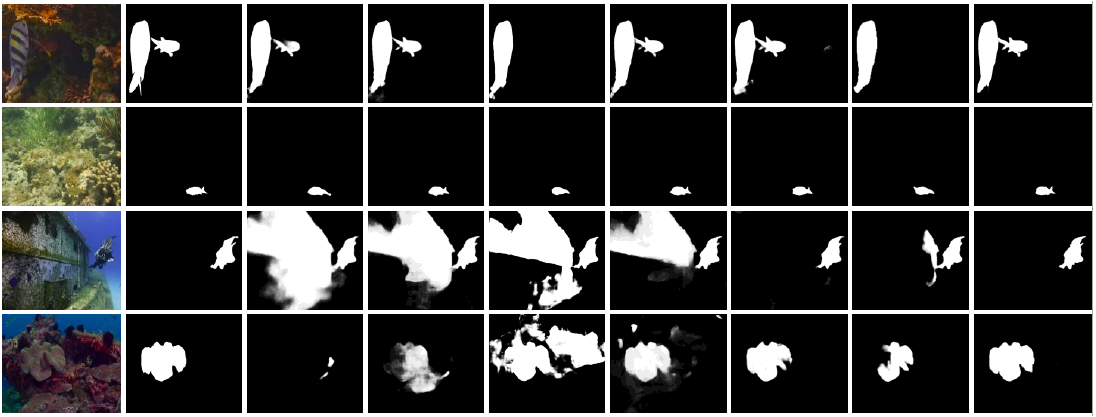}};
         \node[inner sep=1pt] at (0.55,0.3) {(a)};
        \node[inner sep=1pt] at (1.55,0.3) {(b)};
        \node[inner sep=1pt] at (2.54,0.3) {(c)};
        \node[inner sep=1pt] at (3.53,0.3) {(d)};
        \node[inner sep=1pt] at (4.52,0.3) {(e)};
        \node[inner sep=1pt] at (5.51,0.3) {(f)};
        \node[inner sep=1pt] at (6.5,0.3) {(g)};
        \node[inner sep=1pt] at (7.5,0.3) {(h)};
        \node[inner sep=1pt] at (8.5,0.3) {(i)};
        \node[inner sep=1pt] at (8.5,0.3) {(i)};
    \end{scope}
  \end{tikzpicture}
  \caption{Qualitative comparison with six state-of-the-art methods on USOD10K. From (a) to (i): input image, ground truth, TC-USOD~\cite{10102831}, CATNet~\cite{10179145}, PICRNet~\cite{cong2023point}, Dual-SAM~\cite{10657195}, iGAN~\cite {10697198}, STAMF~\cite{ma2025stamf}, and DGN-L (ours).}
  \label{fig11}
\end{figure}
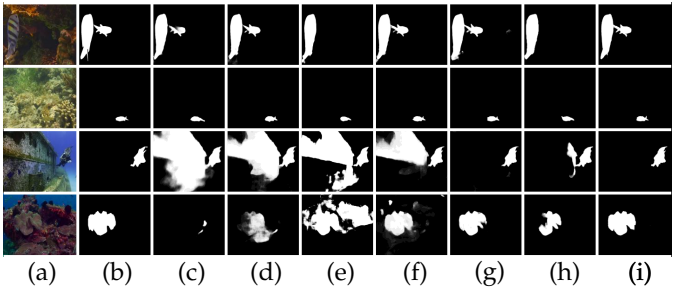

\textbf{Efficiency comparison.} Beyond the quantitative and qualitative performance analysis, Table \ref{tab:average_performance} also compares the computational efficiency of DGN with state-of-the-art methods. Within our framework, DGN-B, which employs a lightweight backbone, reduces the number of parameters and FLOPs by 62.9$\%$ and 56.9$\%$  compared to DGN-L, respectively, and achieves 41.9$\%$ faster inference speed (FPS). Meanwhile, DGN-B maintains competitive performance with acceptable degradation relative to DGN-L, as demonstrated in Table \ref{tab:average_performance}. Despite the performance trade-off, DGN-B still outperforms Transformer-based methods with comparable parameters, namely VST++ \cite{10497889} and MDSAM \cite{mdsam}, by an average of 6.37$\%$, 1.03$\%$, 0.77$\%$, and 0.47$\%$ in MAE, $F^{\max}_\beta$, $S_{m}$, and $E_{m}$, respectively, while achieving 54.4$\%$ faster inference speed on average. Similarly, DGN-L surpasses SAM2UNet \cite{sam2unet}, BirefNet \cite{birefnet}, which have comparable parameter counts, by an average of 7.09$\%$, 1.08$\%$, 0.71$\%$, and 0.82$\%$ in the same metrics, with a 68.6$\%$ improvement in FPS. These results demonstrate that our method, by implementing dual biological principles through a minimalist design philosophy, not only enhances performance but also improves computational efficiency.

\subsection{Generalization on Camouflaged and Underwater Salient Object Detection}
To further demonstrate the generalization and robustness of our method, we conduct evaluations on two additional tasks: camouflaged object detection (COD) and underwater salient object detection (USOD). 

\emph{a) Camouflaged object detection.} The primary challenge in COD lies in the high similarity between targets and backgrounds in terms of texture and color, coupled with variable contextual priors. We retrain DGN-L on the training subset of COD10K \cite{fan2020camouflaged} and evaluate it on four benchmark datasets, including CAMO \cite{le2019anabranch}, NC4K \cite{9577641}, CHAMELEON \cite{skurowski2018animal}, and the testing subset of COD10K \cite{fan2020camouflaged}. We compare our method with 14 state-of-the-art approaches, including Zoomnet \cite{zoomnet}, MRRNet \cite{10180211}, FSPNet \cite{huang2023feature}, DGNet \cite{ji2023deep}, CamoDiffusion \cite{chen2024camodiffusion}, FEDER \cite{10203727}, LSR \cite{10007893}, RISNet \cite{risnet}, PRNet \cite{10379651}, CamoFormer \cite{10623294}, SAM-TTT \cite{yu2025sam}, CGD \cite{zhang2025cgcod}, CTF-Net \cite{CTF-Net}, and BCLNet \cite{bclnet}. For fair comparisons, methods with publicly available code are re-evaluated utilizing their released implementations, while results for the remaining methods are directly cited from the original publications. The evaluation metrics and training configurations remain consistent with those utilized for SOD. As shown in Table \ref{tab:cod_comparison}, our method achieves the best performance on nearly all metrics. The qualitative results in Fig. \ref{fig10} further demonstrate that our method achieves superior object localization and detail reconstruction in challenging scenarios compared to state-of-the-art approaches. 

\emph{b) Underwater salient object detection.} The primary challenges in USOD arise from water-induced spectral attenuation, scattering, and severe color distortion. We retrain DGN-L on the training set of USOD10K \cite{10102831} and evaluate it on the corresponding test set. The evaluation metrics and training configurations remain consistent with those utilized for SOD. We compare our method with 12 state-of-the-art approaches, including TC-USOD~\cite{10102831}, HIDA-Net~\cite{10091765}, CATNet~\cite{10179145}, PICRNet~\cite{cong2023point}, Dual-SAM~\cite{10657195}, VST++\cite{10497889}, iGAN\cite{10697198}, STAMF~\cite{ma2025stamf}, STAMF-DP~\cite{ma2025stamf}, DSPL-DE~\cite{10744585}, HEHP~\cite{11018233}, and FSCDiff~\cite{li2025fscdiff}. Notably, among methods published after 2023, only VST++, STAMF, and our method rely solely on RGB inputs, whereas all others incorporate additional modalities such as depth or polarization maps, along with auxiliary denoising, compensation, or enhancement modules. Despite utilizing only RGB information, our method still achieves competitive performance, as demonstrated in Table \ref{tab:underwater_comparison} and Fig. \ref{fig11}.

Across both COD and USOD benchmarks, DGN-L consistently achieves competitive results while relying solely on RGB information without task-specific enhancements. These results demonstrate the effective generalization, robustness, and adaptability of our biologically-inspired dual-gaze mechanism across diverse detection scenarios.

\begin{table*}[!t]
\centering
  \definecolor{mygreen}{RGB}{0,102,0}
  \definecolor{myred}{RGB}{153,0,0}
  \definecolor{myblue}{RGB}{0,122,122}
  \definecolor{myorange}{RGB}{204,106,0}
\caption{: Ablation results on different input resolutions and pruning across five benchmark SOD datasets. The best results are in \textcolor{myred}{red} marked \textbf{in bold}, second best in \textcolor{myorange}{orange} marked \textbf{in bold}, and third best in \textcolor{mygreen}{green} marked \textbf{in bold}. MAE: lower is better ($\downarrow$); other metrics: higher is better ($\uparrow$).  Please refer to the text for more details.}
\label{tab:sod_comparison222222}
\renewcommand{\arraystretch}{0.8}
\setlength{\tabcolsep}{1pt}
\begin{tabular*}{\textwidth}{@{\extracolsep{\fill}}l|cccc|cccc|cccc|cccc|cccc|cccc@{}}
\toprule
\multirow{2}{*}{\textbf{Method}} &
\multicolumn{4}{c|}{\textbf{DUTS-TE (5019)}} &
\multicolumn{4}{c|}{\textbf{DUT-OMR (5168)}} &
\multicolumn{4}{c|}{\textbf{ECSSD (1000)}} & 
\multicolumn{4}{c|}{\textbf{HKU-IS (4447)}} & 
\multicolumn{4}{c|}{\textbf{PASCAL-S (850)}} &
\multicolumn{4}{c}{\textbf{Average}} \\
\cmidrule(lr){2-5} \cmidrule(lr){6-9} \cmidrule(lr){10-13} \cmidrule(lr){14-17} \cmidrule(lr){18-21} \cmidrule(lr){22-25}
& MAE & $F_\beta^{\max}$ & $S_m$ & $E_{m}$ 
& MAE & $F_\beta^{\max}$ & $S_m$ & $E_{m}$ 
& MAE & $F_\beta^{\max}$ & $S_m$ & $E_{m}$ 
& MAE & $F_\beta^{\max}$ & $S_m$ & $E_{m}$ 
& MAE & $F_\beta^{\max}$ & $S_m$ & $E_{m}$ 
& MAE & $F_\beta^{\max}$ & $S_m$ & $E_{m}$ \\
\midrule
DGN$^{*}$-B$_{224}$ & .033 & .911 & .895 & .931 & .047 & .857 & .855 & .891 & .034 & .955 & .925 & .946 & .028 & .947 & .923 & .955 & .064 & .886 & .859 & .898 & .0412 & .9112 & .8914 & .9242 \\
DGN-B$_{224}$ & .026 & .927 & .919 & .948 & .039 & .879 & .883 & .912 & .023 & .969 & .947 & .965 & .024 & .956 & .935 & .963 & .051 & .906 & .885 & .918 & .0326 & .9274 & .9138 & .9412 \\
\midrule
DGN$^{*}$-B$_{352}$ & .023 & .932 & .924 & .952 & .037 & .874 & .882 & .909 & .023 & .968 & .947 & .965 & \textcolor{mygreen}{\textbf{.021}} & .959 & .940 & .967 & .049 & .906 & .889 & .924 & .0306 & .9278 & .9164 & .9434 \\
DGN-B$_{352}$ & .022 & .938 & \textcolor{mygreen}{\textbf{.930}} & .958 & \textcolor{mygreen}{\textbf{.036}} & .883 & .889 & \textcolor{mygreen}{\textbf{.918}} & \textcolor{mygreen}{\textbf{.020}} & .971 & .951 & .969 & \textcolor{mygreen}{\textbf{.021}} & .959 & .940 & \textcolor{mygreen}{\textbf{.968}} & .045 & .912 & .894 & .929 & .0288 & .9326 & .9208 & .9484 \\
\midrule
DGN$^{*}$-B$_{512}$ & .023 & .942 & .928 & .953 & .038 & .884 & .884 & .910 & \textcolor{mygreen}{\textbf{.020}} & .975 & \textcolor{myorange}{\textbf{.954}} & \textcolor{mygreen}{\textbf{.970}} & .022 & .960 & .939 & .964 & .048 & .916 & .892 & .923 & .0302 & .9354 & .9194 & .9440 \\
DGN-B$_{512}$ & .022 & .945 & \textcolor{mygreen}{\textbf{.930}} & .956 & \textcolor{mygreen}{\textbf{.036}} & \textcolor{mygreen}{\textbf{.892}} & .890 & .917 & .022 & .972 & .948 & .965 & \textcolor{mygreen}{\textbf{.021}} & .963 & .941 & .967 & .046 & .920 & .893 & .926 & .0294 & .9384 & .9204 & .9462 \\
\midrule
DGN$^{*}$-L$_{224}$ & .023 & .939 & .929 & .954 & \textcolor{mygreen}{\textbf{.036}} & .890 & .889 & .913 & .021 & .974 & .952 & .968 & \textcolor{mygreen}{\textbf{.021}} & .962 & .942 & .967 & .044 & .919 & \textcolor{mygreen}{\textbf{.897}} & .928 & .0290 & .9368 & .9218 & .9460 \\
DGN-L$_{224}$ & .022 & .939 & \textcolor{mygreen}{\textbf{.930}} & .957 & \textcolor{mygreen}{\textbf{.036}} & \textcolor{myorange}{\textbf{.893}} & \textcolor{myred}{\textbf{.895}} & \textcolor{myred}{\textbf{.922}} & \textcolor{mygreen}{\textbf{.020}} & .972 & .952 & \textcolor{mygreen}{\textbf{.970}} & \textcolor{myorange}{\textbf{.020}} & .962 & .943 & \textcolor{myorange}{\textbf{.969}} & .045 & .916 & .\textcolor{mygreen}{\textbf{897}} & \textcolor{mygreen}{\textbf{.932}} & .0286 & .9364 & .9234 & \textcolor{mygreen}{\textbf{.9500}} \\
\midrule
DGN$^{*}$-L$_{352}$ & \textcolor{myorange}{\textbf{.020}} & .947 & \textcolor{myorange}{\textbf{.937}} & \textcolor{myorange}{\textbf{.961}} & .037 & .883 & .886 & .911 & \textcolor{mygreen}{\textbf{.020}} & .973 & \textcolor{mygreen}{\textbf{.953}} & \textcolor{mygreen}{\textbf{.970}} & \textcolor{myred}{\textbf{.019}} & \textcolor{mygreen}{\textbf{.964}} & \textcolor{myred}{\textbf{.946}} & \textcolor{myred}{\textbf{.971}} & .044 & .920 & \textcolor{mygreen}{\textbf{.897}} & .931 & \textcolor{mygreen}{\textbf{.0280}} & .9374 & .9238 & .9488 \\
DGN-L$_{352}$ & \textcolor{mygreen}{\textbf{.021}} & \textcolor{mygreen}{\textbf{.949}} & \textcolor{myorange}{\textbf{.937}} & \textcolor{mygreen}{\textbf{.960}} & .037 & .889 & .890 & .912 & \textcolor{myorange}{\textbf{.019}} & \textcolor{mygreen}{\textbf{.977}} & \textcolor{myred}{\textbf{.956}} & \textcolor{myorange}{\textbf{.971}} & \textcolor{mygreen}{\textbf{.021}} & \textcolor{mygreen}{\textbf{.964}} & .943 & \textcolor{mygreen}{\textbf{.968}} & \textcolor{mygreen}{\textbf{.042}} & \textcolor{mygreen}{\textbf{.924}} & \textcolor{myred}{\textbf{.903}} & \textcolor{myred}{\textbf{.936}} & \textcolor{mygreen}{\textbf{.0280}} & \textcolor{mygreen}{\textbf{.9406}} & \textcolor{mygreen}{\textbf{.9258}} & .9494 \\
\midrule
DGN$^{*}$-L$_{512}$ & \textcolor{myred}{\textbf{.019}} & \textcolor{myred}{\textbf{.957}} & \textcolor{myred}{\textbf{.939}} & \textcolor{myred}{\textbf{.962}} & \textcolor{myorange}{\textbf{.035}} & \textcolor{myred}{\textbf{.901}} & \textcolor{mygreen}{\textbf{.892}} & .917 & \textcolor{mygreen}{\textbf{.020}} & \textcolor{myred}{\textbf{.980}} & \textcolor{myorange}{\textbf{.954}} & .969 & \textcolor{myorange}{\textbf{.020}} & \textcolor{myred}{\textbf{.967}} & \textcolor{myorange}{\textbf{.945}} & \textcolor{myorange}{\textbf{.969}} & \textcolor{myred}{\textbf{.040}} & \textcolor{myred}{\textbf{.932}} & \textcolor{myorange}{\textbf{.902}} & \textcolor{myorange}{\textbf{.934}} & \textcolor{myorange}{\textbf{.0268}} & \textcolor{myred}{\textbf{.9474}} & \textcolor{myorange}{\textbf{.9264}} & \textcolor{myorange}{\textbf{.9502}} \\
DGN-L$_{512}$ & \textcolor{myred}{\textbf{.019}} & \textcolor{myorange}{\textbf{.951}} & \textcolor{myred}{\textbf{.939}} & \textcolor{myred}{\textbf{.962}} & \textcolor{myred}{\textbf{.034}} & \textcolor{myorange}{\textbf{.893}} & \textcolor{myorange}{\textbf{.894}} & \textcolor{myorange}{\textbf{.919}} & \textcolor{myred}{\textbf{.018}} & \textcolor{myorange}{\textbf{.978}} & \textcolor{myred}{\textbf{.956}} & \textcolor{myred}{\textbf{.972}} & \textcolor{myred}{\textbf{.019}} & \textcolor{myorange}{\textbf{.966}} & \textcolor{mygreen}{\textbf{.944}} & \textcolor{myred}{\textbf{.971}} & \textcolor{myorange}{\textbf{.041}} & \textcolor{myorange}{\textbf{.926}} & \textcolor{myorange}{\textbf{.902}} & \textcolor{myorange}{\textbf{.934}} & \textcolor{myred}{\textbf{.0262}} & \textcolor{myorange}{\textbf{.9428}} & \textcolor{myred}{\textbf{.9270}} & \textcolor{myred}{\textbf{.9516}}  \\
\bottomrule
\end{tabular*}
\end{table*}

\subsection{Ablation study}
To clearly demonstrate the effectiveness of each module, we conduct a series of ablation studies to quantify the contribution of individual components in DGN-L. We conduct ablation experiments on the SOD datasets from Subsection \ref{SOTA}, utilizing MAE, $F_{\beta}^{\max}$, $S_{m}$, and $E_{m}$ as evaluation metrics. 


\begin{figure}[!t]
  \centering
  \definecolor{mygreen}{RGB}{0,102,0}
  \definecolor{myred}{RGB}{153,0,0}
  \definecolor{myblue}{RGB}{0,122,122}
  \definecolor{myorange}{RGB}{204,106,0}
  \begin{tikzpicture}%
    \begin{scope}[xshift=0cm, yshift=0cm ]
        \node[above right] (fig63) at (0.0,0.4){\hspace{-0.1cm}\includegraphics[width=0.49\textwidth]{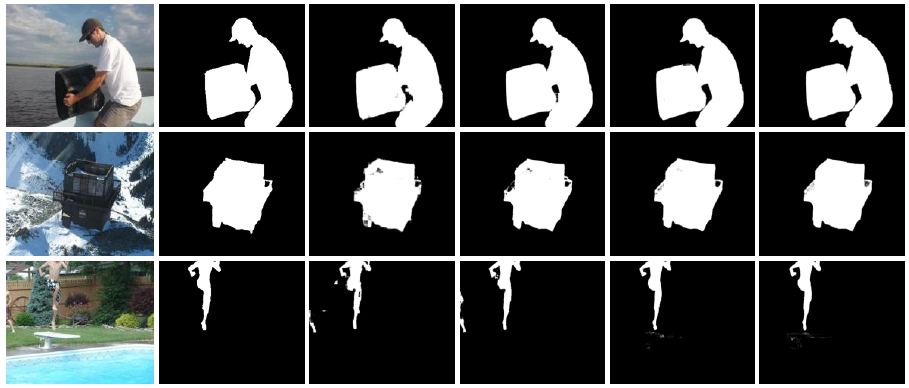}};
         \node[inner sep=1pt] at (0.85,0.3) {(a)};
        \node[inner sep=1pt] at (2.3,0.3) {(b)};
        \node[inner sep=1pt] at (3.75,0.3) {(c)};
        \node[inner sep=1pt] at (5.25,0.3) {(d)};
        \node[inner sep=1pt] at (6.7,0.3) {(e)};
        \node[inner sep=1pt] at (8.2,0.3) {(f)};
    \end{scope}
  \end{tikzpicture}
  \caption{Visualization of models with different backbones. (a) Input image; (b) ground truth; (c) Swin-B backbone (d) PVT-b5 backbone; (e) Hiera*-L backbone; (f) Hiera-L backbone (ours).}
  \label{fig14}
\end{figure}
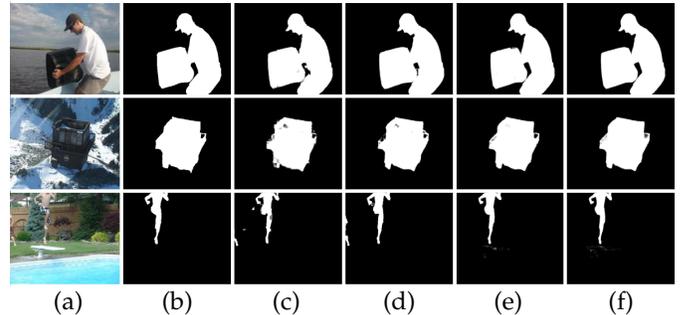

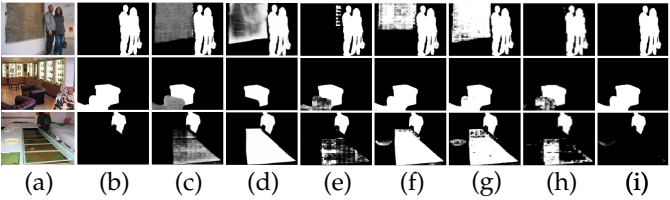
\begin{figure}[!t]
  \centering
  \definecolor{mygreen}{RGB}{0,102,0}
  \definecolor{myred}{RGB}{153,0,0}
  \definecolor{myblue}{RGB}{0,122,122}
  \definecolor{myorange}{RGB}{204,106,0}
  \begin{tikzpicture}%
    \begin{scope}[xshift=0cm, yshift=0cm ]
        \node[above right] (fig63) at (0.0,0.4){\hspace{-0.1cm}\includegraphics[width=0.49\textwidth, height=2.2cm]{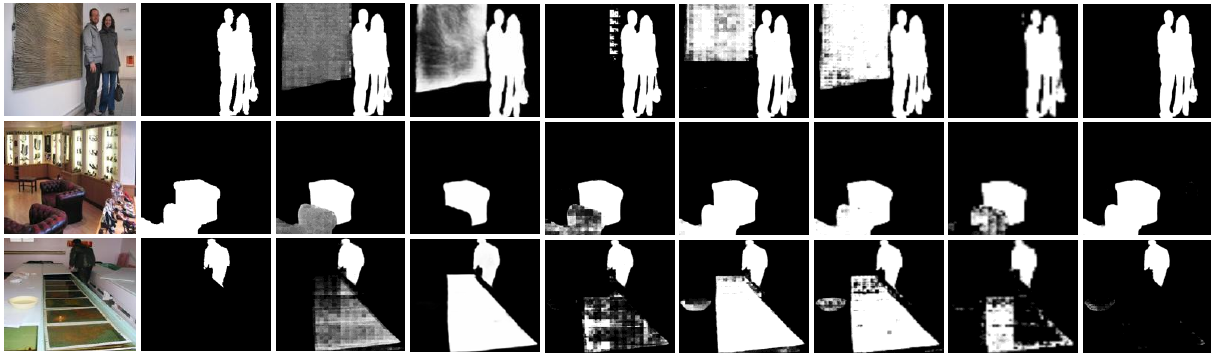}};
         \node[inner sep=1pt] at (0.55,0.3) {(a)};
        \node[inner sep=1pt] at (1.55,0.3) {(b)};
        \node[inner sep=1pt] at (2.54,0.3) {(c)};
        \node[inner sep=1pt] at (3.53,0.3) {(d)};
        \node[inner sep=1pt] at (4.52,0.3) {(e)};
        \node[inner sep=1pt] at (5.51,0.3) {(f)};
        \node[inner sep=1pt] at (6.5,0.3) {(g)};
        \node[inner sep=1pt] at (7.5,0.3) {(h)};
        \node[inner sep=1pt] at (8.5,0.3) {(i)};
        \node[inner sep=1pt] at (8.5,0.3) {(i)};
    \end{scope}
  \end{tikzpicture}
  \caption{Visualizations of models with Hiera-L backbone. (a) Input; (b) GT; (c) fully finetuned encoder; (d) CNN decoder; (e) decoder with 10-dimensional query; (f) decoder with 100-dimensional query; (g) decoder with single-scale feature $F_{3}$; (h) decoder with single-scale feature $F_{4}$; (i) DGN-L (ours).}
  \label{fig15}
\end{figure}

\textbf{Effectiveness of adapter.} To validate the effectiveness of the adapter in Fig. \ref{fig3t}, we unfreeze all parameters of Hiera-L and retrain the model, denoted as DGN-Full. As shown in Table \ref{tab:sod_comparison222}, DGN-Full exhibits notable performance degradation compared to DGN-L. This is because the adapter-based approach preserves the pretrained general visual representations learned from large-scale datasets, whereas fully finetuning the encoder on limited SOD data leads to overfitting and degrades these robust features. The visualization in Fig. \ref{fig15} (c) further confirms this overfitting behavior, where the model tends to over-activate regions and incorrectly identifies non-salient objects as salient, thereby introducing substantial noise into the predictions.

\textbf{Effectiveness of Hiera as the encoder.} In Subsection \ref{Object Detection and Tracking}, we have discussed the advantages of Hiera. To further justify its selection for emulating human robust representation learning capabilities, we replace it with two widely-utilized Transformer architectures, Swin-B and PVT-B5, as encoders for extracting hierarchical features, denoted as DGN-Swin and DGN-PVT, respectively. The remaining model architecture remains unchanged. Since Table \ref{tab:sod_comparison} and existing literature have demonstrated the superiority of Transformers with self-attention and positional encoding for SOD, we do not examine CNN-based encoders in this ablation study. 

As shown in Table \ref{tab:sod_comparison222}, DGN-Swin and DGN-PVT achieve comparable performance to each other, but both underperform DGN-L. Specifically, DGN-L reduces MAE by 17.61$\%$ and 21.56$\%$ compared to DGN-Swin and DGN-PVT, respectively. For $F_{\beta}^{\max}$, $S_{m}$, and $E_{m}$, DGN-L outperforms DGN-Swin by 2.28$\%$, 1.64$\%$, 1.00$\%$, and surpasses DGN-PVT by 2.21$\%$, 1.73$\%$, 1.08$\%$, respectively. This superiority stems from several architectural advantages of Hiera. First, unlike the window-based attention in Swin-B, Hiera facilitates better modeling of long-range dependencies across windows, thereby reducing the constraints imposed by window partitioning on cross-object connectivity. Second, compared to the spatial reduction strategy in PVT, Hiera better preserves spatial details when establishing multi-scale semantic representations. Third, the masked autoencoding pretraining employed in Hiera encourages inherent robustness to occlusions and noisy boundaries during feature extraction, enabling more accurate recovery of object boundary contours. The visual comparisons in Fig. \ref{fig14} among columns (c), (d), and (f) further confirm that the other two backbones may introduce more noise along object boundaries.

\begin{table}[t]
\centering
\caption{: Quantitative comparison of model complexity and efficiency. FPS is measured on an NVIDIA RTX 5090 GPU with uniform settings across all methods.}
\label{tab:model_comparison222222}
\setlength{\tabcolsep}{1.5pt}
\renewcommand{\arraystretch}{0.8}
\begin{tabular}{@{}lcccc|lcccc@{}}
\toprule
\textbf{Method} & 
\textbf{Size} & 
\textbf{Params} & 
\textbf{FLOPs} & 
\textbf{FPS} &
\textbf{Method} & 
\textbf{Size} & 
\textbf{Params} & 
\textbf{FLOPs} & 
\textbf{FPS} \\
\midrule
DGN$^{*}$-B & 224$^2$ & 49.23 & 18.86 & 78 & DGN$^{*}$-L & 224$^2$ & 162.32 & 44.19 & 52 \\
DGN-B & 224$^2$ & 91.92 & 17.89 & 69 & DGN-L & 224$^2$ & 247.56 & 48.59 & 46 \\
\midrule
DGN$^{*}$-B & 352$^2$ & 49.23 & 39.13 & 77 & DGN$^{*}$-L & 352$^2$ & 162.32 & 126.27 & 50 \\
DGN-B & 352$^2$ & 91.92 & 47.95 & 64 & DGN-L & 352$^2$ & 247.56 & 139.07 & 45 \\
\midrule
DGN$^{*}$-B & 512$^2$ & 49.23 & 83.47 & 72 & DGN$^{*}$-L & 512$^2$ & 162.32 & 217.11 & 48 \\
DGN-B & 512$^2$ & 91.92 & 102.78 & 61 & DGN-L & 512$^2$ & 247.56 & 238.52 & 43  \\
\bottomrule
\end{tabular}
\end{table}

\textbf{Effectiveness of proposed decoder.} To validate the effectiveness of the query mechanism in our decoder, we replace the decoder containing MGQM and MGFRM in Fig. \ref{architecture} with the CNN decoder from Sam2UNet \cite{sam2unet}, denoted as DGN-CNN. As shown in Table \ref{tab:sod_comparison222}, replacing the decoder with a CNN decoder maintains comparable performance in $F_{\beta}^{\max}$ but degrades other metrics, with MAE increasing by 9.16$\%$, $E_{m}$ decreasing by 0.17$\%$, and $S_{m}$ decreasing by 0.35$\%$. This degradation occurs because cortical queries encode explicit spatial localization information of salient objects, whereas the CNN decoder lacks such guidance and tends to incorrectly activate non-salient regions, thereby introducing spurious responses. This limitation is evident in the visual comparison between columns (d) and (i) in Fig. \ref{fig15}, which validates the localization capability of cortical queries.

\textbf{Effect of query-token dimensionality.} In \eqref{query}, the dimensionality of cortical queries is set to one. To investigate the relationship between query dimensionality and model performance, we increase the dimensionality to 10 and 100, denoted as DGN-L$_{10}$ and DGN-L$_{100}$, respectively. As shown in Table \ref{tab:sod_comparison222},  the averaged results across five datasets indicate that when dimensionality increases to 10, $F_{\beta}^{\max}$ remains comparable, while MAE increases by 6.11$\%$, and $S_{m}$ as well as $E_{m}$ decrease by 0.11$\%$ and 0.15$\%$, respectively. When dimensionality further increases to 100, the performance degrades notably, with MAE increasing by 15.27$\%$, and $S_{m}$ as well as $E_{m}$  decreasing by 0.84$\%$ and 1.37$\%$, respectively. This phenomenon indicates that low-dimensional cortical queries are sufficient for discriminating salient regions, as saliency detection is inherently a binary task that only requires distinguishing salient from non-salient objects. Conversely, excessively high-dimensional cortical queries tend to over-segment regions and introduce conflicting guidance, leading to spurious activations of non-salient areas. This degradation is also evident in the visual comparison among columns (e), (f), and (i) in Fig. \ref{fig15}.

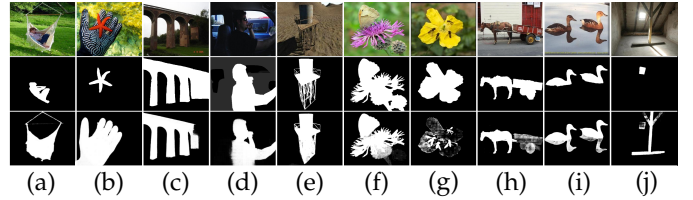
\begin{figure}[!t]
  \centering
  \definecolor{mygreen}{RGB}{0,102,0}
  \definecolor{myred}{RGB}{153,0,0}
  \definecolor{myblue}{RGB}{0,122,122}
  \definecolor{myorange}{RGB}{204,106,0}
  \begin{tikzpicture}%
    \begin{scope}[xshift=0cm, yshift=0cm ]
        \node[above right] (fig63) at (0.0,0.4){\hspace{-0.1cm}\includegraphics[width=0.49\textwidth, height=2.2cm]{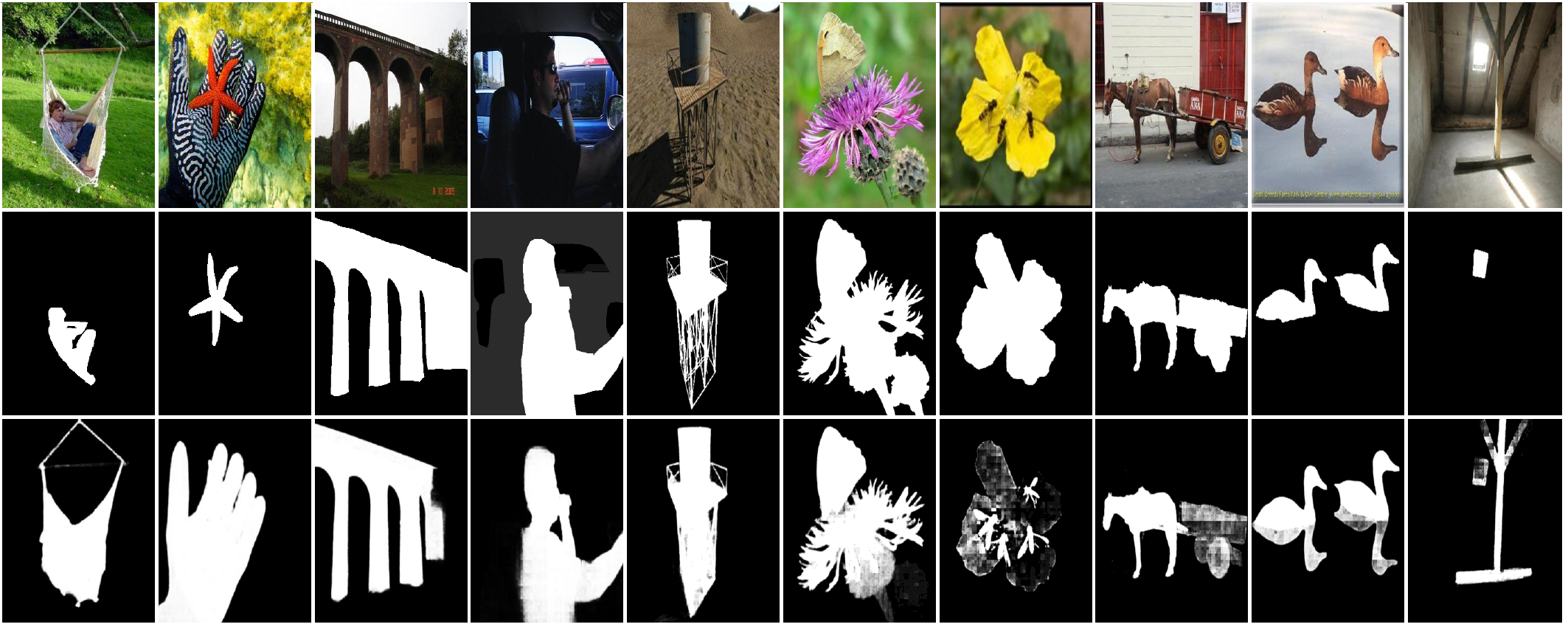}};
         \node[inner sep=1pt] at (0.45,0.3) {(a)};
        \node[inner sep=1pt] at (1.35,0.3) {(b)};
        \node[inner sep=1pt] at (2.54-0.3,0.3) {(c)};
        \node[inner sep=1pt] at (3.53-0.4,0.3) {(d)};
        \node[inner sep=1pt] at (4.52-0.5,0.3) {(e)};
        \node[inner sep=1pt] at (5.51-0.6,0.3) {(f)};
        \node[inner sep=1pt] at (6.5-0.7,0.3) {(g)};
        \node[inner sep=1pt] at (7.5-0.8,0.3) {(h)};
        \node[inner sep=1pt] at (8.5-0.9,0.3) {(i)};
        \node[inner sep=1pt] at (8.5,0.3) {(j)};
    \end{scope}
  \end{tikzpicture}
  \caption{Failure cases of DGN-L. From top to bottom: input images, ground truth, and predictions.}
  \label{fig16}
\end{figure}

\textbf{Decoder performance with single-layer high-level semantics.} Existing literature has demonstrated that relying solely on low-level features is insufficient for SOD tasks. To further investigate the impact of utilizing only high-level semantic information in our decoder, we conduct experiments where MGFRM utilizes only $F_{3}$ with its corresponding cortical query $Q_{3}$, or $F_{4}$ with $Q_{4}$, denoted as DGN-L-F${_{3}}$ and DGN-L-F${_{4}}$, respectively. As shown in Table \ref{tab:sod_comparison222}, utilizing only high-level semantic features leads to notable performance degradation. Moreover, employing higher-level features from $F_4$ results in even greater degradation compared to $F_3$. Specifically, DGN-L-F${_4}$ increases averaged MAE by 16.44$\%$ and decreases $F_{\beta}^{\max}$, $S_{m}$, and $E_{m}$ by 0.84$\%$, 0.74$\%$, and 1.16$\%$ relative to DGN-L-F${_{3}}$. These results lead to two important conclusions. First, integrating both high-level and low-level features yields better performance compared to relying solely on high-level semantics. Second, excessively high-level features, which emphasize semantics at the expense of geometric details, tend to increase the risk of structural discontinuities in salient object structures. This fragmentation effect is also evident in the visualization results shown in Fig. \ref{fig15}(h).

\textbf{Pruning Strategy.} Since DGN-L-F$_{3}$ outperforms DGN-L-F$_{4}$ when utilizing individual high-level features, we hypothesize that the $F_{4}$ level may contain redundant features.  To validate this, we prune the original model by removing the $F_{4}$ layer from both the encoder and decoder in Fig. \ref{architecture}, along with its corresponding query $Q_{4}$, and retrain the model, denoted as DGN$^{*}$-L. As shown in Table \ref{tab:sod_comparison222}, all metrics either remain comparable or even exhibit slight improvements in $F_{\beta}^{\max}$, confirming our hypothesis. To further validate this hypothesis, we conduct experiments with different backbones, Hiera-L and Hiera-B, across various input sizes of 224, 352, and 512, and retrain the pruned models accordingly. As shown in Table \ref{tab:model_comparison222222}, the pruned models consistently achieve higher FPS at the same input size.  However, Table \ref{tab:sod_comparison222222} reveals a more complex performance trend. For the heavy Hiera-L backbone, pruning remains comparable performance only at an input size of 512, whereas performance degrades at smaller resolutions. Moreover, for the lightweight Hiera-B backbone, performance degradation increases as image resolution decreases, indicating that insufficient features remain after pruning. These observations reveal that both feature redundancy and insufficiency degrade performance, highlighting that feature balance is more critical than architectural complexity. This also explains why our method has superior performance compared to existing approaches.

\subsection{Failure Cases}
Although DGN-L demonstrates superior performance and maintains reliable predictions, some failure cases remain, as illustrated in Fig. \ref{fig16}. The failures in Fig. \ref{fig16}(a)-(b) result from misidentifying the primary salient object. In Fig. \ref{fig16}(c)-(d), the model struggles with shadow effects. Fig. \ref{fig16}(e)-(f) demonstrates difficulty in capturing objects with highly intricate boundary details. The failures in Fig. \ref{fig16}(g)-(h) arise from insufficient visual information within the salient regions. Finally, Fig. \ref{fig16}(i)-(j) illustrates cases where limited training samples exist for such scenarios in the dataset. To address these limitations, future work can extend our biologically inspired framework by incorporating additional modalities, such as depth information and photometric cues, as well as leveraging more diverse training datasets to enhance performance across challenging scenarios.


\section{Conclusion}
In this paper, we presented DualGazeNet, a biologically inspired pure Transformer for SOD that first operationalizes robust representation learning and magnocellular–parvocellular dual-pathway processing with cortical attention modulation into a dual-gaze query pipeline. Despite its minimalist decoder and the absence of auxiliary tasks or heavy fusion modules, DualGazeNet achieves a favorable accuracy–efficiency trade-off and consistently outperforms representative CNN and Transformer-based methods across diverse benchmarks, and generalizes well to camouflaged and underwater scenes. These results suggest that biologically grounded design and balanced feature utilization can be more beneficial than further increasing architectural complexity, and motivate future extensions of our framework to multi-modal, video, and 3D saliency settings.





\bibliographystyle{IEEEtran}
\bibliography{mylib.bib}

@string{ECCV = "{Proceedings of the European Conference on Computer Vision}"}

@string{NIPS = "{Proceedings of the Advances in Neural Information Processing Systems}"}

@string{AAAI = "{Proceedings of the AAAI Conference on Artificial Intelligence}"}

@string{TPAMI = "{{IEEE} Transactions on Pattern Analysis and Machine Intelligence}"}

@ARTICLE{long2015fully,
  author={Shelhamer, Evan and Long, Jonathan and Darrell, Trevor},
  journal=TPAMI, 
  title={Fully {C}onvolutional {N}etworks for {S}emantic {S}egmentation}, 
  year={2017},
  volume={39},
  number={4},
  pages={640-651},
  doi={10.1109/TPAMI.2016.2572683}}

@inproceedings{vaswani2017attention,
  title={Attention is all you need},
  author={Vaswani, Ashish and Shazeer, Noam and Parmar, Niki and Uszkoreit, Jakob and Jones, Llion and Gomez, Aidan N and Kaiser, {\L}ukasz and Polosukhin, Illia},
  booktitle= NIPS,
  year={2017}
}

@inproceedings{DETR,
  title={End-to-end object detection with transformers},
  author={Carion, Nicolas and Massa, Francisco and Synnaeve, Gabriel and Usunier, Nicolas and Kirillov, Alexander and Zagoruyko, Sergey},
  booktitle= ECCV,
  pages={213-229},
  year={2020}
}

@article{mae,
  title={Masked Autoencoders Are Scalable Vision Learners},
  author={Kaiming He and Xinlei Chen and Saining Xie and Yanghao Li and Piotr Doll'ar and Ross B. Girshick},
  journal={2022 IEEE/CVF Conference on Computer Vision and Pattern Recognition},
  year={2021},
  pages={15979-15988},
}

@inproceedings{mdsam,
  title={Multi-scale and detail-enhanced segment anything model for salient object detection},
  author={Gao, Shixuan and Zhang, Pingping and Yan, Tianyu and Lu, Huchuan},
  booktitle={Proceedings of the 32nd ACM International Conference on Multimedia},
  pages={9894--9903},
  year={2024}
}

@article{birefnet,
  title={Bilateral reference for high-resolution dichotomous image segmentation},
  author={Zheng, Peng and Gao, Dehong and Fan, Deng-Ping and Liu, Li and Laaksonen, Jorma and Ouyang, Wanli and Sebe, Nicu},
  journal={arXiv preprint arXiv:2401.03407},
  year={2024}
}

@inproceedings{tracer,
  title={Tracer: Extreme attention guided salient object tracing network (student abstract)},
  author={Lee, Min Seok and Shin, WooSeok and Han, Sung Won},
  booktitle={Proceedings of the AAAI Conference on Artificial Intelligence},
  volume={36},
  number={11},
  pages={12993--12994},
  year={2022}
}

@inproceedings{hiera,
  title={Hiera: A hierarchical vision transformer without the bells-and-whistles},
  author={Ryali, Chaitanya and Hu, Yuan-Ting and Bolya, Daniel and Wei, Chen and Fan, Haoqi and Huang, Po-Yao and Aggarwal, Vaibhav and Chowdhury, Arkabandhu and Poursaeed, Omid and Hoffman, Judy and others},
  booktitle={International Conference on Machine Learning},
  pages={29441--29454},
  year={2023},
}

@inproceedings{duts,
  title={Learning to detect salient objects with image-level supervision},
  author={Wang, Lijun and Lu, Huchuan and Wang, Yifan and Feng, Mengyang and Wang, Dong and Yin, Baocai and Ruan, Xiang},
  booktitle={Proceedings of the IEEE Conference on Computer Vision and Pattern Recognition},
  pages={136--145},
  year={2017}
}

@inproceedings{duto,
  title={Saliency detection via graph-based manifold ranking},
  author={Yang, Chuan and Zhang, Lihe and Lu, Huchuan and Ruan, Xiang and Yang, Ming-Hsuan},
  booktitle={Proceedings of the IEEE Conference on Computer Vision and Pattern Recognition},
  pages={3166--3173},
  year={2013}
}

@inproceedings{hkuis,
  title={Visual saliency based on multiscale deep features},
  author={Li, Guanbin and Yu, Yizhou},
  booktitle={Proceedings of the IEEE Conference on Computer Vision and Pattern Recognition},
  pages={5455--5463},
  year={2015}
}

@inproceedings{ecssd,
  title={Hierarchical saliency detection},
  author={Yan, Qiong and Xu, Li and Shi, Jianping and Jia, Jiaya},
  booktitle={Proceedings of the IEEE Conference on Computer Vision and Pattern Recognition},
  pages={1155--1162},
  year={2013}
}

@inproceedings{pascals,
  title={The secrets of salient object segmentation},
  author={Li, Yin and Hou, Xiaodi and Koch, Christof and Rehg, James M and Yuille, Alan L},
  booktitle={Proceedings of the IEEE Conference on Computer Vision and Pattern Recognition},
  pages={280--287},
  year={2014}
}

@inproceedings{sm,
  title={Structure-measure: A new way to evaluate foreground maps},
  author={Fan, Deng-Ping and Cheng, Ming-Ming and Liu, Yun and Li, Tao and Borji, Ali},
  booktitle={Proceedings of the IEEE International Conference on Computer Vision},
  pages={4548--4557},
  year={2017}
}

@article{em,
  title={Enhanced-alignment measure for binary foreground map evaluation},
  author={Fan, Deng-Ping and Gong, Cheng and Cao, Yang and Ren, Bo and Cheng, Ming-Ming and Borji, Ali},
  journal={arXiv preprint arXiv:1805.10421},
  year={2018}
}

@article{cagnet,
  title={CAGNet: Content-aware guidance for salient object detection},
  author={Mohammadi, Sina and Noori, Mehrdad and Bahri, Ali and Majelan, Sina Ghofrani and Havaei, Mohammad},
  journal={Pattern Recognition},
  volume={103},
  pages={107303},
  year={2020},
  publisher={Elsevier}
}

@article{dfi,
  title={Dynamic feature integration for simultaneous detection of salient object, edge, and skeleton},
  author={Liu, Jiang-Jiang and Hou, Qibin and Cheng, Ming-Ming},
  journal={IEEE Transactions on Image Processing},
  volume={29},
  pages={8652--8667},
  year={2020},
  publisher={IEEE}
}

@inproceedings{gatenet,
  title={Suppress and balance: A simple gated network for salient object detection},
  author={Zhao, Xiaoqi and Pang, Youwei and Zhang, Lihe and Lu, Huchuan and Zhang, Lei},
  booktitle={Computer vision--ECCV 2020: 16th European conference, Glasgow, UK, August 23--28, 2020, proceedings, part II 16},
  pages={35--51},
  year={2020},
}

@inproceedings{minet,
  title={Multi-scale interactive network for salient object detection},
  author={Pang, Youwei and Zhao, Xiaoqi and Zhang, Lihe and Lu, Huchuan},
  booktitle={Proceedings of the IEEE/CVF Conference on Computer Vision and Pattern Recognition},
  pages={9413--9422},
  year={2020}
}

@inproceedings{ldf,
  title={Label decoupling framework for salient object detection},
  author={Wei, Jun and Wang, Shuhui and Wu, Zhe and Su, Chi and Huang, Qingming and Tian, Qi},
  booktitle={Proceedings of the IEEE/CVF Conference on Computer Vision and Pattern Recognition},
  pages={13025--13034},
  year={2020}
}

@inproceedings{menet,
  title={Pixels, regions, and objects: Multiple enhancement for salient object detection},
  author={Wang, Yi and Wang, Ruili and Fan, Xin and Wang, Tianzhu and He, Xiangjian},
  booktitle={Proceedings of the IEEE/CVF Conference on Computer Vision and Pattern Recognition},
  pages={10031--10040},
  year={2023}
}

@article{icon,
  title={Salient object detection via integrity learning},
  author={Zhuge, Mingchen and Fan, Deng-Ping and Liu, Nian and Zhang, Dingwen and Xu, Dong and Shao, Ling},
  journal={IEEE Transactions on Pattern Analysis and Machine Intelligence},
  volume={45},
  number={3},
  pages={3738--3752},
  year={2022},
  publisher={IEEE}
}

@article{bbrf,
  title={Boosting broader receptive fields for salient object detection},
  author={Ma, Mingcan and Xia, Changqun and Xie, Chenxi and Chen, Xiaowu and Li, Jia},
  journal={IEEE Transactions on Image Processing},
  volume={32},
  pages={1026--1038},
  year={2023},
  publisher={IEEE}
}

@article{dcnet,
  title={DC-Net: Divide-and-Conquer for Salient Object Detection},
  author={Jiayi Zhu and Xuebin Qin and Abdulmotaleb El Saddik},
  journal={Pattern Recognit.},
  year={2023},
  volume={157},
  pages={110903},
}

@article{sam2unet,
  title={Sam2-unet: Segment anything 2 makes strong encoder for natural and medical image segmentation},
  author={Xiong, Xinyu and Wu, Zihuang and Tan, Shuangyi and Li, Wenxue and Tang, Feilong and Chen, Ying and Li, Siying and Ma, Jie and Li, Guanbin},
  journal={arXiv preprint arXiv:2408.08870},
  year={2024}
}

@inproceedings{seqrank,
  title={Seqrank: Sequential ranking of salient objects},
  author={Guan, Huankang and Lau, Rynson WH},
  booktitle={Proceedings of the AAAI Conference on Artificial Intelligence},
  volume={38},
  number={3},
  pages={1941--1949},
  year={2024}
}

@article{pam,
  title={Pyramidal attention with progressive multi-stage iterative feature refinement for salient object segmentation},
  author={Khan, Rahim and Alzaben, Nada and Daradkeh, Yousef Ibrahim and Zhu, Xianxun and Ullah, Inam},
  journal={Image and Vision Computing},
  pages={105670},
  year={2025},
  publisher={Elsevier}
}

@article{SegR1,
  title={Seg-R1: Segmentation Can Be Surprisingly Simple with Reinforcement Learning},
  author={Zuyao You and Zuxuan Wu},
  journal={ArXiv},
  year={2025},
  volume={abs/2506.22624},
}

@article{atkinson1992early,
  title={Early visual development: Differential functioning of parvocellular and magnocellular pathways},
  author={Atkinson, Janette},
  journal={Eye},
  volume={6},
  number={2},
  pages={129--135},
  year={1992},
  publisher={Nature Publishing Group}
}

@article{han2022diversity,
  title={Diversity of spatiotemporal coding reveals specialized visual processing streams in the mouse cortex},
  author={Han, Xu and Vermaercke, Ben and Bonin, Vincent},
  journal={Nature Communications},
  volume={13},
  number={1},
  pages={3249},
  year={2022},
  publisher={Nature Publishing Group UK London}
}

@ARTICLE{11072100,
  author={Su, Zhuo and Liu, Li and Müller, Matthias and Zhang, Jiehua and Wofk, Diana and Cheng, Ming-Ming and Pietikäinen, Matti},
  journal={IEEE Transactions on Pattern Analysis and Machine Intelligence}, 
  title={Rapid Salient Object Detection With Difference Convolutional Neural Networks}, 
  year={2025},
  volume={47},
  number={10},
  pages={9061-9077},
  doi={10.1109/TPAMI.2025.3583968}}

@ARTICLE{9933726,
  author={Gao, Shanghua and Li, Zhong-Yu and Yang, Ming-Hsuan and Cheng, Ming-Ming and Han, Junwei and Torr, Philip},
  journal={IEEE Transactions on Pattern Analysis and Machine Intelligence}, 
  title={Large-Scale Unsupervised Semantic Segmentation}, 
  year={2023},
  volume={45},
  number={6},
  pages={7457-7476},
  doi={10.1109/TPAMI.2022.3218275}}

@INPROCEEDINGS{10203634,
  author={Li, Hao and Zhang, Dingwen and Liu, Nian and Cheng, Lechao and Dai, Yalun and Zhang, Chao and Wang, Xinggang and Han, Junwei},
  booktitle={2023 IEEE/CVF Conference on Computer Vision and Pattern Recognition}, 
  title={Boosting Low-Data Instance Segmentation by Unsupervised Pre-training with Saliency Prompt}, 
  year={2023},
  volume={},
  number={},
  pages={15485-15494},
  doi={10.1109/CVPR52729.2023.01486}}

@ARTICLE{10549838,
  author={Xiao, Chao and An, Wei and Zhang, Yifan and Su, Zhuo and Li, Miao and Sheng, Weidong and Pietikäinen, Matti and Liu, Li},
  journal={IEEE Transactions on Pattern Analysis and Machine Intelligence}, 
  title={Highly Efficient and Unsupervised Framework for Moving Object Detection in Satellite Videos}, 
  year={2024},
  volume={46},
  number={12},
  pages={11532-11539},
  doi={10.1109/TPAMI.2024.3409824}}

@ARTICLE{6587754,
  author={Ren, Zhixiang and Gao, Shenghua and Chia, Liang-Tien and Tsang, Ivor Wai-Hung},
  journal={IEEE Transactions on Circuits and Systems for Video Technology}, 
  title={Region-Based Saliency Detection and Its Application in Object Recognition}, 
  year={2014},
  volume={24},
  number={5},
  pages={769-779},
  doi={10.1109/TCSVT.2013.2280096}}

@article{zhang2020non,
  title={Non-rigid object tracking via deep multi-scale spatial-temporal discriminative saliency maps},
  author={Zhang, Pingping and Liu, Wei and Wang, Dong and Lei, Yinjie and Wang, Hongyu and Lu, Huchuan},
  journal={Pattern Recognition},
  volume={100},
  pages={107130},
  year={2020},
  publisher={Elsevier}
}

@inproceedings{hong2015online,
  title={Online tracking by learning discriminative saliency map with convolutional neural network},
  author={Hong, Seunghoon and You, Tackgeun and Kwak, Suha and Han, Bohyung},
  booktitle={International Conference on Machine Learning},
  pages={597--606},
  year={2015},
}

@ARTICLE{6871397,
  author={Cheng, Ming-Ming and Mitra, Niloy J. and Huang, Xiaolei and Torr, Philip H. S. and Hu, Shi-Min},
  journal={IEEE Transactions on Pattern Analysis and Machine Intelligence}, 
  title={Global Contrast Based Salient Region Detection}, 
  year={2015},
  volume={37},
  number={3},
  pages={569-582},
  doi={10.1109/TPAMI.2014.2345401}}

@ARTICLE{730558,
  author={Itti, L. and Koch, C. and Niebur, E.},
  journal={IEEE Transactions on Pattern Analysis and Machine Intelligence}, 
  title={A model of saliency-based visual attention for rapid scene analysis}, 
  year={1998},
  volume={20},
  number={11},
  pages={1254-1259},
  doi={10.1109/34.730558}}

@INPROCEEDINGS{6247743,
  author={Perazzi, Federico and Krähenbühl, Philipp and Pritch, Yael and Hornung, Alexander},
  booktitle={2012 IEEE Conference on Computer Vision and Pattern Recognition}, 
  title={Saliency filters: Contrast based filtering for salient region detection}, 
  year={2012},
  volume={},
  number={},
  pages={733-740},
  doi={10.1109/CVPR.2012.6247743}}

@INPROCEEDINGS{7410522,
  author={Zhang, Jianming and Sclaroff, Stan and Lin, Zhe and Shen, Xiaohui and Price, Brian and Mech, Radomír},
  booktitle={2015 IEEE International Conference on Computer Vision}, 
  title={Minimum Barrier Salient Object Detection at 80 FPS}, 
  year={2015},
  volume={},
  number={},
  pages={1404-1412},
  doi={10.1109/ICCV.2015.165}}

@ARTICLE{6243147,
  author={Li, Jian and Levine, Martin D. and An, Xiangjing and Xu, Xin and He, Hangen},
  journal={IEEE Transactions on Pattern Analysis and Machine Intelligence}, 
  title={Visual Saliency Based on Scale-Space Analysis in the Frequency Domain}, 
  year={2013},
  volume={35},
  number={4},
  pages={996-1010},
  doi={10.1109/TPAMI.2012.147}}

@inproceedings{liu2021tritransnet,
  title={TriTransNet: RGB-D salient object detection with a triplet transformer embedding network},
  author={Liu, Zhengyi and Wang, Yuan and Tu, Zhengzheng and Xiao, Yun and Tang, Bin},
  booktitle={Proceedings of the 29th ACM International Conference on Multimedia},
  pages={4481--4490},
  year={2021}
}

@ARTICLE{10227346,
  author={Qiu, Yu and Liu, Yun and Zhang, Le and Lu, Haotian and Xu, Jing},
  journal={IEEE Transactions on Circuits and Systems for Video Technology}, 
  title={Boosting Salient Object Detection With Transformer-Based Asymmetric Bilateral U-Net}, 
  year={2024},
  volume={34},
  number={4},
  pages={2332-2345},
  doi={10.1109/TCSVT.2023.3307693}}

@ARTICLE{10497889,
  author={Liu, Nian and Luo, Ziyang and Zhang, Ni and Han, Junwei},
  journal={IEEE Transactions on Pattern Analysis and Machine Intelligence}, 
  title={VST++: Efficient and Stronger Visual Saliency Transformer}, 
  year={2024},
  volume={46},
  number={11},
  pages={7300-7316},
  doi={10.1109/TPAMI.2024.3388153}}

@ARTICLE{10268450,
  author={Yuan, Junbin and Zhu, Aiqing and Xu, Qingzhen and Wattanachote, Kanoksak and Gong, Yongyi},
  journal={IEEE Transactions on Circuits and Systems for Video Technology}, 
  title={CTIF-Net: A CNN-Transformer Iterative Fusion Network for Salient Object Detection}, 
  year={2024},
  volume={34},
  number={5},
  pages={3795-3805},
  doi={10.1109/TCSVT.2023.3321190}}

@ARTICLE{10184101,
  author={Zeng, Zhihong and Liu, Haijun and Chen, Fenglei and Tan, Xiaoheng},
  journal={IEEE Transactions on Circuits and Systems for Video Technology}, 
  title={AirSOD: A Lightweight Network for RGB-D Salient Object Detection}, 
  year={2024},
  volume={34},
  number={3},
  pages={1656-1669},
  doi={10.1109/TCSVT.2023.3295588}}

@INPROCEEDINGS{8954094,
  author={Feng, Mengyang and Lu, Huchuan and Ding, Errui},
  booktitle={2019 IEEE/CVF Conference on Computer Vision and Pattern Recognition}, 
  title={Attentive Feedback Network for Boundary-Aware Salient Object Detection}, 
  year={2019},
  volume={},
  number={},
  pages={1623-1632},
  doi={10.1109/CVPR.2019.00172}}

@ARTICLE{10155248,
  author={Zhang, Liqian and Zhang, Qing},
  journal={IEEE Transactions on Circuits and Systems for Video Technology}, 
  title={Salient Object Detection With Edge-Guided Learning and Specific Aggregation}, 
  year={2024},
  volume={34},
  number={1},
  pages={534-548},
  doi={10.1109/TCSVT.2023.3287167}}

@article{yuan2024fgnet,
  title={FGNet: Fixation guidance network for salient object detection},
  author={Yuan, Junbin and Xiao, Lifang and Wattanachote, Kanoksak and Xu, Qingzhen and Luo, Xiaonan and Gong, Yongyi},
  journal={Neural Computing and Applications},
  volume={36},
  number={2},
  pages={569--584},
  year={2024},
  publisher={Springer}
}

@article{zhu2024separate,
  title={Separate first, then segment: An integrity segmentation network for salient object detection},
  author={Zhu, Ge and Li, Jinbao and Guo, Yahong},
  journal={Pattern Recognition},
  volume={150},
  pages={110328},
  year={2024},
  publisher={Elsevier}
}

@ARTICLE{10475351,
  author={Qin, Haolin and Xu, Tingfa and Liu, Peifu and Xu, Jingxuan and Li, Jianan},
  journal={IEEE Transactions on Geoscience and Remote Sensing}, 
  title={DMSSN: Distilled Mixed Spectral–Spatial Network for Hyperspectral Salient Object Detection}, 
  year={2024},
  volume={62},
  number={},
  pages={1-18},
  doi={10.1109/TGRS.2024.3379380}}

@ARTICLE{10778650,
  author={Tang, Hao and Li, Zechao and Zhang, Dong and He, Shengfeng and Tang, Jinhui},
  journal={IEEE Transactions on Pattern Analysis and Machine Intelligence}, 
  title={Divide-and-Conquer: Confluent Triple-Flow Network for RGB-T Salient Object Detection}, 
  year={2025},
  volume={47},
  number={3},
  pages={1958-1974},
  doi={10.1109/TPAMI.2024.3511621}}

@ARTICLE{9320524,
  author={Wang, Wenguan and Lai, Qiuxia and Fu, Huazhu and Shen, Jianbing and Ling, Haibin and Yang, Ruigang},
  journal={IEEE Transactions on Pattern Analysis and Machine Intelligence}, 
  title={Salient Object Detection in the Deep Learning Era: An In-Depth Survey}, 
  year={2022},
  volume={44},
  number={6},
  pages={3239-3259},
  doi={10.1109/TPAMI.2021.3051099}}

@ARTICLE{9756227,
  author={Wu, Yu-Huan and Liu, Yun and Zhang, Le and Cheng, Ming-Ming and Ren, Bo},
  journal={IEEE Transactions on Image Processing}, 
  title={EDN: Salient Object Detection via Extremely-Downsampled Network}, 
  year={2022},
  volume={31},
  number={},
  pages={3125-3136},
  doi={10.1109/TIP.2022.3164550}}

@article{fang2022lc3net,
  title={LC3Net: Ladder context correlation complementary network for salient object detection},
  author={Fang, Xian and Zhu, Jinchao and Shao, Xiuli and Wang, Hongpeng},
  journal={Knowledge-Based Systems},
  volume={242},
  pages={108372},
  year={2022},
  publisher={Elsevier}
}

@ARTICLE{9989433,
  author={Wu, Xin and Hong, Danfeng and Chanussot, Jocelyn},
  journal={IEEE Transactions on Image Processing}, 
  title={UIU-Net: U-Net in U-Net for Infrared Small Object Detection}, 
  year={2023},
  volume={32},
  number={},
  pages={364-376},
  doi={10.1109/TIP.2022.3228497}}

@inproceedings{liu2019simple,
  title={A simple pooling-based design for real-time salient object detection},
  author={Liu, Jiang-Jiang and Hou, Qibin and Cheng, Ming-Ming and Feng, Jiashi and Jiang, Jianmin},
  booktitle={Proceedings of the IEEE/CVF Conference on Computer Vision and Pattern Cecognition},
  pages={3917--3926},
  year={2019}
}

@inproceedings{wei2020f3net,
  title={F$^3$Net: fusion, feedback and focus for salient object detection},
  author={Wei, Jun and Wang, Shuhui and Huang, Qingming},
  booktitle={Proceedings of the AAAI Conference on Artificial Intelligence},
  volume={34},
  number={07},
  pages={12321--12328},
  year={2020}
}

@INPROCEEDINGS{9008371,
  author={Zhao, Jiaxing and Liu, Jiang-Jiang and Fan, Deng-Ping and Cao, Yang and Yang, Jufeng and Cheng, Ming-Ming},
  booktitle={2019 IEEE/CVF International Conference on Computer Vision}, 
  title={EGNet: Edge Guidance Network for Salient Object Detection}, 
  year={2019},
  volume={},
  number={},
  pages={8778-8787},
  doi={10.1109/ICCV.2019.00887}}

@article{zhou2025rmfdnet,
  title={RMFDNet: Redundant and Missing Feature Decoupling Network for salient object detection},
  author={Zhou, Qianwei and Wang, Jintao and Li, Jiaqi and Zhou, Chen and Hu, Haigen and Hu, Keli},
  journal={Engineering Applications of Artificial Intelligence},
  volume={139},
  pages={109459},
  year={2025},
  publisher={Elsevier}
}

@article{wang2021feature,
  title={Feature enhancement: predict more detailed and crisper edges},
  author={Wang, Yin and Wang, Lide and Qiu, Ji and Yang, Yueyi},
  journal={Signal, Image and Video Processing},
  volume={15},
  number={7},
  pages={1635--1642},
  year={2021},
  publisher={Springer}
}

@article{liu2022generative,
  title={Generative image inpainting using edge prediction and appearance flow},
  author={Liu, Qian and Ji, Hua and Liu, Gang},
  journal={Multimedia Tools and Applications},
  volume={81},
  number={22},
  pages={31709--31725},
  year={2022},
  publisher={Springer}
}

@article{zhao2024combining,
  title={Combining residual structure and edge loss for face image restoration with generative adversarial networks},
  author={Zhao, Jia and Liu, Bosheng and Wu, Runxiu and Han, Longzhe and Chen, Ming},
  journal={Signal, Image and Video Processing},
  volume={18},
  number={3},
  pages={2571--2582},
  year={2024},
  publisher={Springer}
}

@ARTICLE{9964258,
  author={Xu, Yihong and Ban, Yutong and Delorme, Guillaume and Gan, Chuang and Rus, Daniela and Alameda-Pineda, Xavier},
  journal={IEEE Transactions on Pattern Analysis and Machine Intelligence}, 
  title={TransCenter: Transformers With Dense Representations for Multiple-Object Tracking}, 
  year={2023},
  volume={45},
  number={6},
  pages={7820-7835},
  doi={10.1109/TPAMI.2022.3225078}}

@inproceedings{mao2022poseur,
  title={Poseur: Direct human pose regression with transformers},
  author={Mao, Weian and Ge, Yongtao and Shen, Chunhua and Tian, Zhi and Wang, Xinlong and Wang, Zhibin and den Hengel, Anton van},
  booktitle={European Conference on Computer Vision},
  pages={72--88},
  year={2022},
}

@INPROCEEDINGS{9577946,
  author={Dai, Zhigang and Cai, Bolun and Lin, Yugeng and Chen, Junying},
  booktitle={2021 IEEE/CVF Conference on Computer Vision and Pattern Recognition}, 
  title={UP-DETR: Unsupervised Pre-training for Object Detection with Transformers}, 
  year={2021},
  volume={},
  number={},
  pages={1601-1610},
  doi={10.1109/CVPR46437.2021.00165}}

@INPROCEEDINGS{9711179,
  author={Wang, Wenhai and Xie, Enze and Li, Xiang and Fan, Deng-Ping and Song, Kaitao and Liang, Ding and Lu, Tong and Luo, Ping and Shao, Ling},
  booktitle={2021 IEEE/CVF International Conference on Computer Vision}, 
  title={Pyramid Vision Transformer: A Versatile Backbone for Dense Prediction without Convolutions}, 
  year={2021},
  volume={},
  number={},
  pages={548-558},
  doi={10.1109/ICCV48922.2021.00061}}

@inproceedings{Dosovitskiy2021AnII,
  title={An Image is Worth 16x16 Words: Transformers for Image Recognition at Scale},
  author={Alexey Dosovitskiy and Lucas Beyer and Alexander Kolesnikov and Dirk Weissenborn and Xiaohua Zhai and Thomas Unterthiner and Mostafa Dehghani and Matthias Minderer and Georg Heigold and Sylvain Gelly and Jakob Uszkoreit and Neil Houlsby},
  booktitle={International Conference on Learning Representations},
  year={2021}
}

@inproceedings{zheng2021rethinking,
  title={Rethinking semantic segmentation from a sequence-to-sequence perspective with transformers},
  author={Zheng, Sixiao and Lu, Jiachen and Zhao, Hengshuang and Zhu, Xiatian and Luo, Zekun and Wang, Yabiao and Fu, Yanwei and Feng, Jianfeng and Xiang, Tao and Torr, Philip HS and others},
  booktitle={Proceedings of the IEEE/CVF Conference on Computer Vision and Pattern Recognition},
  pages={6881--6890},
  year={2021}
}

@INPROCEEDINGS{10377177,
  author={Chen, Zheng and Zhang, Yulun and Gu, Jinjin and Kong, Linghe and Yang, Xiaokang and Yu, Fisher},
  booktitle={2023 IEEE/CVF International Conference on Computer Vision}, 
  title={Dual Aggregation Transformer for Image Super-Resolution}, 
  year={2023},
  volume={},
  number={},
  pages={12278-12287},
  doi={10.1109/ICCV51070.2023.01131}}

@INPROCEEDINGS{9710747,
  author={Yuan, Li and Chen, Yunpeng and Wang, Tao and Yu, Weihao and Shi, Yujun and Jiang, Zihang and Tay, Francis E. H. and Feng, Jiashi and Yan, Shuicheng},
  booktitle={2021 IEEE/CVF International Conference on Computer Vision}, 
  title={Tokens-to-Token ViT: Training Vision Transformers from Scratch on ImageNet}, 
  year={2021},
  volume={},
  number={},
  pages={538-547},
  doi={10.1109/ICCV48922.2021.00060}}

@INPROCEEDINGS{9710031,
  author={Wu, Haiping and Xiao, Bin and Codella, Noel and Liu, Mengchen and Dai, Xiyang and Yuan, Lu and Zhang, Lei},
  booktitle={2021 IEEE/CVF International Conference on Computer Vision}, 
  title={CvT: Introducing Convolutions to Vision Transformers}, 
  year={2021},
  volume={},
  number={},
  pages={22-31},
  doi={10.1109/ICCV48922.2021.00009}}

@article{WU2024105039,
title = {SwinSOD: Salient object detection using swin-transformer},
journal = {Image and Vision Computing},
volume = {146},
pages = {105039},
year = {2024},
issn = {0262-8856},
author = {Shuang Wu and Guangjian Zhang and Xuefeng Liu},
}

@InProceedings{Liu_2021_ICCV,
    author    = {Liu, Nian and Zhang, Ni and Wan, Kaiyuan and Shao, Ling and Han, Junwei},
    title     = {Visual Saliency Transformer},
    booktitle = {Proceedings of the IEEE/CVF International Conference on Computer Vision},
    year      = {2021},
    pages     = {4722-4732}
}

@inproceedings{NEURIPS2021_82898892,
 author = {Zhang, Jing and Xie, Jianwen and Barnes, Nick and Li, Ping},
 booktitle = {Advances in Neural Information Processing Systems},
 pages = {15448--15463},
 title = {Learning Generative Vision Transformer with Energy-Based Latent Space for Saliency Prediction},
 volume = {34},
 year = {2021}
}

@ARTICLE{10834569,
  author={Sun, Ke and Chen, Zhongxi and Lin, Xianming and Sun, Xiaoshuai and Liu, Hong and Ji, Rongrong},
  journal={IEEE Transactions on Pattern Analysis and Machine Intelligence}, 
  title={Conditional Diffusion Models for Camouflaged and Salient Object Detection}, 
  year={2025},
  volume={47},
  number={4},
  pages={2833-2848},
  doi={10.1109/TPAMI.2025.3527469}}

@inproceedings{deng2018r3net,
  title={R3net: Recurrent residual refinement network for saliency detection},
  author={Deng, Zijun and Hu, Xiaowei and Zhu, Lei and Xu, Xuemiao and Qin, Jing and Han, Guoqiang and Heng, Pheng-Ann},
  booktitle={Proceedings of the 27th International Joint Conference on Artificial Intelligence},
  volume={684690},
  year={2018},
}

@ARTICLE{10516304,
  author={Bao, Liuxin and Zhou, Xiaofei and Lu, Xiankai and Sun, Yaoqi and Yin, Haibing and Hu, Zhenghui and Zhang, Jiyong and Yan, Chenggang},
  journal={IEEE Transactions on Image Processing}, 
  title={Quality-Aware Selective Fusion Network for V-D-T Salient Object Detection}, 
  year={2024},
  volume={33},
  number={},
  pages={3212-3226},
  doi={10.1109/TIP.2024.3393365}}

@ARTICLE{9669083,
  author={Liu, Jiang-Jiang and Hou, Qibin and Liu, Zhi-Ang and Cheng, Ming-Ming},
  journal={IEEE Transactions on Pattern Analysis and Machine Intelligence}, 
  title={PoolNet+: Exploring the Potential of Pooling for Salient Object Detection}, 
  year={2023},
  volume={45},
  number={1},
  pages={887-904},
  doi={10.1109/TPAMI.2021.3140168}}

@ARTICLE{10330640,
  author={He, Yang and Xiao, Lingao},
  journal={IEEE Transactions on Pattern Analysis and Machine Intelligence}, 
  title={Structured Pruning for Deep Convolutional Neural Networks: A Survey}, 
  year={2024},
  volume={46},
  number={5},
  pages={2900-2919},
  doi={10.1109/TPAMI.2023.3334614}}

@article{ma2023llm,
  title={Llm-pruner: On the structural pruning of large language models},
  author={Ma, Xinyin and Fang, Gongfan and Wang, Xinchao},
  journal={Advances in Neural Information Processing Systems},
  volume={36},
  pages={21702--21720},
  year={2023}
}

@ARTICLE{10428086,
  author={Su, Zhuo and Zhang, Jiehua and Liu, Tianpeng and Liu, Zhen and Zhang, Shuanghui and Pietikäinen, Matti and Liu, Li},
  journal={IEEE Transactions on Neural Networks and Learning Systems}, 
  title={Boosting Convolutional Neural Networks With Middle Spectrum Grouped Convolution}, 
  year={2025},
  volume={36},
  number={2},
  pages={3436-3449},
  doi={10.1109/TNNLS.2024.3355489}}

@article{gou2021knowledge,
  title={Knowledge distillation: A survey},
  author={Gou, Jianping and Yu, Baosheng and Maybank, Stephen J and Tao, Dacheng},
  journal={International Journal of Computer Vision},
  volume={129},
  number={6},
  pages={1789--1819},
  year={2021},
}

@inproceedings{xiao2023smoothquant,
  title={Smoothquant: Accurate and efficient post-training quantization for large language models},
  author={Xiao, Guangxuan and Lin, Ji and Seznec, Mickael and Wu, Hao and Demouth, Julien and Han, Song},
  booktitle={International Conference on Machine Learning},
  pages={38087--38099},
  year={2023},
}

@article{li2023q,
  title={Q-dm: An efficient low-bit quantized diffusion model},
  author={Li, Yanjing and Xu, Sheng and Cao, Xianbin and Sun, Xiao and Zhang, Baochang},
  journal={Advances in Neural Information Processing Systems},
  volume={36},
  pages={76680--76691},
  year={2023}
}

@INPROCEEDINGS{10657220,
  author={Zhao, Yian and Lv, Wenyu and Xu, Shangliang and Wei, Jinman and Wang, Guanzhong and Dang, Qingqing and Liu, Yi and Chen, Jie},
  booktitle={2024 IEEE/CVF Conference on Computer Vision and Pattern Recognition}, 
  title={DETRs Beat YOLOs on Real-time Object Detection}, 
  year={2024},
  volume={},
  number={},
  pages={16965-16974},
  doi={10.1109/CVPR52733.2024.01605}}

@INPROCEEDINGS{10203147,
  author={Jain, Jitesh and Li, Jiachen and Chiu, MangTik and Hassani, Ali and Orlov, Nikita and Shi, Humphrey},
  booktitle={2023 IEEE/CVF Conference on Computer Vision and Pattern Recognition}, 
  title={OneFormer: One Transformer to Rule Universal Image Segmentation}, 
  year={2023},
  volume={},
  number={},
  pages={2989-2998},
  doi={10.1109/CVPR52729.2023.00292}}

@inproceedings{Jin2025KeypointDETR,
  author    = {Hairong Jin and Yuefan Shen and Jianwen Lou and Kun Zhou and Youyi Zheng},
  title     = {KeypointDETR: An End-to-End 3D Keypoint Detector},
  booktitle = {European Computer Vision Association},
  volume    = {15132},
  pages     = {374--390},
  year      = {2025},
}

@article{Nam2025_SA_DETR,
  author  = {Kwangwoon Nam and Jeeheon Kim and Heeyeon Kim and Minyoung Chung},
  title   = {SA-{DETR}: Saliency Attention-based {DETR} for Salient Object Detection},
  journal = {Pattern Analysis and Applications},
  year    = {2025, DOI: 10.1007/s10044-024-01379-5},
  volume  = {28},
}

@article{Cen2025_PDDNet,
  author  = {Chaojun Cen and Fei Li and Zhenbo Li and Yun Wang},
  title   = {Towards Salient Object Detection via Parallel Dual-Decoder Network},
  journal = {Engineering Applications of Artificial Intelligence},
  year    = {2025},
  volume  = {139},
  pages   = {109638},
}

@article{Yang2024_SEF_E2E,
  author  = {Chen Yang and Yang Xiao and Lili Chu and Ziping Yu and Jun Zhou and Huilong Zheng},
  title   = {Saliency and Edge Features-Guided End-to-End Network for Salient Object Detection},
  journal = {Expert Systems with Applications},
  year    = {2024},
  volume  = {257},
  pages   = {125016},
}

@article{vonderHeydt1984Science,
  author  = {von der Heydt, R{\"u}diger and Peterhans, E. and Baumgartner, G.},
  title   = {Illusory contours and cortical neuron responses},
  journal = {Science},
  year    = {1984},
  volume  = {224},
  number  = {4654},
  pages   = {1260--1262},
  doi     = {10.1126/science.6539501}
}

@article{Kovacs1995JNeuro,
  author  = {Kov{\'a}cs, G{\'a}bor and Vogels, Rufin and Orban, Guy A.},
  title   = {Selectivity of macaque inferior temporal neurons for partially occluded shapes},
  journal = {Journal of Neuroscience},
  year    = {1995},
  volume  = {15},
  number  = {3},
  pages   = {1984--1997},
  doi     = {10.1523/JNEUROSCI.15-03-01984.1995}
}

@article{Lerner2002CerebCortex,
  author  = {Lerner, Y. and Hendler, T. and Malach, R.},
  title   = {Object-completion effects in the human lateral occipital complex},
  journal = {Cerebral Cortex},
  year    = {2002},
  volume  = {12},
  number  = {2},
  pages   = {163--177},
  doi     = {10.1093/cercor/12.2.163}
}

@article{Hu2024IJCV_CPNet,
  author  = {Hu, Xihang and Sun, Feng and Sun, Jian and Wang, Fang and Li, Hong},
  title   = {Cross-Modal Fusion and Progressive Decoding Network for RGB-D Salient Object Detection},
  journal = {International Journal of Computer Vision},
  year    = {2024},
  volume  = {132},
  number  = {8},
  pages   = {3067--3085},
  doi     = {10.1007/s11263-024-02020-y}
}

@ARTICLE{10339864,
  author={Zhang, Ni and Liu, Nian and Nan, Fang and Han, Junwei},
  journal={IEEE Transactions on Pattern Analysis and Machine Intelligence}, 
  title={CADC++: Advanced Consensus-Aware Dynamic Convolution for Co-Salient Object Detection}, 
  year={2024},
  volume={46},
  number={5},
  pages={2741-2757},
  doi={10.1109/TPAMI.2023.3336015}}

@article{wang2025novel,
  title={A novel embedded cross framework for high-resolution salient object detection},
  author={Wang, Baoyu and Yang, Mao and Cao, Pingping and Liu, Yan},
  journal={Applied Intelligence},
  volume={55},
  number={4},
  pages={277},
  year={2025},
  publisher={Springer}
}

@inproceedings{zhang2025cgcod,
  title={Cgcod: Class-guided camouflaged object detection},
  author={Zhang, Chenxi and Zhang, Qing and Wu, Jiayun and Pang, Youwei},
  booktitle={Proceedings of the 33rd ACM International Conference on Multimedia},
  pages={4369--4377},
  year={2025}
}

@inproceedings{yu2025sam,
  title={SAM-TTT: Segment Anything Model via Reverse Parameter Configuration and Test-Time Training for Camouflaged Object Detection},
  author={Yu, Zhenni and Zhao, Li and Xiao, Guobao and Zhang, Xiaoqin},
  booktitle={Proceedings of the 33rd ACM International Conference on Multimedia},
  pages={4030--4038},
  year={2025}
}

@ARTICLE{10623294,
  author={Yin, Bowen and Zhang, Xuying and Fan, Deng-Ping and Jiao, Shaohui and Cheng, Ming-Ming and Van Gool, Luc and Hou, Qibin},
  journal={IEEE Transactions on Pattern Analysis and Machine Intelligence}, 
  title={CamoFormer: Masked Separable Attention for Camouflaged Object Detection}, 
  year={2024},
  volume={46},
  number={12},
  pages={10362-10374},
  doi={10.1109/TPAMI.2024.3438565}}

@ARTICLE{10379651,
  author={Hu, Xihang and Zhang, Xiaoli and Wang, Fasheng and Sun, Jing and Sun, Fuming},
  journal={IEEE Transactions on Circuits and Systems for Video Technology}, 
  title={Efficient Camouflaged Object Detection Network Based on Global Localization Perception and Local Guidance Refinement}, 
  year={2024},
  volume={34},
  number={7},
  pages={5452-5465},
  doi={10.1109/TCSVT.2023.3349209}}

@inproceedings{risnet,
  title={Depth-aware concealed crop detection in dense agricultural scenes},
  author={Wang, Liqiong and Yang, Jinyu and Zhang, Yanfu and Wang, Fangyi and Zheng, Feng},
  booktitle={Proceedings of the IEEE/CVF Conference on Computer Vision and Pattern Recognition},
  pages={17201--17211},
  year={2024}
}

@ARTICLE{10007893,
  author={Lv, Yunqiu and Zhang, Jing and Dai, Yuchao and Li, Aixuan and Barnes, Nick and Fan, Deng-Ping},
  journal={IEEE Transactions on Circuits and Systems for Video Technology}, 
  title={Toward Deeper Understanding of Camouflaged Object Detection}, 
  year={2023},
  volume={33},
  number={7},
  pages={3462-3476},
  doi={10.1109/TCSVT.2023.3234578}}

@INPROCEEDINGS{10203727,
  author={He, Chunming and Li, Kai and Zhang, Yachao and Tang, Longxiang and Zhang, Yulun and Guo, Zhenhua and Li, Xiu},
  booktitle={2023 IEEE/CVF Conference on Computer Vision and Pattern Recognition}, 
  title={Camouflaged Object Detection with Feature Decomposition and Edge Reconstruction}, 
  year={2023},
  volume={},
  number={},
  pages={22046-22055},
  doi={10.1109/CVPR52729.2023.02111}}

@inproceedings{chen2024camodiffusion,
  title={CamoDiffusion: Camouflaged object detection via conditional diffusion models},
  author={Chen, Zhongxi and Sun, Ke and Lin, Xianming},
  booktitle={Proceedings of the AAAI Conference on Artificial Intelligence},
  volume={38},
  number={2},
  pages={1272--1280},
  year={2024}
}

@article{ji2023deep,
  title={Deep gradient learning for efficient camouflaged object detection},
  author={Ji, Ge-Peng and Fan, Deng-Ping and Chou, Yu-Cheng and Dai, Dengxin and Liniger, Alexander and Van Gool, Luc},
  journal={Machine Intelligence Research},
  volume={20},
  number={1},
  pages={92--108},
  year={2023},
  publisher={Springer}
}

@inproceedings{huang2023feature,
  title={Feature shrinkage pyramid for camouflaged object detection with transformers},
  author={Huang, Zhou and Dai, Hang and Xiang, Tian-Zhu and Wang, Shuo and Chen, Huai-Xin and Qin, Jie and Xiong, Huan},
  booktitle={Proceedings of the IEEE/CVF Conference on Computer Vision and Pattern Recognition},
  pages={5557--5566},
  year={2023}
}

@ARTICLE{10180211,
  author={Yan, Xinyu and Sun, Meijun and Han, Yahong and Wang, Zheng},
  journal={IEEE Transactions on Neural Networks and Learning Systems}, 
  title={Camouflaged Object Segmentation Based on Matching–Recognition–Refinement Network}, 
  year={2024},
  volume={35},
  number={11},
  pages={15993-16007},
  doi={10.1109/TNNLS.2023.3291595}}

@inproceedings{zoomnet,
  title={Zoom in and out: A mixed-scale triplet network for camouflaged object detection},
  author={Pang, Youwei and Zhao, Xiaoqi and Xiang, Tian-Zhu and Zhang, Lihe and Lu, Huchuan},
  booktitle={Proceedings of the IEEE/CVF Conference on Computer Vision and Pattern Recognition},
  pages={2160--2170},
  year={2022}
}

@ARTICLE{10102831,
  author={Hong, Lin and Wang, Xin and Zhang, Gan and Zhao, Ming},
  journal={IEEE Transactions on Image Processing}, 
  title={USOD10K: A New Benchmark Dataset for Underwater Salient Object Detection}, 
  year={2025},
  volume={34},
  number={},
  pages={1602-1615},
  doi={10.1109/TIP.2023.3266163}}

@ARTICLE{10091765,
  author={Wu, Zongwei and Allibert, Guillaume and Meriaudeau, Fabrice and Ma, Chao and Demonceaux, Cédric},
  journal={IEEE Transactions on Image Processing}, 
  title={HiDAnet: RGB-D Salient Object Detection via Hierarchical Depth Awareness}, 
  year={2023},
  volume={32},
  number={},
  pages={2160-2173},
  doi={10.1109/TIP.2023.3263111}}

@ARTICLE{10179145,
  author={Sun, Fuming and Ren, Peng and Yin, Bowen and Wang, Fasheng and Li, Haojie},
  journal={IEEE Transactions on Multimedia}, 
  title={CATNet: A Cascaded and Aggregated Transformer Network for RGB-D Salient Object Detection}, 
  year={2024},
  volume={26},
  number={},
  pages={2249-2262},
  doi={10.1109/TMM.2023.3294003}}

@inproceedings{cong2023point,
  title={Point-aware interaction and CNN-induced refinement network for RGB-D salient object detection},
  author={Cong, Runmin and Liu, Hongyu and Zhang, Chen and Zhang, Wei and Zheng, Feng and Song, Ran and Kwong, Sam},
  booktitle={Proceedings of the 31st ACM International Conference on Multimedia},
  pages={406--416},
  year={2023}
}

@INPROCEEDINGS{10657195,
  author={Zhang, Pingping and Yan, Tianyu and Liu, Yang and Lu, Huchuan},
  booktitle={2024 IEEE/CVF Conference on Computer Vision and Pattern Recognition}, 
  title={Fantastic Animals and Where to Find Them: Segment Any Marine Animal with Dual SAM}, 
  year={2024},
  volume={},
  number={},
  pages={2578-2587},
  doi={10.1109/CVPR52733.2024.00249}}

@ARTICLE{10697198,
  author={Mao, Yuxin and Zhang, Jing and Wan, Zhexiong and Tian, Xinyu and Li, Aixuan and Lv, Yunqiu and Dai, Yuchao},
  journal={IEEE Transactions on Circuits and Systems for Video Technology}, 
  title={Generative Transformer for Accurate and Reliable Salient Object Detection}, 
  year={2025},
  volume={35},
  number={2},
  pages={1041-1054},
  doi={10.1109/TCSVT.2024.3469286}}

@article{ma2025stamf,
  title={STAMF: Synergistic transformer and mamba fusion network for RGB-Polarization based underwater salient object detection},
  author={Ma, Qianwen and Li, Xiaobo and Li, Bincheng and Zhu, Zhen and Wu, Jing and Huang, Feng and Hu, Haofeng},
  journal={Information Fusion},
  volume={122},
  pages={103182},
  year={2025},
  publisher={Elsevier}
}

@ARTICLE{10744585,
  author={Jin, Jianhui and Jiang, Qiuping and Wu, Qingyuan and Xu, Binwei and Cong, Runmin},
  journal={IEEE Transactions on Circuits and Systems for Video Technology}, 
  title={Underwater Salient Object Detection via Dual-Stage Self-Paced Learning and Depth Emphasis}, 
  year={2025},
  volume={35},
  number={3},
  pages={2147-2160},
  doi={10.1109/TCSVT.2024.3491907}}

@ARTICLE{11018233,
  author={Zha, Mingfeng and Wang, Guoqing and Pei, Yunqiang and Li, Tianyu and Tang, Xiongxin and Li, Chongyi and Yang, Yang and Tao Shen, Heng},
  journal={IEEE Transactions on Image Processing}, 
  title={Heterogeneous Experts and Hierarchical Perception for Underwater Salient Object Detection}, 
  year={2025},
  volume={34},
  number={},
  pages={3703-3717},
  doi={10.1109/TIP.2025.3572760}}

@inproceedings{li2025fscdiff,
  title={FSCDiff: Frequency-Spatial Entangled Conditional Diffusion model for Underwater Salient Object Detection},
  author={Li, Hua and Lin, Gaowei and Li, Zhiyuan and Kwong, Sam and Cong, Runmin},
  booktitle={Proceedings of the 33rd ACM International Conference on Multimedia},
  pages={8379--8388},
  year={2025}
}

@article{CTF-Net,
title = {An effective CNN and Transformer fusion network for camouflaged object detection},
journal = {Computer Vision and Image Understanding},
volume = {259},
pages = {104431},
year = {2025},
issn = {1077-3142},
author = {Dongdong Zhang and Chunping Wang and Huiying Wang and Qiang Fu and Zhaorui Li},
}

@article{bclnet,
title = {Boundary-and-object collaborative learning network for camouflaged object detection},
journal = {Image and Vision Computing},
volume = {161},
pages = {105596},
year = {2025},
issn = {0262-8856},
author = {Chenyu Zhuang and Qing Zhang and Chenxi Zhang and Xinxin Yuan},
}

@inproceedings{fan2020camouflaged,
  title={Camouflaged object detection},
  author={Fan, Deng-Ping and Ji, Ge-Peng and Sun, Guolei and Cheng, Ming-Ming and Shen, Jianbing and Shao, Ling},
  booktitle={Proceedings of the IEEE/CVF Conference on Computer Vision and Pattern Recognition},
  pages={2777--2787},
  year={2020}
}

@article{le2019anabranch,
  title={Anabranch network for camouflaged object segmentation},
  author={Le, Trung-Nghia and Nguyen, Tam V and Nie, Zhongliang and Tran, Minh-Triet and Sugimoto, Akihiro},
  journal={Computer Vision and Image Understanding},
  volume={184},
  pages={45--56},
  year={2019},
  publisher={Elsevier}
}

@INPROCEEDINGS{9577641,
  author={Lv, Yunqiu and Zhang, Jing and Dai, Yuchao and Li, Aixuan and Liu, Bowen and Barnes, Nick and Fan, Deng-Ping},
  booktitle={2021 IEEE/CVF Conference on Computer Vision and Pattern Recognition}, 
  title={Simultaneously Localize, Segment and Rank the Camouflaged Objects}, 
  year={2021},
  volume={},
  number={},
  pages={11586-11596},
  doi={10.1109/CVPR46437.2021.01142}}

@article{skurowski2018animal,
  title={Animal camouflage analysis: Chameleon database},
  author={Skurowski, Przemys{\l}aw and Abdulameer, Hassan and B{\l}aszczyk, Jakub and Depta, Tomasz and Kornacki, Adam and Kozie{\l}, Przemys{\l}aw},
  journal={Unpublished Manuscript},
  volume={2},
  number={6},
  pages={7},
  year={2018}
}
%



%

\vspace{-30pt}
\begin{IEEEbiography}[{\includegraphics[width=1in,height=1.25in,clip,keepaspectratio]{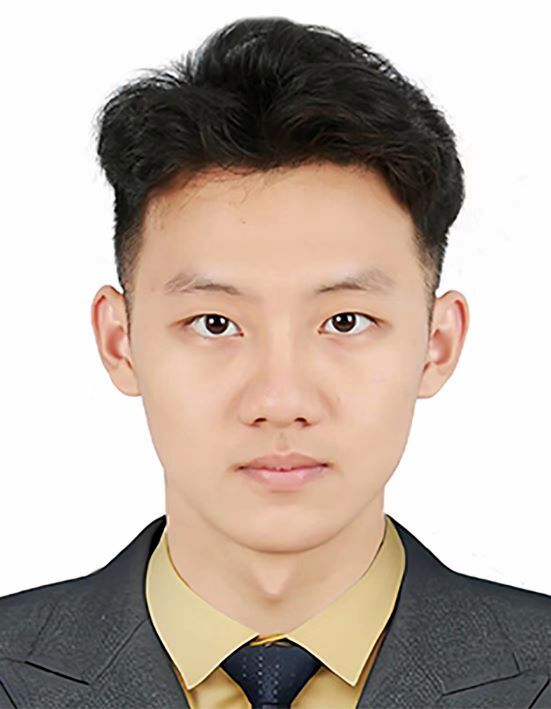}}]{Yu Zhang} received the M.Eng. degree from the School of Intelligence Science and Technology, University of Science and Technology Beijing, Beijing, China, in 2022. He is currently working toward the Ph.D. degree in computer science as a member of the Informatics 6, Technical University of Munich, Munich, Germany.\\
\indent His current research interests include perception, optimization, and control in robotics.
 \end{IEEEbiography}

 \vspace{-30pt}
\begin{IEEEbiography}[{\includegraphics[width=1in,height=1.25in,clip,keepaspectratio]{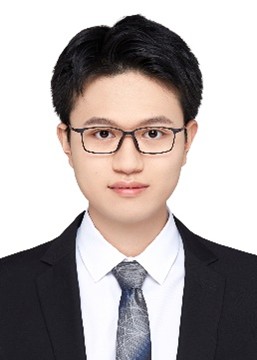}}]{Haoan Ping} received his B.E. from Harbin Institute of Technology in 2021 and his M.Sc. from the Technical University of Munich in 2025. He is currently working at Bosch in the field of autonomous driving, with his research focusing on 2D/3D visual perception, trajectory prediction, and related technologies.
 \end{IEEEbiography}

\vspace{-30pt}

\begin{IEEEbiography}[{\includegraphics[width=1in,height=1.25in,clip,keepaspectratio]{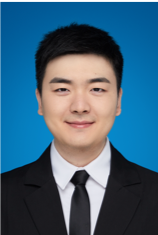}}]{Yuchen Li} received the B.E. degree from the University of Science and Technology Beijing in 2016, and the M.E. degrees from Beihang University in 2020. He is pursuing a Ph.D. degree at Hong Kong Baptist University. He is an intern at WAYTOUS. His research interest covers computer vision, 3D object detection, and autonomous driving.
\end{IEEEbiography}
 

\vspace{-30pt}

\begin{IEEEbiography}[{\includegraphics[width=1in,height=1.25in,clip,keepaspectratio]{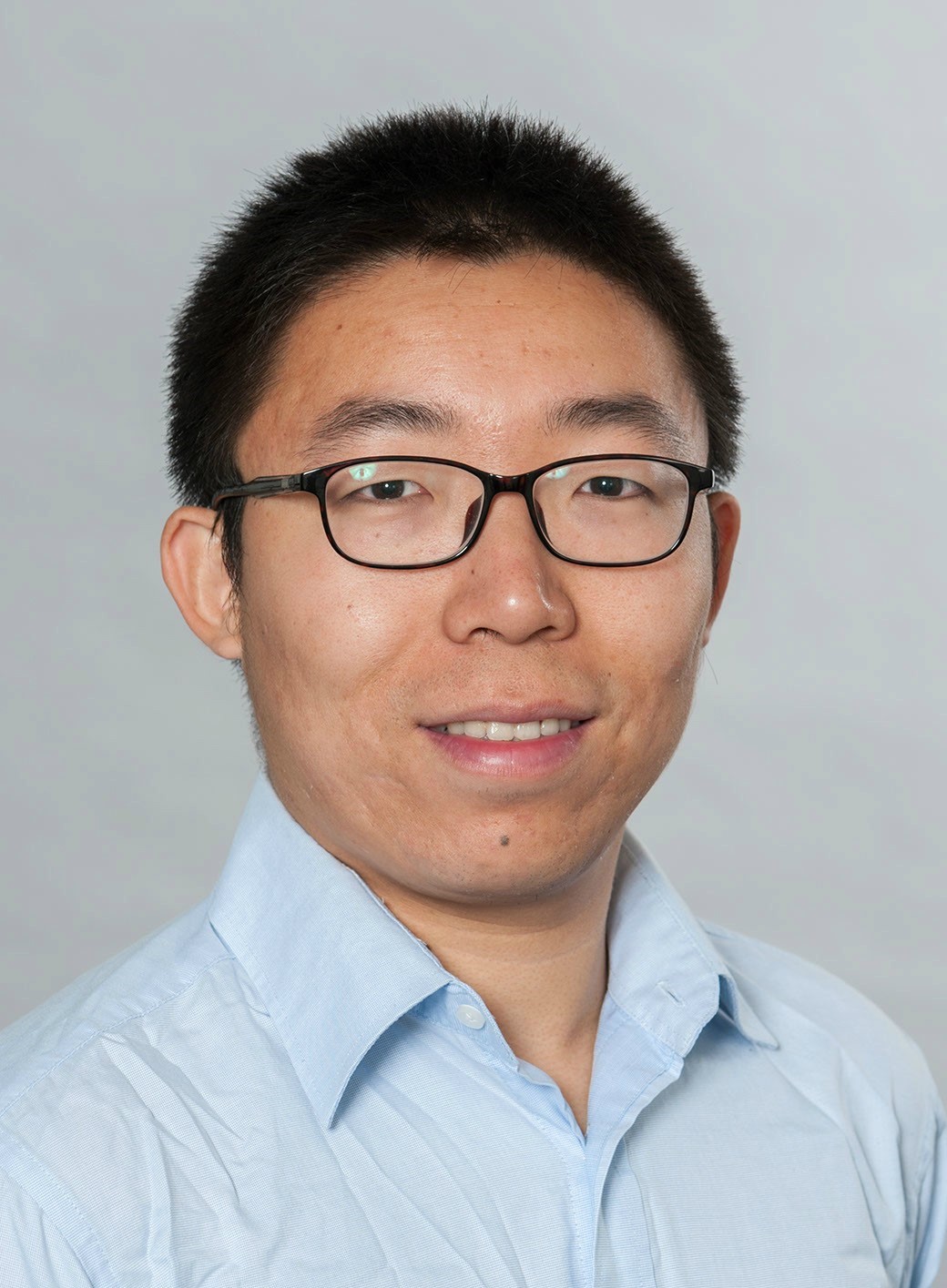}}]{Zhenshan Bing} received the B.S. degree in mechanical design manufacturing and automation and the M.Eng. degree in mechanical engineering from the Harbin Institute of Technology, Harbin, China, in 2013 and 2015, respectively, and the Doctoral degree in computer science from the Technical University of Munich, Munich, Germany, in 2019., He is currently a Postdoctoral Researcher with Informatics 6, Technical University of Munich. His research investigates the snake-like robot which is controlled by artificial neural networks and its related applications.
\end{IEEEbiography}

\vspace{-30pt}

\begin{IEEEbiography}[{\includegraphics[width=1in,height=1.25in,clip,keepaspectratio]{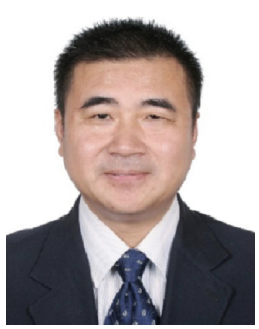}}]{Fuchun Sun} (Fellow, IEEE) is currently a Full Professor with the Department of Computer Science and Technology, Tsinghua University, Beijing, China. His research interests include robotic perception and skill learning, cross-modal learning, and robot
dexterous operations. Dr. Sun is a Chinese Association for Artificial Intelligence/Chinese Association of Automation (CAAI/CAA) Fellow. He was a recipient of the Excellent Doctoral Dissertation Prize of China in 2000 by MOE of China and the Choon-Gang Academic Award by Korea in 2003, and was recognized as a Distinguished Young Scholar in 2006 by the Natural Science Foundation of China. He serves as the Vice Chairperson for the Chinese Association for Artificial Intelligence.
\end{IEEEbiography}

\vspace{-30pt}

\begin{IEEEbiography}[{\includegraphics[width=1in,height=1.25in,clip,keepaspectratio]{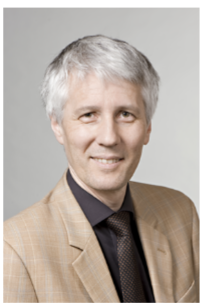}}]{Alois Knoll} (Fellow, IEEE) received his diploma (M.Sc.) degree in Electrical/Communications Engineering from the University of Stuttgart, Germany, in 1985 and his Ph.D. (summa cum laude) in Computer Science from Technical University of Berlin, Germany, in 1988. He served on the faculty of the Computer Science department at TU Berlin until 1993. He joined the University of Bielefeld, Germany as a full professor and served as the director of the Technical Informatics research group until 2001. Since 2001, he has been a professor at the Department of Informatics, Technical University of Munich (TUM), Germany. His research interests include cognitive, medical and sensor-based robotics, multi-agent systems, data fusion, as well as simulation systems for robotics and traffic.
\end{IEEEbiography}





\balance

\end{document}